\documentclass[11pt, a4paper]{thuc3i}
\usepackage[sort&compress]{natbib}
\setheadertext{A Survey of Weight Space Learning: Understanding, Representation, and Generation}

\usepackage[utf8]{inputenc} %
\usepackage[T1]{fontenc}    %
\usepackage{hyperref}       %
\usepackage{url}            %
\usepackage{tabularx}
\usepackage{makecell}
\usepackage{array,ragged2e,booktabs}
\newcolumntype{P}[1]{>{\RaggedRight\arraybackslash}p{#1}}
\usepackage{longtable}
\usepackage{amsfonts}       %
\usepackage{amsthm}         %

\usepackage{nicefrac}       %
\usepackage{microtype}      %
\usepackage{xcolor}         %

\usepackage{algorithm}
\usepackage{algorithmicx}
\usepackage{algpseudocode}

\definecolor{darkblue}{rgb}{0, 0, 0.5}
\hypersetup{colorlinks=true, citecolor=darkblue, linkcolor=darkblue, urlcolor=darkblue}

\usepackage{xurl}

\usepackage{soul}
\usepackage{amsmath}
\usepackage{subcaption}
\usepackage{graphicx}
\usepackage{changes}
\usepackage{amsmath,mathtools}
\usepackage{changes}
\usepackage{pifont}
\usepackage{tocloft}

\usepackage{lipsum}
\usepackage{colortbl}
\usepackage{wrapfig}
\usepackage{xspace}
\usepackage{multirow}
\usepackage{enumitem}

\usepackage{pifont}
\usepackage{listings}
\usepackage{fontawesome5} 

\usepackage{wrapfig}
\usepackage{caption}

\usepackage{makecell}

\usepackage{tabularx}
\usepackage{afterpage}

\usepackage[edges]{forest}
\usepackage[normalem]{ulem}
\usepackage{caption}
\usepackage{CJKutf8}
\usepackage{bbding}
\usepackage[most,skins,theorems]{tcolorbox}
\usepackage{longtable}
\usepackage{booktabs}
\usepackage{multirow}
\usepackage{hyperref}
\usepackage{array}
\usepackage[bottom]{footmisc}

\usepackage[tikz]{bclogo}
\usepackage[framemethod=tikz]{mdframed}
\definecolor{bgblue}{RGB}{245,243,253}
\definecolor{ttblue}{RGB}{91,194,224}

\mdfdefinestyle{mystyle}{%
  rightline=true,
  innerleftmargin=10,
  innerrightmargin=10,
  outerlinewidth=3pt,
  topline=false,
  rightline=true,
  bottomline=false,
  skipabove=\topsep,
  skipbelow=\topsep
}

\newtcolorbox{myboxi}[1][]{
  breakable,
  title=#1,
  colback=red!5,
  colbacktitle=red!5,
  coltitle=black,
  fonttitle=\bfseries,
  bottomrule=0pt,
  toprule=0pt,
  leftrule=2pt,
  rightrule=2pt,
  titlerule=0pt,
  arc=0pt,
  outer arc=0pt,
  colframe=red,
}

\newtcolorbox{myboxnote}[1][]{
  breakable,
  title=#1,
  colback=orange!0,
  colbacktitle=orange!0,
  coltitle=black,
  fonttitle=\bfseries,
  bottomrule=0pt,
  toprule=0pt,
  leftrule=2pt,
  rightrule=2pt,
  titlerule=0pt,
  arc=0pt,
  outer arc=0pt,
  colframe=orange,
}

\usepackage{epigraph}

\newtcolorbox{myboxii}[1][]{
  breakable,
  freelance,
  title=#1,
  colback=white,
  colbacktitle=white,
  coltitle=black,
  fonttitle=\bfseries,
  bottomrule=0pt,
  boxrule=0pt,
  colframe=white,
  overlay unbroken and first={
  \draw[red!75!black,line width=3pt]
    ([xshift=5pt]frame.north west) -- 
    (frame.north west) -- 
    (frame.south west);
  \draw[red!75!black,line width=3pt]
    ([xshift=-5pt]frame.north east) -- 
    (frame.north east) -- 
    (frame.south east);
  },
  overlay unbroken app={
  \draw[red!75!black,line width=3pt,line cap=rect]
    (frame.south west) -- 
    ([xshift=5pt]frame.south west);
  \draw[red!75!black,line width=3pt,line cap=rect]
    (frame.south east) -- 
    ([xshift=-5pt]frame.south east);
  },
  overlay middle and last={
  \draw[red!75!black,line width=3pt]
    (frame.north west) -- 
    (frame.south west);
  \draw[red!75!black,line width=3pt]
    (frame.north east) -- 
    (frame.south east);
  },
  overlay last app={
  \draw[red!75!black,line width=3pt,line cap=rect]
    (frame.south west) --
    ([xshift=5pt]frame.south west);
  \draw[red!75!black,line width=3pt,line cap=rect]
    (frame.south east) --
    ([xshift=-5pt]frame.south east);
  },
}

\definecolor{myblue}{rgb}{0.9, 0.1, 0.94}
\definecolor{mygreen}{rgb}{0.64, 0.56, 0.88}
\definecolor{myyellow}{rgb}{0.68, 0.6, 0.1}
\definecolor{fancygreen}{rgb}{0.33, 0.68, 0.20}
\definecolor{salmon}{rgb}{0.94, 0.52, 0.49}
\definecolor{tablegreen}{rgb}{0.82, 0.94, 0.75}
\definecolor{tableblue}{rgb}{0.81, 0.90, 0.94}
\definecolor{tablered}{rgb}{0.97, 0.85, 0.85}
\definecolor{tableorange}{rgb}{0.96, 0.85, 0.81}

\definecolor{myorange}{rgb}{1.0, 0.49, 0.0}	
\definecolor{tlgreen}{rgb}{0.33, 0.68, 0.20}

\newenvironment{itemize*}%
 {\leftmargini=10pt\begin{itemize}%
  \setlength{\itemsep}{0pt}%
  \setlength{\parskip}{0pt}%
  }%
 {\end{itemize}}
\newenvironment{enumerate*}%
 {\begin{enumerate}%
  \setlength{\itemsep}{0pt}%
  \setlength{\parskip}{0pt}}%
 {\end{enumerate}}

\tikzset{%
    every node/.style={font=\tiny},
    parent/.style =          {align=center,text width=2cm,rounded corners=3pt, line width=0.3mm, fill=gray!10,draw=gray!80},
    child/.style =           {align=center,text width=1.5cm,rounded corners=3pt, fill=blue!10,draw=blue!80,line width=0.3mm},
    grandchild/.style =      {align=center,text width=1.5cm,rounded corners=3pt},
    greatgrandchild/.style = {align=center,text width=1.5cm,rounded corners=3pt},
    greatgrandchild2/.style = {align=center,text width=1.5cm,rounded corners=3pt},    
    referenceblock/.style =  {align=center,text width=1.5cm,rounded corners=2pt},
    pretrain/.style =           {align=center,text width=2.0cm,rounded corners=3pt, fill=blue!10,draw=blue!80,line width=0.3mm},   
    pretrain_work/.style =           {align=center, text width=8.5cm,rounded corners=3pt, fill=blue!10,draw=blue!0,line width=0.3mm},  
    template/.style =           {align=center,text width=2.0cm,rounded corners=3pt, fill=red!10,draw=red!80,line width=0.3mm},   
    template_work/.style =           {align=center,text width=8.5cm,rounded corners=3pt, fill=red!10,draw=red!0,line width=0.3mm},    
    answer/.style =           {align=center,text width=2.0cm,rounded corners=3pt, fill= cyan!10,draw= cyan!80,line width=0.3mm},   
    answer_work/.style =           {align=center,text width=8.5cm,rounded corners=3pt, fill= cyan!10,draw= cyan!0,line width=0.3mm},      
    multiple/.style =           {align=center,text width=2.0cm,rounded corners=3pt, fill= orange!10,draw= orange!80,line width=0.3mm},   
    multiple_work/.style =           {align=center,text width=8.5cm,rounded corners=3pt, fill= orange!10,draw= orange!0,line width=0.3mm},        
    tuning/.style =           {align=center,text width=2.0cm,rounded corners=3pt, fill= magenta!10,draw= magenta!80,line width=0.3mm},   
    tuning_work/.style =           {align=center,text width=8.5cm,rounded corners=3pt, fill= magenta!10,draw= magenta!0,line width=0.3mm},      
    analysis/.style =         {align=center,text width=2.0cm,rounded corners=3pt,fill=green!10,draw=green!80,line width=0.3mm},
    analysis_work/.style =    {align=center,text width=8.5cm,rounded corners=3pt,fill=green!10,draw=green!0 ,line width=0.3mm},
    integration/.style =      {align=center,text width=2.0cm,rounded corners=3pt,fill=purple!10,draw=purple!80,line width=0.3mm},
    integration_work/.style = {align=center,text width=8.5cm,rounded corners=3pt,fill=purple!10,draw=purple!0 ,line width=0.3mm},
}

\usepackage{tikz}

\newcommand{\lstbg}[3][0pt]{{\fboxsep#1\colorbox{#2}{\strut #3}}}

\lstdefinelanguage{diff}{
  basicstyle=\ttfamily\small,
  morecomment=[f][\lstbg{red!20}]-,
  morecomment=[f][\lstbg{green!20}]+,
}
\lstdefinelanguage{diffpython}{
  language=diff,
  morekeywords={def, if, else, for, while, return, import, from, as, class, with, try, except, finally, raise, lambda, and, or, not, in, is, None, True, False},
  morecomment=[l]{\#},
  morestring=[b]",
  morestring=[b]',
}

\definecolor{darkgreen}{RGB}{50,100,0}
\definecolor{darkred}{RGB}{200, 0, 0}
\definecolor{lightblue}{RGB}{220,235,250}

\tcbset{
  takeawaysbox/.style={
    title=Takeaways,
    colback=lightblue!80,
    colframe=black,
    fonttitle=\bfseries\small,
    coltitle=white,
    colbacktitle=black,
    enhanced,
    attach boxed title to top left={xshift=2.5mm,yshift=-2.5mm},
    boxed title style={rounded corners, size=small, colframe=black, colback=black},
    width=\linewidth,
    arc=3.5mm
  }
}

\definecolor{darkgreen}{RGB}{50,100,0}
\definecolor{darkred}{RGB}{200, 0, 0}

\NewDocumentCommand{\kaiyan}
{ mO{} }{\textcolor{purple}{\textsuperscript{\textit{kaiyan}}\textsf{\textbf{\small[#1]}}}}
\NewDocumentCommand{\yuxin}
{ mO{} }{\textcolor{cyan}{\textsuperscript{\textit{yuxin}}\textsf{\textbf{\small[#1]}}}}
\NewDocumentCommand{\bx}
{ mO{} }{\textcolor{green}{\textsuperscript{\textit{bx}}\textsf{\textbf{\small[#1]}}}}
\NewDocumentCommand{\at}
{ mO{} }{\textcolor{red}{\textsuperscript{\textit{AT}}\textsf{\textbf{\small[#1]}}}}
\NewDocumentCommand{\re}
{ mO{} }{\textcolor{blue}{\textsuperscript{\textit{RE}}\textsf{\textbf{\small[#1]}}}}
\NewDocumentCommand{\ybsun}
{ mO{} }{\textcolor{magenta}{\textsuperscript{\textit{youbang}}\textsf{\textbf{\small[#1]}}}}
\NewDocumentCommand{\runze}
{ mO{} }{\textcolor{orange}{\textsuperscript{\textit{runze}}\textsf{\textbf{\small[#1]}}}}

\definecolor{darkgreen}{RGB}{0,100,0} 
\NewDocumentCommand{\add}
{ mO{} }{\textcolor{darkgreen}{\textsuperscript{\textit{Maybe Consider Discuss}}\textsf{\textbf{[#1]}}}}

\setlist[itemize]{leftmargin=20pt}

\usepackage[edges]{forest}
\definecolor{hidden-blue}{RGB}{194,232,247}
\definecolor{hidden-black}{RGB}{20,68,106}

\usepackage{tabularx, booktabs, xcolor, colortbl}
\usepackage{pifont}   %
\usepackage{xstring}  %

\usepackage{multirow}
\usepackage{hyperref}
\usepackage{fontawesome5}

\usepackage[export]{adjustbox}

\definecolor{yes}{HTML}{C6EFCE}      %
\definecolor{no}{HTML}{FFC7CE}       %
\definecolor{partial}{HTML}{FFEB9C}  %
\definecolor{external}{HTML}{D9E1F2} %
\definecolor{hdr}{HTML}{F2F2F2}

\usepackage{amssymb}
\newcommand{\cmark}{\textcolor{darkgreen}{\boldmath$\checkmark$}}
\newcommand{\xmark}{\textcolor{darkred}{\boldmath$\times$}}

\newcommand{\cellstatus}[1]{%
  \begingroup
  \StrTrim{#1}[\statusval]%
  \IfStrEq{\statusval}{Yes}{\cellcolor{yes}\cmark}{}%
  \IfStrEq{\statusval}{No}{\cellcolor{no}\xmark}{}%
  \IfBeginWith{\statusval}{Yes (}{\cellcolor{yes}\cmark~\textit{\statusval\unskip}}{}%
  \IfStrEq{\statusval}{Partial}{\cellcolor{partial}\textbf{Partial}}{}%
  \IfStrEq{\statusval}{External}{\cellcolor{external}\textbf{External}}{}%
  \endgroup
}

\newcommand{\tstyle}[1]{\underline{\textit{#1}}}

\usepackage{array,booktabs,tabularx}
\newcolumntype{C}[1]{>{\centering\arraybackslash}m{#1}}
\newcolumntype{L}[1]{>{\raggedright\arraybackslash}m{#1}}

\renewcommand{\arraystretch}{1.05}

\title{A Survey of Weight Space Learning: Understanding, Representation, and Generation}

\author{%
    Xiaolong Han$^{1,*}$, 
    Zehong Wang$^{2,*,\dagger}$, 
    Bo Zhao$^{3}$, 
    Binchi Zhang$^{4}$, 
    Jundong Li$^{4}$, 
    Damian Borth$^{5}$, 
    Rose Yu$^{3}$, 
    \quad\quad\quad 
    Haggai Maron$^{6,7}$, 
    Yanfang Ye$^{2}$, 
    Lu Yin$^{1}$, 
    Ferrante Neri$^{1,\dagger}$
    \vspace{1mm} \\
    $^1$ University of Surrey \quad
    $^2$ University of Notre Dame \quad
    $^3$ University of California San Diego \quad
    $^4$ University of Virginia \\
    $^5$ University of St.Gallen  \quad
    $^6$ Technion \quad
    $^7$ Nvidia 
    \vspace{1mm} \\
    \textbf{$^*$ Equal Contribution.}~~ $\dagger$ \textbf{Corresponding Authors.}
    \vspace{1mm} \\
    \faEnvelope[regular]~\texttt{zwang43@nd.edu}  \quad \faEnvelope[regular]~\texttt{f.neri@surrey.ac.uk} \quad 
    \faGithub~\href{https://github.com/Zehong-Wang/Awesome-Weight-Space-Learning}{Zehong-Wang/Awesome-Weight-Space-Learning}
}

\begin{abstract}

Neural network weights are typically viewed as the end product of training, while most deep learning research focuses on data, features, and architectures. However, recent advances show that the set of all possible weight values (weight space) itself contains rich structure: pretrained models form organized distributions, exhibit symmetries, and can be embedded, compared, or even generated. Understanding such structures has tremendous impact on how neural networks are analyzed and compared, and on how knowledge is transferred across models, beyond individual training instances. This emerging research direction, which we refer to as Weight Space Learning (WSL), treats neural weights as a meaningful domain for analysis and modeling.
This survey provides the first unified taxonomy of WSL. We categorize existing methods into three core dimensions: Weight Space Understanding (WSU), which studies the geometry and symmetries of weights; Weight Space Representation (WSR), which learns embeddings over model weights; and Weight Space Generation (WSG), which synthesizes new weights through hypernetworks or generative models. We further show how these developments enable practical applications, including model retrieval, continual and federated learning, neural architecture search, and data-free reconstruction.
By consolidating fragmented progress under a coherent framework, this survey highlights weight space as a learnable, structured domain with growing impact across model analysis, transferring, and weight generation.
We release an accompanying resource at \url{https://github.com/Zehong-Wang/Awesome-Weight-Space-Learning}.

\end{abstract}

\begin{document}

\begin{figure}[b!]
    \centering
    \vspace{-30pt}
    \includegraphics[width=1\linewidth, trim={0.8cm 0.5cm 1cm 0cm}, clip, scale=0.5]
    {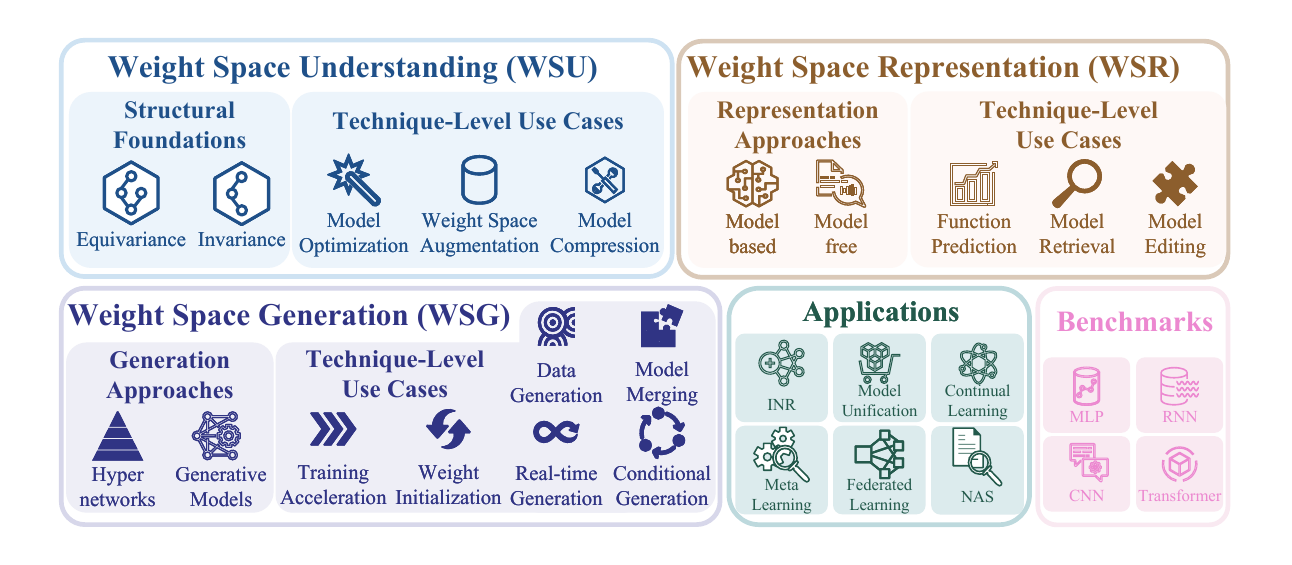}
    \caption{\textbf{Overview of the survey.} We organize weight space learning into three dimensions: understanding weight-space properties, learning compact representations, generating weights via auxiliary models. 
    In addition, it consolidates applications to related domains and benchmarks, paving the way for future research.
    }
    \vspace{-20pt}
    \label{fig:taxonomy}
\end{figure}

\maketitle

\newpage
\begingroup
\setlength{\baselineskip}{1.25\baselineskip}
\tableofcontents
\endgroup

\newpage

\section{Introduction}\label{sec:introduction}

Deep neural networks (DNNs) have revolutionized numerous fields, ranging from computer vision~\citep{voulodimos2018deep} and natural language processing~\citep{khurana2023natural} to decision-making systems~\citep{zhang2025survey}. This progress has been driven by advances in computational hardware, the scaling of datasets and model architectures, and continual innovations in training and optimization techniques~\citep{ gower2019sgd,moradi2020survey}. As a result, modern machine learning practice has produced an unprecedented volume of trained neural networks, publicly released through large-scale model repositories and model zoos. These pretrained weights, accumulated across tasks, architectures, and training regimes, now constitute a rich and rapidly growing source of information.
This widespread availability of trained models suggests a shift in perspective: pretrained weights themselves can be viewed as a new data modality. Beyond encoding task-specific knowledge, weights reflect architectural inductive biases, optimization dynamics, and structural regularities of the learning process. Unlike input features that describe data, weights describe the learners of data~\citep{eilertsen2020classifying,unterthiner2020predicting, schurholt2021self}. This observation raises a fundamental question:
\begin{quote}
    \vspace{-5pt}
    \begin{center}
        \textbf{\textit{Can we treat the weight space itself as a meaningful, learnable domain, and apply machine learning directly to collections of trained models?}}
    \end{center}
    \vspace{-5pt}
\end{quote}
Recent studies have begun to explore this idea by analyzing, representing, and even generating neural network weights~\citep{schurholt2022hyper,navon2023equivariant,erkocc2023hyperdiffusion}. This emerging paradigm, which we refer to as Weight Space Learning (WSL), treats model parameters not merely as optimization artifacts, but as structured objects amenable to modeling and learning.

Early work on neural network weights focused largely on analytical perspectives: studying symmetries, invariances, and reparameterizations to improve optimization or reduce redundancy~\citep{sourek2021lossless,ganev2021universal,saul2023weightbalancing,zhao2022symmetry}. While these studies revealed important structural properties, they treated weights as static end products of training. This view limited our ability to compare different models, capture relationships across training runs, or leverage the growing abundance of pretrained networks.
A key shift occurred when pretrained models themselves began to be treated as data. With the emergence of large model zoos, researchers started learning representations across model weights rather than hand-deriving them. Early approaches such as~\citep{eilertsen2020classifying,unterthiner2020predicting,schurholt2021self} directly operated on model parameters for tasks like retrieval and performance prediction. Subsequent work recognized the need to respect weight symmetries (e.g., neuron permutation invariance~\citep{hecht1990algebraic}), leading to permutation-invariant and graph-based encoders~\citep{navon2023equivariant,zhou2023permutation,kofinas2024graph,lim2024graph} and probing methods that avoid explicit alignment~\citep{kahana2025deep,herrmann2024learning}. More recently, generative models such as hypernetworks~\citep{ha2017hypernetworks,zhang2018graph,schurholt2022hyper} and diffusion-based weight generators~\citep{zhang2024metadiff,erkocc2023hyperdiffusion} have shifted the perspective once more: weights are not merely to be analyzed or represented, but can be generated directly.
Together, these developments establish weight space as a meaningful learning domain shared across architectures, and rich enough to support representation learning and generative modeling.

\begin{figure}[htbp]
    \centering
    \includegraphics[width=1\linewidth, trim={0.5cm 0.8cm 1cm 0.5cm}, clip, scale=0.1]
    {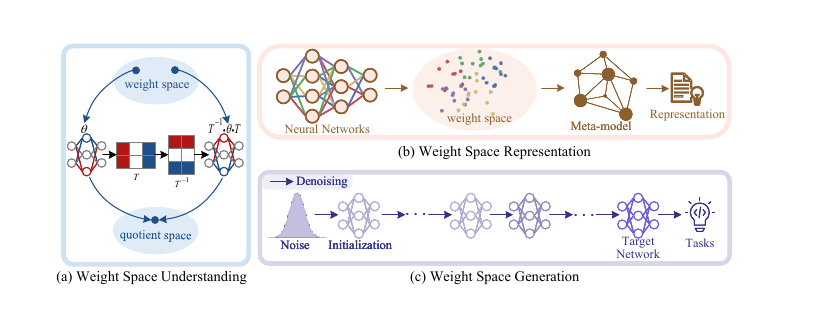}
    \caption{\textbf{Conceptual overview of weight space learning.} (a) Weight space understanding analyzes symmetries in weight space and introduces quotient spaces to reduce redundancy and clarify functional equivalence. (b) Weight space representation learns compact embeddings of neural weights, discriminating symmetry in weight space well. (c) Weight space generation synthesizes network weights, supporting efficient model creation and adaptation.}
    \label{fig:framework}
\end{figure}

Despite the flourish of ideas, existing studies in this area remain conceptually scattered. Researchers have approached the topic from diverse angles, ranging from symmetry analysis and model compression to generative modeling of weights, but often under different names and frameworks. As a result, the field suffers from terminological inconsistency: similar ideas are described with disparate notions such as neural functional networks~\citep{zhou2023permutation} and neural representation for neural networks~\citep{ashkenazi2023nern}. This fragmentation obscures the underlying connections between methods and makes it difficult to accumulate coherent knowledge. Moreover, even though recent surveys have discussed partial aspects of this topic, such as symmetry in weight space~\citep{zhao2025symmetry} or model retrieval~\citep{pal2024model}, there does not exist a unified taxonomy that captures the full spectrum of weight space learning. 

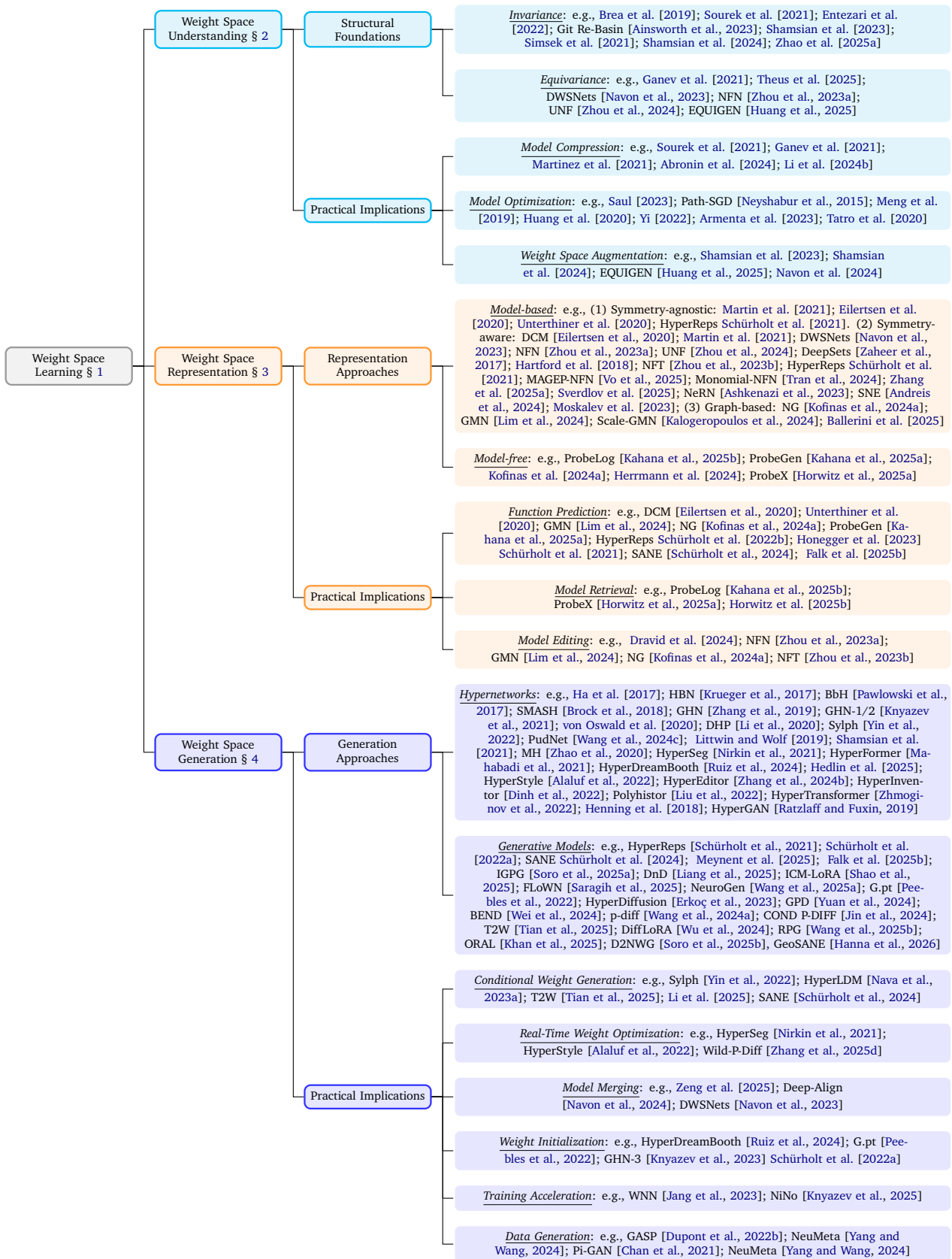
\begin{figure}
\footnotesize
\vspace{0cm}
\centering
\begin{adjustbox}{max width=1\textwidth}
\begin{forest}
    for tree={
        forked edges,
        grow'=0,
        draw,
        rounded corners,
        node options={align=center,},
        text width=2cm,
        s sep=6pt,
        calign=child edge, calign child=(n_children()+1)/2,
    },
    [Weight Space Learning~\S~\ref{sec:introduction}, fill=gray!45, parent
        [Weight Space Understanding~\S~\ref{section:wsu}, for tree={fill=blue!45, answer}
            [Structural Foundations, answer
                [{\tstyle{Invariance}: e.g., \citet{brea2019weight}; \citet{sourek2021lossless}; \citet{entezari2022the}; Git Re-Basin~\citep{ainsworth2023git}; \citet{shamsian2023data}; \citet{simsek2021geometry}; \citet{shamsian2024improved}; \citet{zhao2025understanding}}
                , answer_work]
                [{\tstyle{Equivariance}: e.g., \citet{ganev2021universal}; \citet{theus2025generalized}; DWSNets~\citep{navon2023equivariant}; NFN~\citep{zhou2023permutation}; UNF~\citep{zhou2024universal}; EQUIGEN~\citep{huang2025few}}
                , answer_work]
            ]
            [Practical Implications, answer
                [{\tstyle{Model Compression}: e.g., \citet{sourek2021lossless}; \citet{ganev2021universal}; \citet{martinez2021permute}; \citet{abronin2024tqcompressor}; \citet{li2024merge}}
                , answer_work]
                [{\tstyle{Model Optimization}: e.g., \citet{saul2023weightbalancing}; Path-SGD~\citep{neyshabur2015path}; \citet{meng2018gsgd}; \citet{huang2020projection}; \citet{yi2022accelerating}; \citet{armenta2023neural}; \citet{tatro2020optimizing}}
                , answer_work]
                [{\tstyle{Weight Space Augmentation}: e.g., \citet{shamsian2023data}; \citet{shamsian2024improved}; EQUIGEN~\citep{huang2025few}; \citet{navon2024equivariant}}
                , answer_work]
            ]
        ]
        [Weight Space Representation~\S~\ref{section:wsr}, for tree={fill=blue!45, multiple}
            [Representation Approaches , multiple
                [{\tstyle{Model-based}: e.g., (1) Symmetry-agnostic: \citet{martin2021predicting}; \citet{eilertsen2020classifying}; \citet{unterthiner2020predicting}; HyperReps~\citet{schurholt2021self}. (2) Symmetry-aware: DCM~\citep{eilertsen2020classifying}; \citet{martin2021predicting}; DWSNets~\citep{navon2023equivariant}; NFN~\citep{zhou2023permutation}; UNF~\citep{zhou2024universal}; DeepSets~\citep{zaheer2017deep}; \citet{hartford2018deep}; NFT~\citep{zhou2023neural}; HyperReps~\citet{schurholt2021self}; MAGEP-NFN~\citep{vo2025equivariant}; Monomial-NFN~\citep{tran2024monomial}; \citet{zhang2025beyond}; \citet{sverdlov2025revisiting}; NeRN~\citep{ashkenazi2023nern}; SNE~\citep{andreis2024set}; \citet{moskalev2023genuine}; (3) Graph-based: NG~\citep{kofinas2024graph}; GMN~\citep{lim2024graph}; Scale-GMN~\citep{kalogeropoulos2024scale}; \citet{ballerini2025weight}}
                , multiple_work]
                [{\tstyle{Model-free}: e.g., ProbeLog~\citep{kahana2025can}; ProbeGen~\citep{kahana2025deep}; ~\citet{kofinas2024graph}; \citet{herrmann2024learning}; ProbeX~\citep{horwitz2025learning}}
                , multiple_work]
            ]
            [Practical Implications , multiple
                [{\tstyle{Function Prediction}: e.g., DCM~\citep{eilertsen2020classifying}; \citet{unterthiner2020predicting}; GMN~\citep{lim2024graph}; NG~\citep{kofinas2024graph}; ProbeGen~\citep{kahana2025deep}; HyperReps~\citet{schurholt2022model}; \citet{honegger2023sparsified} \citet{schurholt2021self}; SANE~\citep{schurholt2024towards}; ~\citet{falk2025learning}}
                , multiple_work]
                [{\tstyle{Model Retrieval}: e.g., ProbeLog~\citep{kahana2025can}; ProbeX~\citep{horwitz2025learning}; \citet{horwitz2025we}}
                , multiple_work]
                [{\tstyle{Model Editing}: e.g., ~\citet{dravid2024interpreting}; NFN~\citep{zhou2023permutation}; GMN~\citep{lim2024graph}; NG~\citep{kofinas2024graph}; NFT~\citep{zhou2023neural}}
                , multiple_work]
            ]
        ]
        [Weight Space Generation~\S~\ref{section:wsg}, for tree={fill=blue!45, pretrain}
            [Generation Approaches , pretrain
                [{\tstyle{Hypernetworks}: e.g., \citet{ha2017hypernetworks}; HBN~\citep{krueger2017bayesian}; BbH~\citep{pawlowski2017implicit}; SMASH~\citep{brock2018smash}; GHN~\citep{zhang2018graph}; GHN-1/2~\citep{knyazev2021parameter}; \citet{Oswald2020Continual}; DHP~\citep{li2020dhp}; Sylph~\citep{yin2022sylph}; PudNet~\citep{wang2024learning}; ~\citet{littwin2019deep}; \citet{shamsian2021personalized}; MH~\citep{zhao2020meta}; HyperSeg~\citep{nirkin2021hyperseg}; HyperFormer~\citep{mahabadi2021parameter}; HyperDreamBooth~\citep{ruiz2024hyperdreambooth}; \citet{hedlin2025hypernet}; HyperStyle~\citep{alaluf2022hyperstyle}; HyperEditor~\citep{zhang2024hypereditor}; HyperInventor~\citep{dinh2022hyperinverter}; Polyhistor~\citep{liu2022polyhistor}; HyperTransformer~\citep{zhmoginov2022hypertransformer}; \citet{henning2018approximating}; HyperGAN~\citep{ratzlaff2019hypergan}}
                , pretrain_work]
                [{\tstyle{Generative Models}: e.g., HyperReps~\citep{schurholt2021self}; \citet{schurholt2022hyper}; SANE~\cite{schurholt2024towards}; ~\citet{meynent2025structure}; ~\citet{falk2025learning}; IGPG~\citep{soro2025instructionguided}; DnD~\citep{liang2025drag}; ICM-LoRA~\citep{shao2025context}; FLoWN~\citep{saragih2025flow}; NeuroGen~\citep{wang2025neurogen}; G.pt~\citep{peebles2022learning}; HyperDiffusion~\citep{erkocc2023hyperdiffusion}; GPD~\citep{yuan2024spatiotemporal}; BEND~\citep{wei2024bend}; p-diff~\citep{wang2024neural}; COND P-DIFF~\citep{jin2024conditional}; T2W~\citep{tian2025text2weight}; DiffLoRA~\citep{wu2024difflora}; RPG~\citep{wang2025recurrent}; ORAL~\citep{khan2025oral}; D2NWG~\citep{soro2025diffusionbased}, GeoSANE~\citep{hanna2026GeoSANE}}
                , pretrain_work]
            ]
            [Practical Implications , pretrain
                [{\tstyle{Conditional Weight Generation}: e.g., Sylph~\citep{yin2022sylph}; HyperLDM~\citep{nava2023meta}; T2W~\citep{tian2025text2weight}; \citet{li2025continual}; SANE~\citep{schurholt2024towards}}
                , pretrain_work]
                [{\tstyle{Real-Time Weight Optimization}: e.g., HyperSeg~\citep{nirkin2021hyperseg}; HyperStyle \citep{alaluf2022hyperstyle}; Wild-P-Diff \citep{zhang2025reimagining}}
                , pretrain_work]
                [{\tstyle{Model Merging}: e.g., \citet{zeng2025generative}; Deep-Align \citep{navon2024equivariant};  DWSNets \citep{navon2023equivariant}}
                , pretrain_work]
                [{\tstyle{Weight Initialization}: e.g., HyperDreamBooth~\citep{ruiz2024hyperdreambooth}; G.pt~\citep{peebles2022learning}; GHN-3~\citep{knyazev2023can} \citet{schurholt2022hyper}}
                , pretrain_work]
                [{\tstyle{Training Acceleration}: e.g., WNN~\citep{jang2023learning}; NiNo~\citep{knyazev2025accelerating}}
                , pretrain_work]
                [{\tstyle{Data Generation}: e.g., GASP~\citep{dupont2022generative}; NeuMeta~\citep{yang2024neural}; Pi-GAN~\citep{chan2021pi}; NeuMeta \citep{yang2024neural}}
                , pretrain_work]
            ]
        ]
    ]
\end{forest}
\end{adjustbox}
\caption{Taxonomy of foundational components and representative works for weight space learning.}
\label{fig:related_works}
\end{figure}

To address this gap, we propose a systematic perspective on WSL that consolidates its diverse developments under a cohesive structure. Specifically, we categorize existing works into three complementary dimensions as shown in Figure \ref{fig:framework}: \textit{Weight Space Understanding (WSU)}, which investigates the intrinsic structures and theoretical principles of weight space; \textit{Weight Space Representation (WSR)}, which learns embeddings or descriptors of weights for downstream tasks; and \textit{Weight Space Generation (WSG)}, which explores how new weights can be synthesized through hypernetworks or generative models. By establishing a unified terminology and taxonomy (Figure \ref{fig:related_works}), this survey aims to unify dispersed  studies from diverse communities into a coherent research paradigm, laying the conceptual foundation for future exploration in weight space learning. 

The remainder of this survey is structured as follows. In \textbf{Section~\ref{section:wsu}}, we investigate the intrinsic properties of weight space, including its symmetries, invariances, and underlying principles, forming the basis for understanding weight-space behavior. \textbf{Section~\ref{section:wsr}} reviews approaches for learning compact and informative representations of network weights, enabling tasks such as function prediction, model editing, and model retrieval. \textbf{Section~\ref{section:wsg}} discusses methods for generating or synthesizing neural network weights through auxiliary models, encompassing hypernetworks, and generative models. Beyond these core dimensions, \textbf{Section~\ref{section:application}} highlights representative application scenarios that leverage WSL in practical domains, and \textbf{Section~\ref{section:benchmark}} surveys available benchmarks and datasets that facilitate evaluation and comparison. Finally, we conclude by summarizing emerging trends, open challenges, and potential directions for future research in \textbf{Section~\ref{section:conclusion}}. We hope this structured overview not only consolidates current progress but also provides a conceptual roadmap for exploring the broader potential of Weight Space Learning.

\newpage

\section{Weight Space Understanding} \label{section:wsu}



Weight Space Understanding (WSU) aims to characterize the intrinsic structure of the neural network weight space, independent of any specific dataset or training objective. Rather than viewing model parameters as unstructured vectors governed by a flat metric, WSU investigates the underlying structures, e.g., symmetries, redundancies, and manifold topologies, that govern how different weight configurations correspond to equivalent model behaviors. This perspective provides fundamental insights into why deep networks generalize, how optimization landscapes are shaped, and what structural properties emerge from overparameterization. 

At its core, WSU reveals that the weight space of modern neural networks is a highly organized domain with rich geometric and topological structures. These structures arise from architectural symmetries, parameter sharing, and the functional equivalence between different permutations of neurons or filters. Understanding these properties is essential not only for theoretical insights, such as explaining model identifiability and loss landscape connectivity, but also for developing practical tools that exploit weight space regularities for more efficient learning. Building on these insights, research in WSU has inspired several practical directions. Analyses of symmetry and redundancy have led to compression methods that remove functionally redundant parameters, optimization strategies that navigate symmetry-invariant subspaces, and even data augmentation schemes that perturb weights while preserving model semantics. By bridging theoretical structure with empirical utility, WSU provides a principled foundation for rethinking how we design, train, and interpret neural networks.

In the following subsections, we first formalize symmetry as a structural property in parameter space, and focus on its induced functional relations (invariance and equivariance). We then discuss representative applications that leverage these insights for model compression, optimization guidance, and augmentation. Finally, we summarize open challenges and future research directions toward a unified understanding of weight space geometry and its role in deep learning.

\begin{figure}[!b]
    \centering
    \includegraphics[width=0.8\linewidth, trim={0cm 1cm 0cm 0.8cm}, clip, scale=0.5]
    {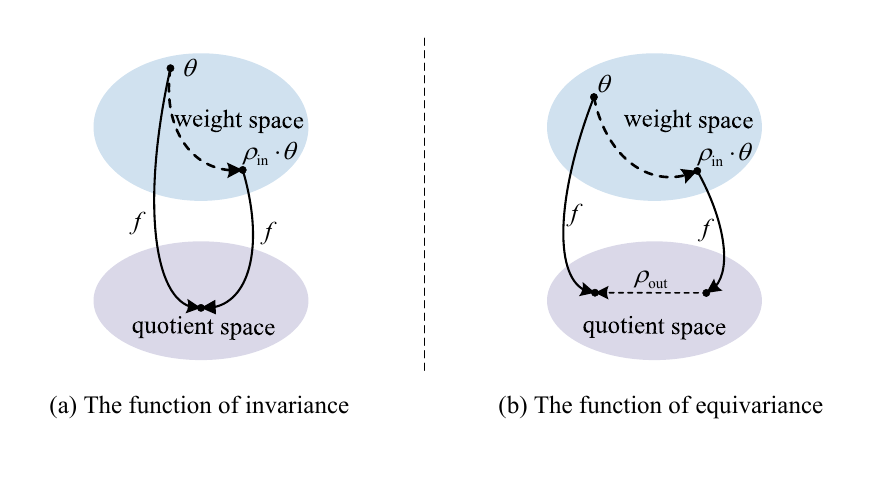}
    \caption{\textbf{Weight space symmetry and its induced functional manifestations.} (a) illustrates the invariance under group action: applying a transformation $\rho_{in}$ to weights $\theta$ in weight space does not alter the induced function in quotient space\protect\footnotemark. (b) illustrates the equivariance under group action: transforming weights $\theta$ by $\rho_{in}$ in weight space induces a corresponding transformation $\rho_{out}$ in quotient space.}
    \label{fig:symmetry}
\end{figure}
    \footnotetext{The quotient space is defined as the set of equivalence classes of weights that parameterize the same function. Mathematically, let $G$ be the symmetry group acting on the weight space $\mathcal{W}$. The quotient space is defined as the orbit space $\mathcal{W}/G \coloneqq \{ [\theta] \mid \theta \in \mathcal{W} \}$, where the equivalence class  $[\theta] = \{ g \cdot \theta \mid g \in G \}$ contains all weight configurations reachable via symmetry transformations.}

\subsection{Structural Foundations}

\begin{myboxi}[Takeaways]
\begin{itemize}
    \item Under parameter-space symmetries, \textbf{invariance} exposes the redundancy of neural parameterizations: many weight configurations encode the same function. Recognizing these classes clarifies optimization degeneracy and motivates symmetry-invariant algorithms and representations.  
    \item Under parameter-space symmetries, \textbf{equivariance} captures structured transformations in weight space that induce predictable functional transformations. It provides a geometric foundation for designing models and meta-models that generalize across architectures through predictable parameter–function mappings.  
\end{itemize}
\end{myboxi}

A central perspective in understanding weight space lies in identifying its inherent \textit{symmetries} as group actions on parameters, independent of data or loss functions. At the functional level, these parameter-space symmetries induce two canonical relations: \textit{invariance} and \textit{equivariance}, as illustrated in Figure \ref{fig:symmetry}. This structural view helps explain functional redundancy, optimization degeneracy, and generalization patterns in overparameterized networks. Symmetry is a central enabler for learning in weight space: accounting for it is crucial for downstream tasks ranging from model merging and model-level representation learning to weight generation.

To build intuition, we contrast the two induced functional relations. Invariance means that parameter transformations leave the realized function unchanged, whereas equivariance means that parameter transformations induce predictable change in function/output space. This functional distinction provides a practical lens for organizing structural regularities in neural networks, from permutation and scaling effects to structured head-wise transformations.
We summarize representative examples in Table \ref{tab:symmetry example}, and then detail invariance and equivariance in the following subsections.

\subsubsection{Functional Invariance}

Invariance refers to transformations of the weight parameters that leave the represented function completely unchanged.  
Intuitively, these transformations map one parameterization of a network to another that produces exactly the same input--output behavior.  
Such invariances often arise from redundancies in neural architectures, reflecting the fact that many weight configurations correspond to the same underlying function.
Formally, let $\theta$ denote a set of model weights and $\rho_{in}: \Theta \rightarrow \Theta$ be a transformation acting on the weight space.  
If for all inputs $x$ we have
\begin{equation}
    f(\rho_{in}(\theta); x) = f(\theta; x),
\end{equation}
then $\rho_{in}$ defines an invariance transformation.  
This equivalence partitions the weight space into functional equivalence classes, where each class represents a single neural function realized by multiple parameter settings.

In practice, several symmetry transformations induce functioanl invariance in neural networks: 
\begin{itemize}
    \item \textbf{Neuron or filter permutations.}  
    Swapping neurons within a hidden layer (and applying the inverse swap in the next layer) yields an identical function.
    This is perhaps the most studied form of parameter redundancy, observed across Multilayer Perceptrons (MLPs), Convolutional Neural Networks (CNNs), and self-attention.
    \item \textbf{Positive scaling invariance.}  
    Layers equipped with Batch Normalization scale the incoming weights or activations by a positive constant does not change the normalized output.  
    Batch Normalization amplifies this property by absorbing scale parameters into normalization statistics.  
    \item \textbf{Bias or logit translation.}  
    Functions such as Softmax are invariant to adding a constant to all logits, implying that certain shifts in weights or biases do not affect the output distribution.
\end{itemize}

From a theoretical viewpoint, these invariances reveal why the optimization landscape of neural networks is highly degenerate: 
many distinct minima correspond to the same function~\citep{brea2019weight}, producing flat valleys in weight space.  
Recognizing these symmetries allows one to design optimization algorithms that avoid redundant search directions and to construct parameter-invariant representations of networks for meta-analysis~\citep{zhao2025symmetry, herrmann2024learning}. \citet{sourek2021lossless} exploit permutation invariance in GNNs for lossless compression, while \citet{entezari2022the} demonstrate that accounting for permutation invariance places SGD solutions in the same basin, enabling linear interpolation without barriers. Extending these insights, Git Re-Basin~\citep{ainsworth2023git} develops practical algorithms to align permutations and merge independently trained models via linear mode connectivity. \citet{shamsian2023data} extend the linear mode connectivity to design weight space mixup, achieving augmentation in weight space. Specifically, weight space mixup operates by interpolating the weights $w$ and biases $b$ of two models (typically after alignment), creating new parameter sets that reside on the line segment connecting them in weight space. \citet{theus2025generalized} extend the linear mode connectivity exploration to the transformer-based architecture.

\noindent\textbf{Why it matters.}
Understanding invariance in weight space opens up new opportunities for learning algorithms. 
Because multiple configurations correspond to the same function, optimization does not search for a single global minimum but for an entire equivalence class of solutions~\citep{brea2019weight,simsek2021geometry}. 
Recognizing this redundancy explains the success of model compression~\citep{sourek2021lossless} and weight space augmentation~\citep{shamsian2024improved}.
Moreover, explicitly modeling invariance enables the development of optimization algorithms that operate on functionally meaningful manifolds rather than raw parameters, enabling linear mode connectivity between independently trained models~\citep{entezari2022the,zhao2025understanding} for improved model merging~\citep{ainsworth2023git}, and principled weight space interpolation~\citep{shamsian2023data,shamsian2024improved}.
In short, invariance provides the theoretical foundation for understanding when different networks truly represent the same function and how this redundancy can be exploited for efficient and robust learning.

\begin{table}[!t]
    \centering
    \caption{\textbf{Illustrative examples of functional invariance and equivariance induced by weight space symmetry.}
    Under weight space symmetry transformations, invariance means that the realized function remains unchanged, while equivariance means that the realized function changes in a predictable, structured manner. These relations explain that many different parameter configurations correspond to the same or systematically related functions. See \citet{zhao2025symmetry} for a theoretical discussion.}
    \label{tab:symmetry example}
    \resizebox{\linewidth}{!}{
        \begin{tabular}{llll}
            \toprule
            \textbf{Functional Relation} & \textbf{Transformation Example} & \textbf{Functional Effect}                                          \\ \midrule
                                   & Neuron permutation              & Functionally identical network under reordered hidden units         \\ \cmidrule{2-3}
            \textbf{Invariance}    & Positive scaling                & Scaling is neutralized in batch normalization \\ \cmidrule{2-3}
                                   & Bias translation in Softmax     & Constant shift of logits leaves probabilities unchanged             \\ \midrule
                                   & Orthogonal rotation             & Rotating neuron basis with output changing accordingly              \\ \cmidrule{2-3}
            \textbf{Equivariance}  & Sign flipping                       & Output changes predictably under the corresponding sign transformation       \\ \cmidrule{2-3}
                                   & Attention-head permutation      & Reindexing heads yields equivariant attention structure              \\ \bottomrule
        \end{tabular}
    }
\end{table}


\subsubsection{Functional Equivariance}

While invariance describes transformations that leave a network’s function unchanged, \emph{equivariance} captures structured transformations where changes in parameters induce predictable changes in model behavior.  
Equivariance formalizes the idea that if weights are transformed in a particular way, the output transforms in a corresponding manner, mirroring the geometric equivariances exploited in convolutional or group-equivariant networks.
Mathematically, let $\rho_{in}: \Theta \rightarrow \Theta$ be a transformation on weights and $\rho_{out}: \mathcal{Y} \rightarrow \mathcal{Y}$ be a corresponding transformation on the output space.  
A function $f$ is said to be \emph{equivariant} with respect to $(\rho_{in}, \rho_{out})$ if
\begin{equation}
    f(\rho_{in}(\theta); x) = \rho_{out}(f(\theta; x)).
\end{equation}
Unlike invariance, here the function is not identical after transformation, but its change follows a well-defined pattern.  
This property highlights how weight space symmetries can encode the same structural regularities that are imposed explicitly in architectures such as CNNs or transformers.

Examples of weight space symmetry that induce functional equivariance include:
\begin{itemize}
    \item \textbf{Orthogonal or rotation symmetries.}  
    For networks with radial activations, rotating neurons within a layer makes the output rotate accordingly.
    \item \textbf{Sign flipping.} Flipping the sign of hidden units via parameter transformations causes the output representation to flip its sign in the same way, preserving structural consistency.
    \item \textbf{Headwise transformations in attention.}  
    Permuting or linearly mixing attention heads leads to corresponding transformations in attention maps, maintaining equivalence up to reindexing.
\end{itemize}

Functional equivariance induced by weight space symmetries offers a geometric perspective on how architectures enforce functional consistency through structured transformations.  
It provides a bridge between group-theoretic learning principles and practical neural architecture design: from the study of weight permutations and orthogonal rotations~\citep{ganev2021universal,theus2025generalized} to recent graph-based models that explicitly encode parameter dependencies as symmetry-preserving structures~\citep{kofinas2024graph,lim2024graph}. Deep Weight-Space Networks (DWSNets)~\citep{navon2023equivariant} formalize neuron-permutation symmetry in MLP weight spaces and constructs DWS layers as a complete basis of affine equivariant maps using pooling, broadcasting, and linear blocks. Meanwhile, Neural Functional Networks (NFNs)~\citep{zhou2023permutation} enforce permutation equivariance via parameter-sharing and extends it to CNNs. Later, Universal Neural Functionals (UNFs)~\citep{zhou2024universal} generalize permutation-equivariant modeling to arbitrary architectures, including Recurrent Neural Networks (RNNs) and Transformers. More recently, ~\citet{kofinas2024graph} utilize GNN to achieve permutation equivariance. EQUIGEN~\citep{huang2025few} encodes the weight space into an equivariant latent space for generation tasks.

\noindent\textbf{Why it matters.}
Equivariance reveals that neural networks are not only functionally redundant (arise from invariance) but also structurally organized: weight transformations induce predictable, lawful changes in model behavior~\citep{navon2023equivariant}. 
By exploiting this structure, one can design meta-models~\citep{zhou2024universal} and generative frameworks~\citep{huang2025few} that reason directly over families of networks rather than individual instances. 
Equivariance thus provides a blueprint for building weight space representations that generalize across architectures or tasks, enabling model editing~\citep{navon2023equivariant,zhou2023permutation,kofinas2024graph}, and even generative synthesis of new networks~\citep{huang2025few}. 
In essence, it views neural network weights as points on a manifold, which enables mathematical tools from differential geometry and representation theory to be applied to the analysis and improvement of neural network behavior.

\subsection{Technique-Level Use Cases}

\begin{myboxi}[Takeaways]
\begin{itemize}
    \item Understanding weight space symmetry enables principled compression and optimization by exploiting invariance and equivariance among parameter configurations.

    \item Viewing models as points on symmetry orbits motivates new augmentation and learning strategies that enhance efficiency and generalization.
\end{itemize}
\end{myboxi}

By treating weight configurations through symmetry orbits and equivalence classes rather than isolated points, WSU translates theoretical insights into new strategies for model compression, optimization, and data augmentation. 
These applications demonstrate how symmetry-aware analysis bridges fundamental theory with practical improvements in efficiency, robustness, and generalization.

\subsubsection{Model Compression}

One of the most direct consequences of WSU lies in revealing redundancy induced by overparameterization. 
Symmetries such as neuron permutation, scaling, or orthogonal transformation imply that many parameters encode equivalent functions, suggesting that large models often inhabit lower-dimensional manifolds in weight space. 
By identifying and exploiting these equivalence classes, compression methods can remove functionally redundant parameters while preserving performance. Early work formalized this idea by merging symmetric computation graphs~\citep{sourek2021lossless} or exploiting orthogonal invariances in radial architectures through factorized representations~\citep{ganev2021universal}. 
Later approaches extended this principle to modern convolutional and attention-based architectures, where symmetry-guided pruning and low-rank decomposition exploit structural regularities for efficient deployment~\citep{martinez2021permute,abronin2024tqcompressor,li2024merge}. Beyond efficiency, symmetry-aware compression provides interpretability: the identification of equivalent subnetworks reveals how architectural design induces functional redundancy, offering a lens to study overparameterization beyond empirical scaling laws.

\subsubsection{Model Optimization}

Symmetry profoundly influences optimization dynamics by shaping the loss landscape into connected manifolds of equivalent minima. 
Rather than discrete points, optima form \emph{orbits} under symmetry transformations, explaining why stochastic optimizers often traverse wide valleys instead of sharp minima.  
This understanding motivates optimization strategies that either exploit or explicitly remove these redundant directions.
Orbit-based optimization explores symmetry-preserving trajectories to improve convergence and escape degenerate saddles. ~\citet{zhao2022symmetry} propose Symmetry Teleportation, which allows the optimizer to jump to a different position on the same orbit that has more favorable local geometry. ~\citet{saul2023weightbalancing} further investigates weight-balancing flows, demonstrating that explicit rebalancing of weight magnitudes between layers can fix ill-conditioned gradients. 
Alternatively, symmetry-invariant optimization constructs algorithms whose updates are insensitive to specific transformations, operating directly in the quotient space of functionally distinct solutions. 
Path-SGD~\citep{neyshabur2015path} performs steep descent with respect to a path-regularized metric rather than the Euclidean norm, effectively ensuring update steps are invariant to rescaling. Similarly, $\mathcal{G}$-SGD~\citep{meng2018gsgd} optimizes weights directly on the positively scale-invariant manifold, while \citet{yi2022accelerating} and \citet{huang2020projection} develop projection-based methods that remove the radial components of gradients, forcing updates to remain on the tangent space of the unit sphere.
Understanding symmetry also sheds light on phenomena like mode connectivity and weight interpolation~\citep{tatro2020optimizing}, explaining why independently trained models can often be connected through low-loss paths in weight space.  
Thus, WSU reframes optimization not as searching for a point but as navigating a manifold of equivalent solutions.

\subsubsection{Weight Space Augmentation}

Symmetry understanding also inspires new data augmentation paradigms, which are performed not in input or feature space but directly in the weight space itself.  
Because symmetric transformations yield functionally equivalent models, these transformations can generate diverse yet semantically consistent variations of a network, enriching model ensembles and regularizing training.
Recent studies have operationalized this idea mainly in implicit neural representations (INRs), where data scarcity amplifies overfitting risks, while earlier model-zoo work also reported gains from symmetry-aware permutation augmentations~\citep{schurholt2021self}..  
Weight space MixUp~\citep{shamsian2023data} and its alignment-based variants~\citep{shamsian2024improved} interpolate between weight configurations to produce smoother function families. In addition, \citet{navon2024equivariant} leverage permutation symmetry to learn equivariant mappings between networks, enabling efficient augmentation strategies for INR.
EQUIGEN~\citep{huang2025few} extends this notion by applying permutation-based equivariant transformations within weight space to augment INR datasets and improve few-shot generation.  
These techniques highlight a broader insight: we can treat symmetry as a roadmap to navigate the weight space, enabling the controlled representation and even generation of functionally meaningful variations. 
In doing so, WSU bridges the gap between theoretical geometry and practical generalization in modern neural networks.

\subsection{Discussion and Perspective}

\begin{myboxi}[Takeaways]
\begin{itemize}
    \item 
    WSU reveals weight space as a geometric domain shaped by symmetry, offering a unified lens for understanding redundancy, optimization, and generalization.  

    \item 
    Future progress depends on turning WSU’s descriptive insights into measurable, predictive principles that connect weight space geometry with real model behavior.  
\end{itemize}
\end{myboxi}

Weight space understanding provides the conceptual backbone for treating neural parameters as structured, learnable objects. 
By uncovering symmetries and geometric regularities, it reframes how we interpret redundancy~\citep{sourek2021lossless}, optimization~\citep{zhao2025symmetry}, and generalization~\citep{entezari2022the} in modern networks. 
Yet despite its theoretical development, WSU remains an emerging and evolving lens with both distinctive advantages and persistent challenges.

\noindent\textbf{Strengths and Contributions.}
The symmetry-based view of WSU has brought coherence to several previously disconnected observations in deep learning. 
It explains why different parameterizations can encode the same function~\citep{shen2024exploring}, why optimization landscapes exhibit connected minima~\citep{brea2019weight}, and why large models can be merged or interpolated without loss~\citep{entezari2022the,ainsworth2023git}.  
These insights have enabled a class of algorithms, ranging from quotient-space optimization to symmetry-aware meta-models, that explicitly exploit the structure of weight space rather than ignoring it~\citep{navon2023equivariant,zhou2023permutation,zhou2024universal,zhou2023neural,lim2024graph,kofinas2024graph,kalogeropoulos2024scale,kiwitt2025symmetries}.  
Furthermore, by recognizing weight configurations as data, WSU opens the door for new modalities of learning, such as weight space augmentation and editing, thereby forming a theoretical foundation for the next sections on weight space representation and generation.

\noindent\textbf{Limitations and Open Issues.}
Despite rapid progress, WSU still faces conceptual and practical barriers.  
Many symmetries remain only partially characterized: real-world architectures such as transformers introduce complex parameter couplings that break simple group structures, making analytical treatment difficult~\citep{theus2025generalized,zhang2025beyond,xu2024permutation,lavie2024transformers,tran2025equivariant}.  
Bridging theory and practice is equally challenging: symmetry principles derived in idealized settings often fail to transfer to large-scale models with nontrivial normalization, residual connections, or mixed-precision constraints~\citep{zhang2022set,zhang2025symmetry,imani2024exploring}.  
Furthermore, computational scalability poses a severe bottleneck. Operating within the quotient space requires factoring out symmetries, a process that often reduces to solving large-scale combinatorial optimization problems (e.g., graph matching) known to be NP-hard or computationally intensive~\citep{ainsworth2023git}.
In addition, the diversity of architectures demands meta-models that can reason over heterogeneous weight spaces while remaining permutation symmetry, a goal that remains largely unsolved~\citep{zhou2024universal}.  
Beyond symmetry, weight space also exhibits other mathematical structures such as metric geometry and category-theoretic formulations, which provide alternative lenses for understanding optimization and compositionality~\citep{amari1998natural,bronstein2021geometric}.  
Integrating these perspectives with symmetry-based approaches remains an open challenge and a promising direction for future research toward a unified theory of weight space.

\noindent\textbf{Outlook.}
Looking ahead, several open directions highlight the limits of our current understanding of weight space structure.  
First, the geometry of practical networks remains only partially observable: while low-dimensional symmetries explain redundancy in small models, the weight manifolds of large-scale architectures exhibit far richer, data-dependent organization that is yet to be formally characterized.  
Second, most symmetry analyses remain theoretical; translating these insights into measurable quantities, such as curvature, connectivity, or spectral statistics of pretrained models, will be crucial for grounding WSU in empirical evidence.  
Finally, the next frontier lies in transforming WSU from a descriptive framework into a predictive one: understanding how geometric properties of the weight space can affect generalization, robustness, or adaptability.  
Progress in these directions will determine whether WSU becomes merely a diagnostic tool or a genuine theoretical substrate for modern machine learning.

\section{Weight Space Representation} \label{section:wsr}

Once the intrinsic structure of weight space is understood, a natural next question arises: \emph{can we represent neural weights in a way that captures model semantics, behavior, or function?}  
Weight Space Representation (WSR) addresses this question by learning compact and transferable embeddings of network parameters.  
Unlike traditional representation learning, WSR focuses on encoding the learners of data, allowing reasoning, retrieval, or prediction directly in the model space. 
Formally, let a neural network be parameterized by $\theta \in \Theta$, where $\Theta$ denotes the weight space, and let $\phi: \Theta \rightarrow \mathbb{R}^d$ denote a representation function that maps the weights to a low-dimensional embedding:
\begin{equation}
    z = \phi(\theta),
\end{equation}
where $z$ captures structural or functional properties of the model.  
The learned embedding can then support downstream objectives such as performance prediction, similarity retrieval, or model editing:
\begin{equation}
    y = g(z), \quad \text{with} \quad g: \mathbb{R}^d \rightarrow \mathcal{Y},
\end{equation}
where $\mathcal{Y}$ denotes the task-specific target space (e.g., accuracy regression, or a discrete set for property classification).
Here, $\phi$ and $g$ together form a meta-model that learns from networks instead of data, providing a bridge between model-level understanding and functional inference.

Two major paradigms have emerged for implementing WSR.  
Model-based approaches learn $\phi(\cdot)$ explicitly from weight statistics~\citep{unterthiner2020predicting,martin2021predicting} or structure~\citep{schurholt2021self, zhou2023permutation,lim2024graph,kalogeropoulos2024scale, schurholt2024towards}. 
In contrast, model-free approaches infer embeddings implicitly by probing the functional behavior of neural networks, constructing embeddings from surrogate responses without direct access to raw weights~\citep{kahana2025deep,herrmann2024learning}.  
These two views offer complementary strengths: the former captures explicit structural regularities, while the latter encodes functional equivalence and generalization patterns.
Despite rapid progress, WSR faces several fundamental challenges.  
First, \textit{modality heterogeneity}: different architectures exhibit different parameter organizations and scaling behaviors, making unified representations difficult to learn~\citep{lim2024graph,kofinas2024graph,zhou2024universal}.  
Second, \textit{high dimensionality}: modern architectures contain millions of weights, and the resulting overparameterization poses challenges to learning compact and semantically consistent representations~\citep{zhao2025symmetry}.  
Bridging these challenges requires efficient structure-aware architectures~\citep{navon2023equivariant,zhou2023permutation} and regularization strategies~\citep{ashkenazi2023nern,andreis2024set} that respect weight-space structures while maintaining scalability.

In the following sections, we survey the key methodologies that enable WSR, ranging from explicit statistical features and learning-based formulations to probing-driven approaches, and review how these representations power practical tasks such as model retrieval, property prediction, and implicit neural representation.  
Through this lens, WSR emerges as a crucial link between theoretical understanding of weight space structure and data-independent model analysis.

\subsection{Representation Approaches}

\begin{myboxi}[Takeaways]
\begin{itemize}
    \item WSR evolves from handcrafted symmetry modeling toward automatic discovery of functional regularities. 
    \item Model-based and model-free paradigms offer complementary perspectives: the former captures internal geometry of weights, while the latter embeds functional behavior. 
\end{itemize}
\end{myboxi}

\begin{figure}[t!]
    \centering
    \includegraphics[width=0.9\linewidth, trim={0cm 0.6cm 0cm 0cm}, clip, scale=0.9]
    {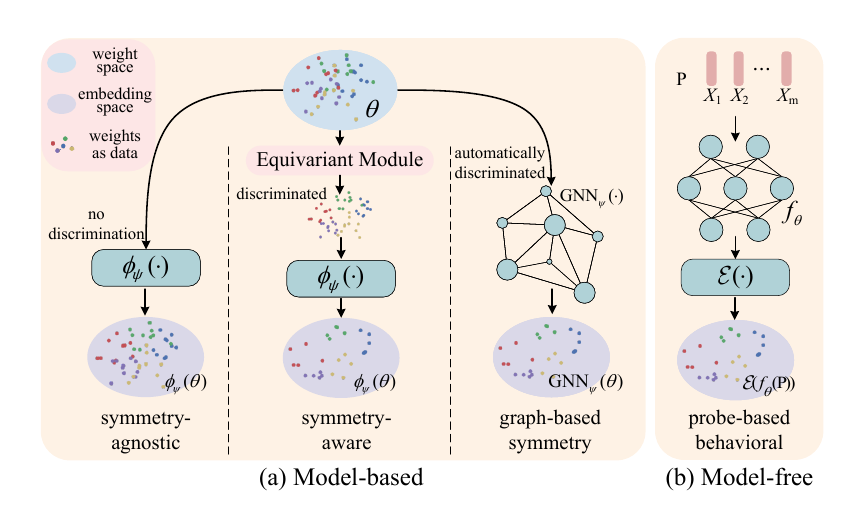}
    \caption{\textbf{Weight Space Representation Methods: Overview and Comparison.} (a) Model-based approaches. All take raw weight space $\theta$ as input but differ in how they handle symmetry. Symmetry-agnostic methods learn representations directly from weight distributions without explicitly encoding neuron permutation symmetry. Symmetry-aware methods incorporate dedicated equivariant modules to respect permutation invariance. Graph-based symmetry approaches leverage GNNs to automatically capture structural symmetries among neurons. (b) Model-free approaches. Probe-based behavioral learning infers representations from functional responses to designed probes, enabling embeddings that encode behavioral properties and functional symmetries of the underlying network.}
    \label{fig:representation}
\end{figure}

From a methodological viewpoint, learning in weight space centers on designing the representation function 
$\phi: \Theta \rightarrow \mathbb{R}^d$ that maps network parameters $\theta$ into an informative, low-dimensional embedding as shown in Figure \ref{fig:representation}.  
Different assumptions about how to define and learn $\phi$ give rise to distinct methodological families. 
We provide a conceptual taxonomy of WSR:  
model-based approaches explicitly encode the values and connectivity of weights, while model-free approaches treat the network as a function and learn from its predictive outputs on reference data.  

\subsubsection{Model-based Representation}

Model-based approaches use a model to learn explicit mappings from network parameters to their embeddings by directly operating on the weights themselves. 
These methods treat the weight tensors as structured signals and design auxiliary models to encode them, typically under varying degrees of symmetry awareness. 
Conceptually, they can be categorized into three evolutionary stages: (\textit{i}) symmetry-agnostic encoders, (\textit{ii}) symmetry-aware architectures with handcrafted inductive biases, and (\textit{iii}) graph-based frameworks that automatically capture structural symmetries. 

\noindent\textbf{Symmetry-agnostic encoders.}
Early studies approached weight representation as a statistical or regression problem, using classical learning models to fit the distribution of parameters. 
For instance, \citet{martin2021predicting} analyzed spectral norms and singular value statistics of weight matrices to predict model quality, while \citet{eilertsen2020classifying} and \citet{unterthiner2020predicting} employed CNNs, MLPs, and boosted trees to classify or rank networks based solely on their weights. 
Similarly, \citet{schurholt2021self} proposed self-supervised encoders for model zoos \citep{unterthiner2020predicting, schurholt2022model}, leveraging transformer-based hyper-representations. 
Formally, these approaches learn a deterministic mapping 
\begin{equation}
    z = \phi_{\psi}(\theta), \quad \text{where $\psi$ denotes the learnable parameters of the encoder},
\end{equation}
without imposing constraints on $\phi_{\psi}$ to respect architectural symmetries.
While simple and flexible, such models often fail to generalize across architectures, as they ignore the inherent permutation and scaling invariances of neural networks.

\noindent\textbf{Symmetry-aware architectures.}
Building upon these limitations, recent works incorporate explicit geometric priors into $\phi_{\psi}$ to ensure that equivalent parameterizations yield consistent representations.
Representative examples include DWSNets~\citep{navon2023equivariant} and NFN~\citep{zhou2023permutation}, which construct permutation-equivariant mappings through layerwise decomposition and shared weights over symmetry group orbits. 
DWSNets achieve permutation invariance by hcaracterzing the space of all linear equivaraint layers for the permutation symmetries of MLP weight spaces. Interstingly, they linear layers combine layers inspired by DeepSets~\citep{zaheer2017deep} and its genralizations to multiple sets ~\citep{hartford2018deep}, ensuring that neuron reordering leaves the embedding unchanged. 
NFN derived an equivalent architecture with a slightly improved scaling due to the use of smart positional encodings for input/output neurons. They also included experiments on  CNNs. 
Further extensions, such as Neural functional transformers (NFT)~\citep{zhou2023neural}, integrate attention mechanisms to model inter-layer dependencies, maintaining equivariance when adjacent layers are jointly permuted. MAGEP-NFN~\citep{vo2025equivariant} and Monomial-NFN~\citep{tran2024monomial} broaden NFN’s symmetry group to include scaling transformations, with the latter further accommodating sign-flipping symmetries. Universal Neural Functionals (UNFs)~\citep{zhou2024universal} propose maximally expressive equivariant linear layers and for a broader range of architectures, including RNNs and the transformers. ~\citet{zhang2025beyond} investigate rotational symmetries in transformer architectures, revealing new structural regularities beyond permutation. ~\citet{sverdlov2025revisiting} decomposes the input weight space of MLPs into irreducible representations, simplifying the design of symmetry-respecting neural architectures.
Other studies like Neural Representation for Neural Networks (NeRN)~\citep{ashkenazi2023nern} and Set-based Neural network Encoder (SNE)~\citep{andreis2024set} introduce regularization-based strategies to approximate invariance, e.g., Logit Invariance Regularization~\citep{moskalev2023genuine}, enabling smoother weight representations under local symmetry perturbations.
Overall, these models explicitly encode known symmetries, but their handcrafted nature limits scalability and adaptability to heterogeneous architectures.

\noindent\textbf{Graph-based weight space architectures.}
To overcome the rigidity of manually designed architectures, recent work reinterprets neural networks as structured graphs and employs Graph Neural Networks (GNNs)~\citep{wang2024gft,wang2025git,wang2025beyond,wang2025scalable} to learn over them. 
This formulation provides a natural mechanism for permutation invariance, since GNN message passing is inherently equivariant to node reordering. 
In this view, each network $\mathcal{N}$ is modeled as a directed acyclic graph $\mathcal{G}=(\mathcal{V}, \mathcal{E})$ with node features (biases) and edge features (weights), and the encoder learns
\begin{equation}
    z = \text{GNN}_{\psi}(\mathcal{G}),
\end{equation}
automatically discovering structural regularities.
Notable examples include Neural Graph (NG)~\citep{kofinas2024graph} and Graph Metanetworks (GMNs)~\citep{lim2024graph}, which represent layers or modules as subgraphs and propagate information along computational dependencies.  
Scale-GMN~\citep{kalogeropoulos2024scale} further incorporates scaling symmetry to handle activation-level transformations, complementing prior permutation-focused designs. \citet{ballerini2025weight} extend this paradigm to Neural Radiance Fields (NeRFs), demonstrating that GMNs can produce architecture-agnostic embeddings across diverse parameterizations.
By modeling weight space as a graph manifold, these approaches unify structure-aware encoding with flexibility, achieving symmetry handling without manual specification.

In summary, model-based methods have evolved from direct parameter regression to symmetry-aware and finally graph-based frameworks that internalize architectural invariances. 
This progression reflects a broader trend in WSR: shifting from handcrafted formulations toward automatic encoding of the geometric principles that govern weight space.

\subsubsection{Model-free Representation}

While model-based approaches encode the internal structure of weight tensors, model-free methods represent a neural network through its externally observable behavior.  
Rather than analyzing parameters directly, they infer representations by probing how a model transforms inputs into outputs, thereby sidestepping challenges such as permutation symmetry, scaling invariance, and architecture-specific parameterization.
Formally, a model-free representation defines an implicit embedding function 
\begin{equation}
    z = \phi_f(f_\theta), \quad \text{where } f_\theta: \mathcal{X} \rightarrow \mathcal{Y},
\end{equation}
mapping a network’s functional behavior to a latent descriptor $z$.  
Since $f_\theta$ is accessible only through queries, $\phi_f$ is typically estimated by evaluating the model’s responses to a finite set of probes $\mathcal{P} = \{x_i\}_{i=1}^m$, yielding
\begin{equation}
    z = \mathcal{E}\big(\{f_\theta(x_i)\}_{x_i \in \mathcal{P}}\big),
\end{equation}
where $\mathcal{E}(\cdot)$ aggregates the model’s responses into a fixed-length behavioral signature.  
This framework turns the abstract weight space into an empirical function space, embedding each network according to how it acts rather than how it is built.

\paragraph{Probe-based behavioral learning.}
A dominant research line in this paradigm is probing-based representation, 
which seeks to characterize a model through its reaction to carefully designed or learned stimuli.  
Early methods used random or handcrafted probes to estimate the representational similarity between models.
For instance, ProbeLog~\citep{kahana2025can} exemplifies a static probing paradigm at the logit level, where a fixed, ordered set of probe inputs is fed to candidate models and each output dimension (logit) is represented by its response signature. 
More recent approaches learn probes jointly with the embedding function to maximize discriminability and transferability.  
ProbeGen~\citep{kahana2025deep} constructs deep linear generators that produce structured probes aligned with the functional manifold of neural networks, improving generalization across architectures.  
\citet{kofinas2024graph} integrate probing into GNN-based frameworks, enriching node features with probe-induced activations.  
ProbeX~\citep{horwitz2025learning} extends the idea to hidden layers using a mixture-of-experts design, capturing diverse internal behaviors within the same probing protocol.  
In the temporal domain, \citet{herrmann2024learning} introduce interactive probing for RNNs, where query inputs adapt to model responses, achieving exponential gains in efficiency over static probes for complex sequential reasoning tasks. 

These representation methods provide a complementary perspective to model-based weight encoders: they are \textit{architecture-agnostic}, naturally invariant to parameter symmetries, and readily extensible to black-box settings where weights are inaccessible.  

\subsection{Technique-Level Use Cases}

\begin{myboxi}[Takeaways]
\begin{itemize}
    \item WSR enables model-level reasoning: once networks are embedded as latent vectors, their behaviors can be predicted, retrieved, or modified directly in embedding space.  
    \item By decoupling analysis from data, WSR transforms weight space into a functional manifold where structure and behavior align, enabling scalable model evaluation and reuse.  
\end{itemize}
\end{myboxi}

By embedding neural network parameters into a compact latent representation $z = \phi(\theta)$, 
Weight space representation (WSR) enables reasoning directly over models rather than data.  
Such embeddings capture functional and structural information about networks, 
allowing downstream operations such as predicting model behavior, retrieving suitable pretrained models, 
and editing network functions, all without accessing the original training datasets.

\subsubsection{Behavior Prediction}

The most direct application of WSR is to predict a model’s behavior from its weight embedding.  
Given a learned representation $z = \phi(\theta)$, one can train a predictor $\psi(z)$ to estimate performance metrics, hyperparameters, or generalization properties.  
This paradigm transforms model assessment into a regression problem in the embedding space:
\begin{equation}
    \hat{y} = \psi(\phi(\theta)), \quad \text{where } y \text{ denotes a target property of the model.}
\end{equation}
Early works such as DCM~\citep{eilertsen2020classifying} and \citet{unterthiner2020predicting} demonstrated that network accuracy and hyperparameters 
can be effectively predicted from weight statistics or meta-learned representations~\citep{lim2024graph,kofinas2024graph,kahana2025deep}.  
Subsequent studies extended this idea to large model zoos~\citep{schurholt2022model,honegger2023sparsified}, and~\citet{schurholt2021self,schurholt2024towards} built self-supervised hyper-representations 
to forecast generalization gaps and learning dynamics, 
while~\citet{falk2025learning} applied unsupervised encoding to undocumented Hugging Face models for accuracy prediction.  
These results suggest that weight embeddings encode rich predictive signals about model capacity, optimization history, and task fit, 
providing a foundation for large-scale automated model analysis.

\subsubsection{Model Retrieval}

The proliferation of pretrained models in public repositories such as Hugging Face has shifted the challenge from model creation to model selection.  
WSR offers a principled solution: by embedding each model’s parameters into a shared latent space, 
one can measure cross-model similarity and retrieve networks that are functionally aligned with a given query.  
Formally, given a set of models $\mathcal{M} = \{\theta_i\}$ and a query embedding $z_q$, retrieval is achieved by selecting the candidate from the discrete set $\mathcal{M}$ that minimizes the distance in the embedding space:
\begin{equation}
    \text{Retrieve:} \quad \theta^* = \underset{\theta_i \in \mathcal{M}}{\arg\min} \, \| \phi(\theta_i) - z_q \|.
\end{equation}
This approach enables data-independent model matching, selecting networks by what they have learned, not merely by their metadata.  
Recent work operationalizes this idea at scale:  
ProbeX~\citep{horwitz2025learning} aligns models via probe-induced functional embeddings, 
ProbeLog~\citep{kahana2025can} uses probing-based logit descriptors to index pretrained networks, 
and \citet{horwitz2025we} construct a ``model atlas'' over thousands of Hugging Face models, 
facilitating retrieval based on latent functional proximity.  
Together, these methods redefine model discovery as an embedding-level search problem, bridging information retrieval and model understanding.

\subsubsection{Model Editing}

Beyond retrieval, WSR also enables controlled manipulation of model behavior within the embedding space.  
Instead of retraining, one can directly modify the latent representation $z = \phi(\theta)$ to steer a model’s function toward desired outcomes.  
Conceptually, editing can be formulated as an operation
\begin{equation}
    \tilde{\theta} = \phi^{-1}(z + \Delta z),
\end{equation}
where $\Delta z$ encodes the semantic or functional change to be applied.  
This formulation underlies recent advances in model editing and adaptation, 
which aim to adjust pretrained networks efficiently without full fine-tuning.  
For example, \citet{dravid2024interpreting} and \citet{schurholt2024towards} demonstrate that small, structured perturbations in embedding space 
can implement targeted behavioral modifications, such as bias removal or capability enhancement, while preserving overall model integrity.
Alternatively, some approaches operate directly in weight space and describe editing as
\begin{equation}
    \tilde{\theta} = \theta + \Delta(\theta),
\end{equation}
where $\Delta(\theta)$ is learned as a structured mapping over weights. For instance, \citet{zhou2023permutation,lim2024graph,kofinas2024graph,zhou2023neural} explore permutation-equivariant transformations in weight space to edit INRs and generate new data without full retraining.
This paradigm opens a new avenue for lightweight model refinement, 
where weight-space operations replace expensive gradient-based adaptation.

\subsection{Discussion and Perspective}

\begin{myboxi}[Takeaways]
\begin{itemize}
    \item WSR reframes neural networks as geometric data points in a latent manifold, enabling direct reasoning, comparison, and manipulation at the model level.  
    \item The next frontier lies in cross-architecture, self-supervised, and generative representations that unify understanding and creation within weight space.  
\end{itemize}
\end{myboxi}

Weight space representation establishes a paradigm where neural networks are not merely functions parameterized by weights but data points embedded in a structured latent space~\citep{dupont2022data}.  
By learning mappings $\phi: \Theta \rightarrow \mathbb{R}^d$, WSR enables direct reasoning, comparison, and manipulation of models as geometric objects, connecting structural regularities in parameters with observable functional behavior.  
This perspective unifies previously disconnected practices, from function prediction~\citep{unterthiner2020predicting} to model retrieval~\citep{horwitz2025learning} and editing~\citep{zhou2023permutation,dravid2024interpreting}, under a single representational framework.

\noindent\textbf{Strengths and Contributions.}
The key strength of WSR lies in its unifying capacity~\citep{zhou2024universal,pala2025gnn}.  
It bridges the gap between weight-level analysis and function-level generalization, offering a common language to describe models across architectures, datasets, and training regimes.  
By abstracting networks into embeddings, WSR supports model-level reasoning, enabling tasks such as zero-shot performance prediction~\citep{eilertsen2020classifying,unterthiner2020predicting}, scalable model search~\citep{kahana2025can,horwitz2025learning}, and efficient adaptation~\citep{schurholt2024towards,lim2024graph} without explicit access to data.  
In contrast to classical feature learning, WSR directly encodes the learners themselves~\citep{schurholt2021self}, opening a new dimension for understanding and controlling generalization.

\noindent\textbf{Limitations and Open Challenges.}
Despite its conceptual clarity, WSR remains constrained by several practical barriers.  
First, the high dimensionality and heterogeneity of modern architectures make representation learning sensitive to scale and design, impeding cross-architecture transfer~\citep{schurholt2024towards,kofinas2024graph}.  
Second, most data-driven approaches depend on large repositories of pretrained models, i.e., resources that remain unevenly distributed across domains~\citep{eilertsen2020classifying}. 
Third, the absence of standardized benchmarks and evaluation metrics hampers progress: while function prediction accuracy provides a partial signal, it fails to capture representation quality in terms of symmetry preservation or transferability~\citep{schurholt2022model,falk2025model}.  
Bridging these gaps requires both theoretical formalization and empirical grounding.

\noindent\textbf{Outlook.}
Looking forward, the future of WSR lies in evolving from descriptive encoders toward predictive and generative representations.  
Emerging directions include scalable cross-architecture embeddings that align models with diverse inductive biases, self-supervised formulations that exploit implicit symmetries, and integration with generative paradigms that not only represent but also synthesize weight configurations.  
As these lines converge, WSR will likely form the connective tissue between weight space understanding and weight space generation, turning weight space itself into a domain of learning, where networks can be compared, composed, and created within a unified representational space.




\section{Weight Space Generation} \label{section:wsg}



Weight Space Generation (WSG) extends the paradigm of weight space learning from \textit{understanding} and \textit{representation} to \textit{creation}.  
Rather than viewing weights merely as static parameters to be optimized, WSG treats them as generative objects that can be sampled, synthesized, or adapted within a learned distribution.  
In essence, WSG seeks to answer a natural question that follows weight space representation: \textit{If network weights can be represented, can they also be generated?} Formally, WSG aims to learn a mapping from the latent variables to model weight distributions:
\begin{equation}
    G_{\phi}: \mathcal{Z} \rightarrow \Theta,
\end{equation}
where a latent code $z \in \mathcal{Z}$ is transformed into the parameter set $\theta = G_{\phi}(z)$ of a target model.  
This mapping may capture global regularities of weight space, such as architectural symmetries, scaling relations, or functional similarity, enabling the generator to produce networks that are coherent, diverse, and task-aligned.

The motivation behind this framework is twofold.  
First, WSG provides a mechanism for model synthesis based on learned priors over weights, allowing new networks to be generated efficiently without full retraining.  
Second, it offers a foundation for adaptive model generation, where weight distributions are conditioned on external signals such as tasks, data domains, or performance objectives.  
Through this lens, the training of a model becomes the learning of a generative process over parameters rather than a pointwise search in weight space.
Recent studies have applied WSG in diverse settings, including meta-learning~\citep{nava2023metalearning,zhao2020meta}, continual learning~\citep{Oswald2020Continual,li2025continual}, federated learning~\citep{shamsian2021personalized,lai2025pfedgpa,li2024beyond}, and model fusion~\citep{tang2025data,tian2025text2weight}.
However, several challenges remain: the extreme dimensionality of weight space hinders efficient sampling; the learned priors often depend heavily on data distributions; and ensuring stability and generalization across architectures is still an open problem.

In the following sections, we summarize the main methodologies of WSG, categorize them into representative paradigms, and discuss their applications and limitations.  
Ultimately, WSG represents the creative frontier of weight space learning, where neural networks are no longer only optimized or represented, but generated.

\begin{figure}
    \centering
    \includegraphics[width=1\linewidth, trim={0cm 0.8cm 0cm 0.5cm}]
    {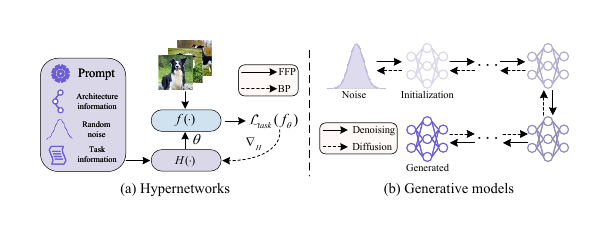}
    \caption{\textbf{Weight Space Generation Methods: Overview and Comparison.} \textbf{(a) Hypernetworks.} A conditioning prompt is processed by a hypernetwork $H(\cdot)$ to generate the parameters $\theta$ of a target model $f(\cdot)$. The hypernetwork is trained end-to-end using the downstream task loss $\mathcal{L}_{\text{task}}(f_\theta)$, enabling fast and conditional weight generation. \textbf{(b) Generative models.} Taking a diffusion-based approach as an example, the process begins by sampling from a noise distribution in weight space. Through iterative denoising steps, the model progressively refines the noisy initialization toward a structured representation of network parameters. This forward–reverse diffusion mechanism learns the underlying distribution of weights and enables conditional generation, producing diverse and task-aligned neural network configurations without explicit gradient-based optimization.}
    \label{fig:generation}
\end{figure}

    



\subsection{Generation Approaches}

\begin{myboxi}[Takeaways]
\begin{itemize}
    \item Hypernetworks generate weights through task-supervised, prompt-conditioned mapping, enabling fast instantiation and efficient adaptation, but their expressiveness is constrained by the conditioning signal and architecture of the generator.
    \item Generative model-based approaches treat trained weights as data and learn their underlying distribution, supporting diversity, model-level exploration, and the synthesis of weights beyond those seen during training.
    \item The key frontier lies in unifying these two paradigms, combining the controllability of hypernetworks with the diversity and scalability of generative models, to enable both precise adaptation and open-ended model synthesis.
\end{itemize}
\end{myboxi}

\begin{table*}[htbp]
\centering
\caption{\textbf{Comparison of Hypernetworks and generative models.} \text{\cmark} indicates full satisfaction, \text{\xmark} indicates the condition is not satisfied, and P denotes partial satisfaction.}
\label{tab:wsg comparison}
\renewcommand{\arraystretch}{1.15}
\setlength{\tabcolsep}{6pt}
\begin{tabular}{L{2.7cm} C{2.1cm} C{1.9cm} C{3.0cm} L{4.0cm}}
\toprule
\textbf{Method}   & \textbf{Conditioning} & \textbf{Diversity} & \textbf{Data Source} & \textbf{Suitable Scenarios} \\
\midrule
Hypernetworks     & \cmark                & \xmark             & real-world data            & Few-shot and fast adaptation \\
Generative models & P                     & \cmark             & model zoo                  & Large-scale and high-fidelity generation \\
\bottomrule
\end{tabular}
\end{table*}

Approaches to weight space generation can be broadly grouped according to the generative mechanism.  
At a high level, there are two major paradigms as shown in Figure \ref{fig:generation} and we compare them in Table \ref{tab:wsg comparison}:  
(i) Hypernetwork, where an auxiliary network produces the weights of the target model; and  
(ii) generative model, where a model learns to sample weights from the distribution of trained network weights.

\subsubsection{Hypernetworks}

Hypernetworks generate the parameters of a target neural network using another auxiliary network~\citep{ha2017hypernetworks}.  
Instead of optimizing the target model's weights directly through gradient descent, the hypernetwork learns a function that maps a conditioning signal (e.g., task description, architecture structure, class identity, or noise) to the weights of the main network.  
Once trained, the hypernetwork can instantly synthesize weights, enabling fast model instantiation, adaptation, or personalization.
Formally, given a target model with layers $\{W_1, \dots, W_L\}$, a hypernetwork $H$ outputs these weights conditioned on input $x_i$:
\begin{equation}
    W_i = H(x_i), \quad i = 1, \dots, L.
\end{equation}
During training, the main model processes task data and computes a loss.  
Gradients from this loss flow back to update only the hypernetwork, so the hypernetwork implicitly learns weight-generation patterns.  
After training, generating a new model reduces to a forward pass through $H$, avoiding iterative optimization.

Unlike standard generative models that learn a distribution over pretrained weights, a hypernetwork is trained end-to-end using task supervision. 
During training, the hypernetwork $H$ does not receive direct supervision on the weights it produces. 
Instead, it generates the parameters of the target model, which is then applied to the task, and the resulting task loss is backpropagated through the target model and into $H$. 
Thus, $H$ learns to produce weights that minimize downstream task loss:
\begin{equation}
    \theta = H(x), \qquad 
    \mathcal{L}_{task}(f_{\theta}) \xrightarrow{\text{backprop}} \nabla_{\!H}.
\end{equation}
This training scheme turns weight generation into a bi-level optimization problem: 
the target network is never explicitly optimized, but its performance defines the learning signal for the hypernetwork. 
In practice, the hypernetwork captures reusable structure in how weights vary across layers, tasks, domains, or input conditions, enabling fast instantiation of new models via a single forward pass~\citep{chauhan2024brief}.

\noindent\textbf{Parameter Efficiency: Why Hypernetworks Reduce Training Costs?}
The parameter savings of hypernetworks arise from replacing per-layer parameter learning with a shared weight-generation mechanism. 
Consider a convolutional network with $L$ layers, each containing $N_{\text{in}} \times N_{\text{out}}$ kernels of size $f \times f$, requiring $L \cdot N_{\text{in}} \cdot N_{\text{out}} \cdot f^2$
parameters to be learned directly.  
In a hypernetwork framework, the target layer weights are instead generated as $W_\ell = H(e_\ell)$, where $e_\ell \in \mathbb{R}^{d}$ is a compact embedding describing the layer (e.g., its index, shape, or role).  
The hypernetwork parameters are shared across all layers, so its parameter count depends only on the dimensions of $H$ rather than $L$.
For example, a two-layer hypernetwork with hidden width $h$ requires roughly
$L \cdot d + h \cdot d \cdot N_{\text{in}} + h \cdot N_{\text{out}}f^2$
parameters, where the first term accounts for per-layer embeddings, the second term corresponds to the first hidden layer that expands embeddings to $N_{\text{in}}$, and the third term generates the final convolutional weights for each output channel (biases omitted for simplicity).  
Since the second layer no longer scales with $N_{\text{in}}$ for every output filter, the hypernetwork significantly reduces complexity.
This design amortizes weight generation across layers and tasks, lowering optimization cost while capturing reusable patterns in weight structure.

\noindent\textbf{Input Conditioning: What Should a Hypernetwork Be Conditioned On?}
A central design choice in hypernetwork-based generation is how to construct the input embedding that conditions weight synthesis, as this embedding determines the specificity and adaptability of the generated weights. Early approaches such as HBN and BbH~\citep{krueger2017bayesian,pawlowski2017implicit} adopt stochastic embeddings sampled from Gaussian noise, enabling the hypernetwork to generate distributions over weights rather than deterministic outputs. Moving beyond noise-based prompts, \citet{brock2018smash} encode architectural topology directly, while~\citet{zhang2018graph} and~\citet{knyazev2021parameter} employ graph neural networks to extract structural embeddings from the target model’s computation graph before generating weights. To better handle multi-task or lifelong adaptation settings,~\citet{Oswald2020Continual} introduce chunked task embeddings that allow the hypernetwork to selectively reuse shared knowledge, and~\citet{beck2023hypernetworks} explore reinforcement-learned embeddings for adaptive control. Other works expand the expressive capacity of prompts: DHP~\citep{li2020dhp} maps input embeddings into higher-dimensional spaces followed by pruning to retain salient task structure; Sylph~\citep{yin2022sylph} augments prompts with class-level semantic information for emerging-category recognition; and more recent methods \citep{wang2024learning,littwin2019deep} bypass handcrafted embeddings entirely by conditioning the hypernetwork directly on input images or features. Collectively, these developments highlight the shift from generic prompts toward semantically structured conditioning, which increasingly positions the hypernetwork not just as a generator of weights, but as a mechanism for encoding and transferring task or model-level information.

\noindent\textbf{Output Granularity: What Does a Hypernetwork Generate?}
Another key axis of design concerns the form of the output produced by the hypernetwork. 
Early hypernetworks primarily generated full weight tensors for compact target networks \citep{ha2017hypernetworks,shamsian2021personalized,zhang2018graph}, but subsequent work broadened the notion of what it means to ``generate weights''. 
For instance, \citet{zhao2020meta} propose generating reusable weight chunks that can be linearly combined to assemble deep convolutional architectures, rather than producing each layer independently. 
In semantic segmentation, HyperSeg \citep{nirkin2021hyperseg} generates decoder weights conditioned on the encoder, thereby adapting downstream heads to varied visual domains. 
Within transformer-based language models, HyperFormer \citep{mahabadi2021parameter} generates the parameters of task-specific adapter modules, enabling parameter-efficient fine-tuning across tasks. 
More recently, HyperDreamBooth \citep{ruiz2024hyperdreambooth} uses a hypernetwork to initialize Low-Rank Adaptation (LoRA) modules for text-to-image personalization, effectively replacing costly gradient-based adaptation with direct weight synthesis. \citet{hedlin2025hypernet} learn to generate neural network weights by modeling the entire training trajectory via gradient supervision, allowing direct prediction of task-specific weights without requiring ground-truth optimization.
Together, these developments illustrate a shift from full-model generation to module-level and residual weight generation, allowing hypernetworks to scale to larger architectures and serve as a flexible interface for model adaptation, specialization, and personalization.

\noindent\textbf{Architectural Capacity: How Should a Hypernetwork Be Built?}
Early hypernetworks were instantiated as simple feed-forward or recurrent models \citep{ha2017hypernetworks}, sufficient for generating small parameter sets but limited in expressiveness.  
To scale toward deeper and more complex target networks, subsequent work enhanced the internal architecture of the hypernetwork itself.  
Residual connections have been incorporated to improve stability and allow deeper hypernetwork stacks \citep{alaluf2022hyperstyle,nirkin2021hyperseg}.  
Alternative designs draw from generative modeling: GAN-inspired hypernetworks \citep{ratzlaff2019hypergan,henning2018approximating} introduce adversarial feedback to encourage realistic weight distributions, while Transformer-based hypernetworks \citep{zhmoginov2022hypertransformer} treat weight generation as a sequence-to-sequence mapping, enabling parameter synthesis that scales to large architectures.  
These developments collectively transform hypernetworks from small auxiliary modules into expressive and scalable generators capable of producing high-dimensional, structurally consistent weight spaces.

\noindent\textbf{Generation Strategy: How Does a Hypernetwork Produce?}
Beyond directly emitting full weight tensors, hypernetworks increasingly adopt alternative generation modes that modify how weights are produced and applied.  
A first direction treats weights as distributions rather than point estimates: Bayesian hypernetworks such as HBN and BbH \citep{krueger2017bayesian,pawlowski2017implicit} sample weights from learned posterior approximations, enabling uncertainty-aware model instantiation.  
A second direction focuses on generating partial or task-specific components instead of entire networks.  
For example, Sylph \citep{yin2022sylph} predicts classifier weights to expand category sets in detection models, while HyperFormer \citep{mahabadi2021parameter} and HyperDreamBooth \citep{ruiz2024hyperdreambooth} generate adapter or LoRA parameters to enable efficient fine-tuning in transformers and diffusion models.  
A third line shifts from generating full weights to producing residual updates: methods such as HyperStyle, HyperInventor, and HyperEditor \citep{alaluf2022hyperstyle,dinh2022hyperinverter,zhang2024hypereditor} synthesize structured offsets that modulate existing pretrained generators, allowing targeted editing while preserving model identity.  
Finally, to address the extreme dimensionality of weight tensors, Polyhistor \citep{liu2022polyhistor} decomposes generation into low-rank factors, enabling scalable approximation for large architectures such as HyperTransformer \citep{zhmoginov2022hypertransformer}.  
Together, these generative strategies expand hypernetworks from direct parameter emitters to flexible controllers of how model weights are synthesized, adapted, or transformed.

\subsubsection{Generative Models}

Unlike hypernetworks that generate weights via a deterministic mapping $W = H(x)$ conditioned on prompts, generative model-based approaches treat trained network weights themselves as data and learn a distribution over the weight space. The goal is not to produce weights for one specific configuration, but to learn a generative model 
\begin{equation}
    p_\phi(W) \approx p_{\text{train}}(W),
\end{equation}
where $p_{\text{train}}(W)$ denotes the underlying distribution of model weights. Once $p_\phi$ is learned, it can be sampled to obtain new, diverse, and potentially improved network weights:
\begin{equation}
    W \sim p_\phi(W).
\end{equation}
This paradigm offers two practical advantages over hypernetwork-based generation:  
(1) it separates weight generation from task-specific conditioning, enabling the creation of weights even without explicit prompts;  
and (2) it supports diversity and exploration in the weight space, potentially discovering models that perform better than any single checkpoint in the training repository. We group generative WSG methods into four main families: (1) encoder-decoder approaches, which map weights into compact representations for efficient generation; (2) generative adversarial approaches, which synthesize weights via a generator trained in a minimax game against a discriminator; (3) autoregressive approaches, which treat model weights as sequences and generate them token-by-token; (4) iterative denoising approaches, which synthesize weights through a gradual refinement process from noise distributions.

\noindent\textbf{Autoencoders.}
These methods treat weight space generation as a dimensionality reduction and reconstruction problem. 
Formally, a generator (or decoder) $G_\phi$ maps a latent code $z$ sampled from a simple prior $p(z)$ (e.g., $\mathcal{N}(0, \mathbf{I})$) to the weight space:
\begin{equation}
    \theta = G_\phi(z), \quad \text{with} \quad z \in \mathbb{R}^d, \, d \ll \dim(\theta).
\end{equation}
The training objective depends on the specific architecture. For Variational Autoencoders (VAEs), the model maximizes the Evidence Lower Bound (ELBO) to align the approximate posterior with the prior.

Early work such as \citep{schurholt2021self,schurholt2022hyper} utilizes transformer encoders to compress flattened weight tensors, enabling interpolation within the latent space.
Subsequently, focus shifted to INRs, where weights represent data.
Functa \citep{dupont2022data} learns a meta-manifold of functions, decoding latent vectors into weight modulations for complex signals.
Recent work addresses the structure of deep networks and fine-tuning modules.
\citet{meynent2025structure} augment autoencoders with behavioral loss (matching outputs of $\theta$ vs.\ $\hat{\theta}$) and structural loss (preserving architectural constraints) to better capture functional semantics.
Most recently, several works focus on generating partial weights, particularly LoRA-style adaptation modules.  
DnD \citep{liang2025drag} derives task prompts using a text encoder and decodes them into LoRA parameters via a hyper-convolutional generator.  
Similarly, ICM-LoRA \citep{shao2025context} learns a task vector from a fine-tuned LLM and employs a conditional VAE to produce adaptation weights $\hat{\theta}_{\text{LoRA}} = D(z)$ where $z \sim q(z \mid \text{task})$.
SANE \citep{schurholt2024towards} extends sequential autoencoding to deeper networks, and \citet{falk2025learning} show scalability to large model repositories (e.g., Hugging Face), enabling zero-shot generation even for ViT-base.

\noindent\textbf{Generative Adversarial Networks.}
Instead of minimizing an explicit reconstruction error, Generative Adversarial Networks (GANs) optimize the generator $G_\phi$ via a minimax game against a discriminator $D_\psi$.
Instead of minimizing the distance between input and output weights directly, the generator aims to synthesize weights $\theta_{gen} = G_\phi(z)$ that are indistinguishable from real trained weights $\theta_{real}$, while the discriminator attempts to classify samples as real or fake.
Formally, the training objective is defined as:
\begin{equation}
    \min_{G_\phi} \max_{D_\psi} V(D, G) = \mathbb{E}_{\theta \sim p_{\text{data}}}[\log D_\psi(\theta)] + \mathbb{E}_{z \sim p(z)}[\log(1 - D_\psi(G_\phi(z)))].
\end{equation}
In the context of 3D-aware image synthesis, this paradigm has proven particularly effective for generating the parameters of neural radiance fields.
Generative Radiance Fields (GRAF) \citep{schwarz2020graf} introduces a framework that maps shape and appearance codes to the parameters of a conditional radiance field, enabling view-consistent scene generation from unposed 2D images.
Later, pi-GAN \citep{chan2021pi} significantly enhances generation  and employs a mapping network to modulate layer weights via frequencies and phase shifts.

\noindent\textbf{Autoregressive Approaches.}
Autoregressive methods treat weight space generation as a sequence modeling problem, analogous to natural language processing. These approaches decompose the weights into a sequence of discrete tokens or continuous chunks $\mathcal{S} = \{s_1, s_2, \dots, s_L\}$ and generate them step-by-step.
The joint probability distribution of the weights $p_\phi(\theta)$ is factorized into a product of conditional probabilities using the chain rule:
\begin{equation}
    p_\phi(\theta) = \prod_{t=1}^{L} p_\phi(s_t \mid s_{<t}, c),
\end{equation}
where $s_{<t}$ denotes the history of generated weight tokens and $c$ represents optional conditioning context (e.g., task embeddings).
During inference, weights are synthesized sequentially, allowing the model to capture complex inter-parameter dependencies and structural hierarchies.

To respect hierarchical architectures, Instruction-guided Parameter Generation (IGPG) \citep{soro2025instructionguided} decomposes generation layer-by-layer, conditioning each layer on previously generated history to ensure structural coherence.
Recurrent Parameter Generator (RPG) \citep{wang2025recurrent} adopts a hybrid strategy, utilizing a recurrent mechanism to iteratively generate tokenized weight representations, thereby scaling to large convolutional architectures.
The most recent frontier leverages Large Language Models (LLMs) to bridge natural language and weight space.
NeuroGen \citep{wang2025neurogen} fine-tunes LLMs to interpret task descriptions and output valid model parameters directly.
Similarly, Text-to-Weight (T2W) \citep{tian2025text2weight} aligns semantic text embeddings with weight space, enabling the zero-shot synthesis of network modules from natural language prompts.

\noindent\textbf{Iterative Denoising Approaches.}
Diffusion-based methods treat weight-space generation as a denoising problem, analogous to image or audio synthesis.  
Instead of producing weights directly, a diffusion process gradually adds noise to real weights $\theta$ to obtain a noisy version $\theta_t$, and a denoising model learns to reverse this process.  
Formally, the forward (noising) process is defined as:
\begin{equation}
    q(\theta_t \mid \theta_0) = \mathcal{N}(\theta_t;\sqrt{\alpha_t}\theta_0, (1-\alpha_t)\mathbf{I}),
\end{equation}
and a denoising model $f_\phi$ is trained to approximate the reverse transitions:
\begin{equation}
    p_\phi(\theta_{t-1} \mid \theta_t) \approx q(\theta_{t-1} \mid \theta_t, \theta_0).
\end{equation}
After training, sampling proceeds by iteratively denoising from $\theta_T$ (pure noise) back to $\theta_0$, producing model weights without gradient-based optimization.  
Compared with hypernetworks, which condition weight generation on prompts, diffusion-based methods learn the intrinsic distribution of weight space itself, enabling synthesis that preserves functional diversity and architectural semantics.

Early work demonstrates that diffusion models can successfully model the weight distributions of compact architectures such as MLPs and small CNNs.  
G.pt \citep{peebles2022learning} trains a transformer denoiser directly on historical checkpoints, enabling the generation of new networks from scratch.  
HyperDiffusion \citep{erkocc2023hyperdiffusion} applies a similar denoising strategy to implicit neural representations, while GPD \citep{yuan2024spatiotemporal} extends the formulation to spatio-temporal GNNs.  
BEND \citep{wei2024bend} ensembles multiple denoised samples to improve robustness, and p-diff \citep{wang2024neural} introduces a latent representation $z = E(\theta)$ prior to diffusion, allowing generation of larger backbone models such as ViT-Base and ConvNeXt-Base. In parallel, FLoWN \citep{saragih2025flow} introduces a flow matching framework that learns continuous trajectories in latent weight space via conditional vector fields.
Following that, recent work scales diffusion-based weight generation to large convolutional networks and even large language models.  
COND P-DIFF \citep{jin2024conditional} highlights the role of conditional inputs, enabling rapid task adaptation for LoRA modules.
DiffLoRA \citep{wu2024difflora} uses a diffusion transformer conditioned on visual features to generate weights for text-to-image diffusion models.  
RPG \citep{wang2025recurrent} proposes a recurrent denoising architecture, scaling to models like ConvNeXt-Large.  
ORAL \citep{khan2025oral} adds conditional control to the denoising trajectory, enabling personalized LoRA generation for models such as Mistral-7B and Qwen-7B-VL.  
D2NWG \citep{soro2025diffusionbased} generalizes conditioning to dataset and task descriptions, demonstrating weight generation for GPT-2 in latent space $\theta_0 = D\big(f_\phi(\theta_T, c)\big)$, where $c$ encodes semantic task information.

\subsection{Technique-Level Use Cases}

\begin{myboxi}[Takeaways]
\begin{itemize}
    \item WSG serves as a unified mechanism for adaptive model specialization, enabling weights to be synthesized on demand for new tasks, domains, or conditions without full retraining.
    \item WSG provides a principled way to compose, initialize, and even generate models or data directly in weight space.
\end{itemize}
\end{myboxi}

While WSG establishes a generative perspective on neural network parameters, its significance extends beyond theoretical interest.  
By enabling the direct synthesis of task- or domain-specific weights, WSG supports efficient model adaptation, composition, and reconstruction.  
We highlight several representative application settings where WSG demonstrates practical benefits, ranging from conditional and real-time weight generation to model merging and data synthesis.

\subsubsection{Conditional Weight Generation}

A key practical benefit of WSG is the ability to synthesize model parameters that are specialized for a particular task, domain, or semantic condition.  
Rather than learning a single static set of weights, conditional generation enables the model to produce weights that adapt to changing input distributions or class sets.
For example, Sylph \citep{yin2022sylph} uses a class-conditional hypernetwork to generate classifier weights for novel categories in incremental few-shot object detection.  
This allows the detector to expand its label space without retraining the full model.  
Similarly, in text-to-image and vision-language models, conditional diffusion-based methods such as HyperLDM \citep{nava2023meta} condition the denoising trajectory on multimodal CLIP features, enabling the generated weights to align with task-specific semantics. T2W \citep{tian2025text2weight} conditions a diffusion transformer on CLIP text embeddings to generate classifier heads directly from natural language task descriptions. 
In continual adaptation for object detection, \citet{li2025continual} integrate target-domain features as conditioning signals, allowing LoRA weights to be generated specifically for newly encountered domains.  
Across these lines of work, the shared idea is to treat weight synthesis as a conditioned mapping. 
This perspective frames WSG as an explicit mechanism for semantic control over model behavior, supporting flexible and data-efficient adaptation.

\subsubsection{Real-Time Weight Optimization}

A separate practical motivation for WSG concerns computational efficiency, especially when models must respond quickly, for instance in streaming, interactive, or real-time deployments.  
In these settings, performing gradient-based fine-tuning for each input instance is prohibitively slow.  
WSG offers a mechanism to bypass iterative optimization by synthesizing weights directly in a single forward pass.
HyperSeg \citep{nirkin2021hyperseg} illustrates this idea in semantic segmentation, where the encoder simultaneously extracts features and acts as a hypernetwork to generate decoder weights, reducing inference-time latency.  
HyperStyle \citep{alaluf2022hyperstyle} similarly trains a ResNet-based hypernetwork to rapidly modulate pretrained StyleGAN generators for fast image inversion.  
In wildlife classification, Wild-P-Diff \citep{zhang2025reimagining} uses location-conditioned diffusion to directly synthesize classifier heads suitable for new environments without retraining.
These methods are particularly valuable in latency-sensitive scenarios, where they enable rapid model adjustment to new tasks or domains without requiring full retraining or gradient-based fine-tuning.

\subsubsection{Model Merging}

WSG also provides a principled framework for model merging, where the goal is to combine the capabilities of multiple pretrained networks into a single model.  
Conventional merging approaches typically operate directly on raw weight tensors, such as averaging or interpolation \citep{wortsman2022model,tang2024merging}, but these strategies are highly sensitive to permutation symmetries, scaling mismatches, and optimization paths, often resulting in degraded performance.
In contrast, WSG-based merging does not treat weights as isolated numerical parameters, but instead models them as samples drawn from a structured distribution in weight space.  
By learning a shared representation of multiple networks, WSG enables merging to take place in a latent or hyper-representation space where structural relationships among models are encoded.  
This view is exemplified by work that encodes an entire model zoo into a shared representation space and samples from it to synthesize new models with strong performance \citep{schurholt2022hyper}.  
Similarly, \citet{zeng2025generative} show that generative weight models implicitly memorize structural patterns across many trained networks, enabling meaningful model recombination rather than naive averaging.  
To address permutation symmetry explicitly, Deep-Align \citep{navon2024equivariant} incorporates equivariant architectures such as DWSNets \citep{navon2023equivariant} to align models before merging, ensuring that weight correspondences are preserved across network instances.
Overall, WSG reframes model merging from direct parameter arithmetic to distribution-aware synthesis, enabling the construction of merged models that preserve functional behavior, avoid destructive interference, and generalize beyond any single source model.

\subsubsection{Weight Initialization}

Another practical use of WSG is to provide informed weight initialization for fine-tuning.  
Instead of starting optimization from a random initialization, WSG synthesizes initial weights that reflect prior structure learned from related models or tasks, reducing the number of optimization steps required and improving convergence stability.
Early works such as G.pt~\citep{peebles2022learning} and GHN-3~\citep{knyazev2023can} predict initialization weights for unseen models, outperforming conventional initialization methods.~\citet{schurholt2022hyper} decode the hyper-representation obtained from model zoos to initialize the weights of other models. 
For example, HyperDreamBooth \citep{ruiz2024hyperdreambooth} generates LoRA initialization weights conditioned on example images, enabling rapid personalization of text-to-image models without full fine-tuning.  
Across these methods, WSG reframes initialization as a learnable process, providing task-aware starting points that accelerate adaptation while preserving model quality.

\subsubsection{Training Acceleration}
WSG can also accelerate training by predicting future weights rather than relying solely on iterative gradient updates.
Different from SGD or Adam that update network weights step by step, WSG leverages structural priors and historical trajectories to forecast near-future weights, enabling the optimizer to “jump ahead” and skip redundant steps.
For example, WNN~\citep{jang2023learning} applies a lightweight predictor trained on past weight sequences to forecast the next update and replaces several gradient steps. NiNo~\citep{knyazev2025accelerating} extends this idea by leveraging neuron connectivity and GNNs to more accurately nowcast.
Across these approaches, WSG reframes optimization as a learnable forecasting process: instead of treating weight evolution as an opaque trajectory, it encodes historical and structural signals to anticipate future states. This perspective transforms training from a purely reactive procedure into a proactive one, reducing computational overhead while maintaining or even improving convergence stability.

\subsubsection{Data Generation}

When data itself is represented in weight space, WSG can be used not only to synthesize models, but also to generate data.  
Implicit Neural Representations (INRs) encode signals such as images or shapes directly as the weights of small neural networks.  
In this setting, generating new weights corresponds to generating new data instances.
Early work demonstrates this idea by using hypernetworks to generate INRs for 3D shape datasets such as ShapeNet \citep{littwin2019deep}.  
GASP \citep{dupont2022generative} extends this by integrating an adversarial discriminator, allowing the hypernetwork to better match the distribution of INR parameters for both ShapeNet and CelebA.  
Other approaches employ generative latent-variable models: NeuMeta \citep{yang2024neural} maps INR weights into a VAE latent space and samples new INRs to reconstruct data on MNIST and CelebA.  
In 3D-aware generation, Pi-GAN \citep{chan2021pi} uses a SIREN-based generator to synthesize view-consistent 3D scenes from 2D image collections.
Across these works, the key insight is that once data is embedded into weight space, WSG becomes a mechanism for data synthesis, enabling model-based data augmentation and generative 3D reconstruction.

\subsection{Discussion and Perspective} 

\begin{myboxi}[Takeaways]
\begin{itemize}
    \item WSG reframes neural networks as entities that can be directly synthesized in weight space, rather than obtained solely through gradient-based optimization.
    \item The main challenge ahead lies in reliably modeling the high-dimensional geometry and symmetries of weight space.
\end{itemize}
\end{myboxi}

Weight space generation reframes neural networks not as the result of an optimization trajectory, but as objects that can be directly synthesized.  
In contrast to conventional pipelines where weights emerge through gradient descent, WSG treats trained models as samples drawn from a structured space and seeks to learn the underlying generative mechanisms that govern this space~\citep{peebles2022learning,yuan2024spatiotemporal}.  
This view aligns naturally with weight space representation: where WSR embeds models into a latent space, WSG moves one step further by constructing new models from that space.  
Together, they shift the focus of learning from fitting functions to modeling the distribution of learners themselves.

\noindent\textbf{Strengths and Contributions.}
The core contribution of WSG lies in its ability to expose the global structure of model space.  
By learning how weights vary across tasks, datasets, or architectures, WSG decouples the act of constructing a model from the process of training it.  
This enables capabilities that are difficult to express in gradient-based learning: synthesizing families of models from semantic conditions~\citep{jin2024conditional,tian2025text2weight}, interpolating or recombining pretrained solutions~\citep{zeng2025generative,navon2024equivariant}, and instantiating models that reflect prior knowledge without revisiting data~\citep{schurholt2022hyper,schurholt2022model}.  
WSG therefore opens a new degree of freedom in neural network design: treating models not only as solutions to learning problems, but as generatable entities in their own right.

\noindent\textbf{Limitations and Open Challenges.}
Despite its conceptual clarity, WSG remains technically challenging.  
Weight space is high-dimensional, heterogeneous, and shaped by architectural symmetries that are not explicitly encoded in most generative models~\citep{navon2023equivariant}.  
Learning distributions over such spaces often requires large collections of pretrained networks, and the resulting generators can be fragile: small conditioning shifts may lead to large functional deviations~\citep{shamsian2024improved,zeng2025generative}.  
Moreover, evaluating generated models is nontrivial: performance alone does not reveal whether structural, functional, or symmetry-preserving properties are retained~\citep{zhao2025symmetry}.  
These difficulties highlight the need for both principled modeling assumptions and rigorous benchmarks.

\noindent\textbf{Outlook.}
We anticipate future progress along three methodological directions.  
First, integrating structural invariances from WSR into WSG may enable geometry-aware generation that respects permutation and scaling symmetries. 
Second, developing compact and expressive latent spaces for models may reduce dimensional complexity and improve controllability.  
Third, hybrid paradigms that blend generative synthesis with selective fine-tuning may offer a practical path to reliability.  
Ultimately, WSG points toward a learning regime in which neural networks are composed, adapted, and created in weight space directly, suggesting a perspective where models themselves become the fundamental units of inference and generalization.

\section{Applications to Related Domains} \label{section:application}

\begin{myboxi}[Takeaways]
\begin{itemize}
    \item Weight space methods provide a unified view of learning systems, where models themselves become manipulable objects, enabling retrieval, adaptation, composition, and generation without returning to data or training from scratch.
    \item By operating directly in weight space, these methods support scalable and architecture-agnostic workflows across diverse domains, connecting signal modeling, model reuse, continual adaptation, distributed learning, and automated architecture design under a shared framework.
\end{itemize}
\end{myboxi}

The previous sections have examined weight space methods from foundational manipulation (WSU), to representation and comparison (WSR), and finally to direct synthesis (WSG).  
Together, these components form a functional toolkit for operating on neural networks at the level of their parameters rather than their activations or outputs.  
We now turn to downstream applications where this perspective becomes operational: settings in which models are stored, transferred, adapted, combined, or generated directly in weight space, enabling new forms of generalization and system-level learning behavior (Related works are summarized in Table \ref{tab:app_domains}).

\subsection{Implicit Neural Representations} \label{sec:inr}

Implicit Neural Representations (INRs) offer a natural setting in which weight space becomes the primary medium for storing information.  
Instead of representing data (e.g., an image, a 3D shape, or a signal) as discrete arrays, an INR parameterizes it as a continuous function
$f_\theta : \mathbf{x} \mapsto \mathbf{y}$, where the signal is encoded directly in the weights $\theta$ of a small neural network.  
This weight-centric view aligns closely with the premise of weight space learning: models are no longer trained solely to solve tasks, but also serve as carriers of data.  
Thus, INRs provide a setting where WSG is not just a means of constructing models, but also a mechanism for data modeling and data synthesis.
Traditional INR research has focused on improving the expressivity and optimization of the implicit function itself, including spectral bias mitigation~\citep{rahaman2019spectral,fathony2020multiplicative} and activation designs such as sinusoidal or wavelet-inspired parameterizations~\citep{sitzmann2020implicit,saragadam2023wire,ramasinghe2022beyond,liu2024finer}.  
However, once signals are represented as weights, the resulting weight space becomes a structured dataset amenable to manipulation, transfer, retrieval, and generation, i.e., precisely the strengths of WSG.

\begin{table*}[!t]
\centering
\caption{Applications of Weight space learning to Related Domains.}
\label{tab:app_domains}
\resizebox{\linewidth}{!}{
\begin{tabular}{C{5cm} L{8.5cm} L{4.5cm}}
\toprule
\textbf{Domain} & \textbf{Representative References} & \textbf{Why WSL Fits} \\
\midrule
\textbf{Implicit Neural Representations (INR)} &
\citet{shamsian2023data}; \citet{shamsian2024improved}; \citet{huang2025few}; \citet{zhou2023neural}; inr2vec~\citep{luigi2023deep}; Functa~\citep{dupont2022data}; Spatial Functa~\citep{bauer2023spatial}; NeoMLP~\citep{kofinas2024mlp}; GRAF~\citet{schwarz2020graf}; pi-GAN~\citet{chan2021pi}; HyperDiffusion~\citet{erkocc2023hyperdiffusion}; \citet{yang2024neural}; \citet{gielisse2025end}. &
Data can be naturally represented by weights. \\ \midrule

\textbf{Model Unification} &
GEM~\citep{du2021learning}; Functa~\citep{dupont2022data}; Spatial Functa~\citep{bauer2023spatial}; uGNN~\citep{pala2025gnn}. &
Architecture-agnostic model space \\ \midrule

\textbf{Continual Learning} &
\citet{Oswald2020Continual}; W-Diff~\citep{xie2024weight}; \citet{li2025continual}. &
Forgetting mitigation via weight regeneration \\ \midrule

\textbf{Meta Learning} &
MetaDiff~\citep{zhang2024metadiff}; Mc-Di~\citep{guan2025learning}; MH~\citep{zhao2020meta}. &
Fast gradient-free adaptation \\ \midrule

\textbf{Federated Learning} &
FedDiff~\citep{li2024beyond}; pFedHN~\citep{shamsian2021personalized}; pFedGPA~\citep{lai2025pfedgpa}. &
Personalized weights, low communication \\ \midrule

\textbf{Neural Architecture Search} &
GHN~\citep{zhang2018graph}; GHN-1/2~\citep{knyazev2021parameter}. &
Efficient weights prediction \\
\bottomrule
\end{tabular}
}
\end{table*}

A first line of work explores weight space augmentation for improving INR robustness and data efficiency.  
Mixup in weight space is shown to generate functionally meaningful synthetic samples~\citep{shamsian2023data}, later extended to alignment-based semi-supervised learning~\citep{shamsian2024improved}.  
Permutation-equivariant augmentation further enforces weight-level symmetry constraints~\citep{huang2025few}, while neural functional transport~\citep{zhou2023neural} learns structured shifts in weight space to enrich INR datasets.
A second direction introduces latent representations of INRs to support retrieval, clustering, and classification.  
inr2vec~\citep{luigi2023deep} proposes a linear encoder that embeds INR weights for efficient model-level retrieval and comparison.  
Functa~\citep{dupont2022data} factorizes weights into a shared base representation and modulation parameters, significantly reducing storage cost, with Spatial Functa~\citep{bauer2023spatial} extending this idea to spatially coherent large-scale data.  
NeoMLP~\citep{kofinas2024mlp} represents INR weights as graphs and applies attention-based aggregation for classification tasks.
Finally, a rapidly growing direction uses generative models to synthesize INRs directly~\citep{wang2026molecular}.  
GAN-based approaches~\citep{schwarz2020graf,chan2021pi} learn distributions over INR parameters to generate new 3D-aware scenes, while diffusion-based and VAE-based methods~\citep{dupont2022data,erkocc2023hyperdiffusion,huang2025few,yang2024neural} sample INR weights to produce new images, shapes, or scene representations. 
In these settings, weight space generation effectively becomes data generation: sampling weights corresponds to synthesizing new continuous signals. In addition, ~\citep{gielisse2025end} uses an end-to-end training strategy to optimize the weight quality of INRs. These developments illustrate that INRs are not merely an application domain of WSG, but a setting in which the connection between weight and data is explicit.  
As WSG matures, INR-based pipelines may increasingly move toward scalable, data-independent, and architecture-agnostic learning over continuous signals, turning weight space into a unified medium for representation, retrieval, and synthesis.

\subsection{Model Unification} \label{sec:mu}

As neural networks are developed across diverse architectures, modalities, and tasks, a fundamental challenge arises: how can we represent and operate on these models within a shared space?  
Weight space representation provides a natural foundation for such unification.  
By treating models themselves as objects embedded in a structured representation space, weight space methods enable comparison, transfer, and manipulation across architectures, rather than within a fixed model class.  
This shifts the goal from designing universal architectures to developing architecture-agnostic model spaces.

Early efforts demonstrate the feasibility of representing heterogeneous networks within a common latent structure. GEM~\citep{du2021learning} uses MLP to represent arbitrary modalities and embed them in a unified latent manifold.
Functa~\citep{dupont2022data} encodes signals as implicit neural representations, enabling a unified representation space that spans datasets and modalities.  
Spatial Functa~\citep{bauer2023spatial} extends this principle to large-scale vision architectures by introducing spatially structured latent representations, improving transferability to models such as transformers and diffusion networks.
A parallel line of work focuses directly on aligning and comparing models in weight space.  
uGNN~\citep{pala2025gnn} interprets the weights of heterogeneous neural networks as graph-structured objects and introduces shared graph-meta parameters that permit joint optimization across architectures.  
More recently, architecture-agnostic weight embedding has been explored through graph-based meta-networks~\citep{ballerini2025weight}, enabling retrieval, clustering, and classification across fundamentally distinct model weights.
These developments illustrate an emerging perspective:  
model unification does not require a universal architecture—only a universal weight space.  
By constructing shared latent geometries over learned models, weight space methods open a path toward cross-domain transfer, scalable model retrieval, and composition across previously incompatible network families.

\subsection{Continual Learning} \label{sec:cl}

Continual Learning (CL) seeks to enable neural networks to acquire new tasks over time while retaining performance on previously learned ones.  
The core difficulty lies in catastrophic forgetting, where gradient updates for new tasks overwrite information encoded in the weights of earlier tasks~\citep{wang2024comprehensive}.  
From the perspective of weight space learning, this challenge can be reinterpreted not as a failure of optimization, but as a problem of weight management: each task corresponds to a position or trajectory in weight space, and forgetting arises when task-specific structure cannot be preserved.
Weight space learning offers a natural mechanism for addressing this issue. 
Rather than maintaining a single evolving set of weights, WSG enables task-conditioned synthesis of model parameters.  
Task-parameterized hypernetworks~\citep{Oswald2020Continual} illustrate this idea: a hypernetwork learns a mapping from compact task embeddings to the full model parameters, allowing past tasks to be recalled by regenerating their weights from stored embeddings.  
This approach shifts the memory burden from storing entire models to storing lightweight task descriptors.
More recent work explores generative modeling of differences in weight space.  
Diffusion-based methods such as W-Diff~\citep{xie2024weight} model residual weight updates between tasks, enabling controlled adaptation across evolving domains.  
Conditional variants further incorporate semantic or domain metadata into the denoising trajectory~\citep{li2025continual}, allowing the generated weights to align with new data without overwriting prior structure.
These developments suggest a shift in how continual learning is conceptualized:  
rather than preserving a single mutable model, CL can be understood as learning a generative model over task-specific weights.  
By treating models as regenerable rather than constantly overwritten objects, WSG provides a principled framework for mitigating forgetting while supporting scalable and data-efficient lifelong learning.

\subsection{Meta Learning} \label{sec:ml}

Meta-learning aims to enable models to adapt rapidly to new tasks from limited data, typically by extracting shared structure across a distribution of tasks~\citep{vettoruzzo2024advances}.  
Classical approaches such as gradient-based meta-learning~\citep{finn2017model} learn a set of meta-parameters that can be quickly fine-tuned for new tasks.  
However, these methods often rely on bi-level optimization, requiring backpropagation through inner learning loops, which introduces high computational cost and instability.
Weight space learning offers an alternative perspective: instead of learning to initialize a model for future fine-tuning, the meta-learner can directly learn to generate the task-specific weights themselves.  
This reframes meta-learning as a mapping from task descriptors to model weights, bypassing iterative adaptation entirely.

Diffusion-based approaches operationalize this idea by treating optimized weights as samples drawn from a task-conditioned distribution.  
MetaDiff~\citep{zhang2024metadiff} trains a conditional diffusion model to map noisy initial weights to task-specific optima, effectively replacing inner-loop updates with a denoising trajectory in weight space.  
Mc-Di~\citep{guan2025learning} further refines this process by constructing targets from intermediate optimization checkpoints, allowing the generative model to learn smoother weight trajectories across tasks.
Hypernetwork-based methods similarly frame adaptation as task-conditioned generation rather than gradient descent.  
Meta HyperNetworks (MH)~\citep{zhao2020meta} learn compact task embeddings and generate the corresponding model parameters via a hypernetwork, concentrating learning on the embedding space rather than full weight matrices.
These developments illustrate a shift in meta-learning:  
from optimizing initializations to directly synthesizing task-specific models.  
Viewed through the lens of weight space learning, meta-learning becomes a problem of modeling the structure of task-conditioned weight distributions, enabling fast, scalable, and gradient-free adaptation.

\subsection{Federated Learning} \label{sec:fl}

Federated learning (FL) enables multiple clients to collaboratively train models without sharing raw data~\citep{wen2023survey}, but faces well-known challenges such as data heterogeneity, communication cost, and model personalization.  
Weight space learning provides a natural perspective for addressing these limitations: instead of aggregating gradients or full models, the server can learn to generate model weights tailored to each client or shared across clients.

A first line of work focuses on improving global model aggregation.  
FedDiff~\citep{li2024beyond} learns a diffusion model on the server to directly synthesize global model weights from noisy initial parameters, reducing the need to transmit full client updates and improving communication efficiency.
A second direction emphasizes personalization by mapping client identity or statistics to client-specific model weights.  
pFedHN~\citep{shamsian2021personalized} employs a server-side hypernetwork that takes client embeddings as input and generates personalized weights (or gradients), enabling heterogeneous clients to retain local specialization while still benefiting from shared knowledge.  
Similarly, pFedGPA~\citep{lai2025pfedgpa} trains a conditional diffusion model that samples personalized client weights conditioned on learned embeddings, achieving personalization and privacy preservation without accessing client data.
Viewed through the lens of WSG, FL shifts from communicating gradients to communicating representations of model families.  
This enables more flexible and scalable collaboration: the server maintains a generative model of weight space, while individual client models become samples drawn from a shared but personalized distribution.

\subsection{Neural Architecture Search} \label{sec:nas}

Neural architecture search (NAS) seeks to automatically discover high-performing network architectures~\citep{han2024sadenas}, but the search process is often expensive, as evaluating each candidate typically requires partial or full training.  
Weight space learning offers an alternative perspective: if a model can generate the weights of a candidate architecture directly, then costly training-in-the-loop can be largely bypassed.
A representative line of work uses hypernetworks to generate weights conditioned on architecture structure.  
GHN~\citep{zhang2018graph} represents a candidate architecture as a computation graph, processes it using a graph neural network, and then predicts the corresponding model weights in a single forward pass.  
This enables rapid evaluation without gradient-based training.  
GHN-1/2~\cite{knyazev2021parameter} further scale this idea by training on large collections of diverse architectures, demonstrating strong zero-shot generalization to unseen networks.
Viewed through the lens of WSG, NAS shifts from searching architectures and training them to searching architectures and sampling their weights.  
This reframes architecture search as a problem of weight-conditioned structure modeling, reducing computational overhead while preserving functional fidelity.

\section{Benchmarks} \label{section:benchmark}

\begin{myboxi}[Takeaways]
\begin{itemize}
    \item Model zoos provide the empirical foundation for WSL by offering large, structured collections of pretrained weights, enabling the study of weight geometry, representation learning, and weight space generative modeling across architectures and training regimes.
    \item The progression from MLPs to CNNs, RNNs, and Transformers reflects an increasing richness of weight space structure, which in turn opens opportunities for more expressive WSR and more controllable, semantically aligned WSG in complex, modern architectures.
\end{itemize}
\end{myboxi}


\begin{table*}[!t]
  \centering
  \caption{Representative model zoos used in weight space research, covering diverse architectures, parameters, datasets, and scales.}
  \label{tab:model_zoo}
  \scriptsize
  \setlength{\tabcolsep}{6pt}
  \renewcommand{\arraystretch}{1.08}

  \begin{adjustbox}{width=\textwidth,center}
    \begin{tabular}{
      >{\raggedright\arraybackslash}p{3cm}
      >{\raggedright\arraybackslash}p{2.5cm}
      >{\raggedright\arraybackslash}p{5cm}
      >{\raggedright\arraybackslash}p{4cm}
      >{\centering\arraybackslash}p{0.9cm}}
      \toprule
      \textbf{Model Zoo} & \textbf{Architecture} & \textbf{Parameters} & \textbf{Datasets} & \textbf{Size} \\
      \midrule

      \multirow{3}{*}[\dimexpr-3ex]{\citet{schurholt2021self}}
        & \multirow{1}{*}[\dimexpr-1.3ex]{2 layers FFN} & Activation, Learning rate, Initialization, Seed & \multirow{1}{*}[\dimexpr-1.3ex]{tetris} & \multirow{1}{*}[\dimexpr-1.3ex]{3.9K} \\
      \cmidrule(lr){2-5}
        & \multirow{1}{*}[\dimexpr-2ex]{3 layers CNN} & \multirow{2}{*}[\dimexpr-0.5ex]{\parbox[t]{5cm}{Activation, Learning rate, Initialization, Seed, Dropout rate}} & MNIST & 9K+ \\
      \cmidrule(lr){4-5}
        &  &  & Fashion-MNIST & 1K \\
      \midrule

      \multirow{4}{*}[\dimexpr-2.2ex]{\parbox[t]{3cm}{Implicit-Zoo~\citep{ma2024implicit}}} 
        & 3 layers SIREN & \multirow{4}{*}[\dimexpr-2ex]{Implicit neural representation} & CIFAR-10 & 60K \\
      \cmidrule(lr){2-2} \cmidrule(lr){4-5}
        & 4 layers SIREN &  & ImageNet-1K & 1.4M+ \\
      \cmidrule(lr){2-2} \cmidrule(lr){4-5}
        & 5 layers SIREN &  & Cityscapes & 23.4K+ \\
      \cmidrule(lr){2-2} \cmidrule(lr){4-5}
        & 4 layers NERF &  & OmniObject3D & 5.9K+ \\
      \midrule

      \multirow{2}{*}[\dimexpr-0.7ex]{\citet{amaduzzi2025scaling}}
        & \multirow{2}{*}[\dimexpr-0.7ex]{3 layers NERF} & \multirow{2}{*}[\dimexpr-0.7ex]{Implicit neural representation} & G-Buffer Objaverse & 280K \\
      \cmidrule(lr){4-5}
        &  &  & ShapeNet & 40K \\
      \midrule

      \multirow{1}{*}[\dimexpr-1.2ex]{\parbox[t]{3cm}{NWS~\citep{eilertsen2020classifying}}} 
        & \multirow{1}{*}[\dimexpr-2.2ex]{3--5 layers CNN} 
        & \parbox[t]{5cm}{Learning rate, Batch size, Augmentation, Optimizer, Activation, Initialization, Filter size, Layer depth, Layer width}
        & \multirow{1}{*}[\dimexpr-1.2ex]{\parbox[t]{4cm}{MNIST, Fashion-MNIST, CIFAR-10, SVHN, STL-10}}
        & \multirow{1}{*}[\dimexpr-2.2ex]{320K} \\
      \midrule

      \multirow{4}{*}[\dimexpr-1.8ex]{\citet{unterthiner2020predicting}} 
        & \multirow{4}{*}[\dimexpr-1.8ex]{3 layers CNN}
        & \multirow{4}{*}[\dimexpr-1.8ex]{\parbox[t]{5cm}{Learning rate, $\mathscr{l}_2$ regularization coefficient, Optimizer, Dropout rate, Initialization, Activation}}
        & MNIST & 30K \\
      \cmidrule(lr){4-5}
        &  &  & Fashion-MNIST & 30K \\
      \cmidrule(lr){4-5}
        &  &  & CIFAR-10 & 30K \\
      \cmidrule(lr){4-5}
        &  &  & SVHN & 30K \\
      \midrule

      \multirow{1}{*}[\dimexpr-1.3ex]{\parbox[t]{3cm}{\citet{honegger2023sparsified}}}
        & 3 layers CNN (sparsified)
        & \parbox[t]{5cm}{Learning rate, Optimizer, Activation, Dropout rate, Weight decay, Seed}
        & \parbox[t]{4cm}{MNIST, Fashion-MNIST, CIFAR-10, SVHN, STL-10, USPS}
        & \multirow{1}{*}[\dimexpr-1.3ex]{1.7M+} \\
      \midrule

      \citet{kurtenbach2025open}
        & \multirow{1}{*}[\dimexpr-1.1ex]{LeNet-5}
        & \multirow{1}{*}[\dimexpr-1.1ex]{Image pairs}
        & \multirow{1}{*}[\dimexpr-1.1ex]{Imagenette V2 320px}
        & \multirow{1}{*}[\dimexpr-1.1ex]{10K} \\
      \midrule

      \multirow{4}{*}[\dimexpr-3ex]{\citet{schurholt2022model}}
        & \multirow{2}{*}[\dimexpr-0.4ex]{\parbox[t]{4cm}{3 layers CNN (small)}}
        & \multirow{3}{*}[\dimexpr-1.5ex]{\parbox[t]{5cm}{Learning rate, Optimizer, Activation, Dropout rate, Weight decay, Seed}}
        & \multirow{4}{*}[\dimexpr-1.ex]{\parbox[t]{4cm}{MNIST, Fashion-MNIST, CIFAR-10, SVHN, STL-10, USPS, CIFAR-100, Tiny ImageNet}}
        & \multirow{4}{*}[\dimexpr-3ex]{3.8M+} \\
        & & & & \\
        \cmidrule(lr){2-2}
        & \multirow{2}{*}[\dimexpr-0.4ex]{\parbox[t]{3cm}{3 layers CNN (medium)}} & & & \\
        & & & & \\
        \cmidrule(lr){2-4}
        & ResNet-18 & Seed & CIFAR-10 &  \\
      \midrule

      \multirow{1}{*}[\dimexpr-1ex]{\citet{herrmann2024learning}}
        & \multirow{1}{*}[\dimexpr-1ex]{LSTM}
        & \multirow{1}{*}[\dimexpr-1ex]{Learning rate, Steps, Seed}
        & \parbox[t]{4cm}{Formal Language, Sequential MNIST}
        & \multirow{1}{*}[\dimexpr-1ex]{9K} \\
      \midrule

      \multirow{1}{*}[\dimexpr-1.5ex]{\citet{dravid2024interpreting}}
        & LoRA (latent diffusion models)
        & \multirow{1}{*}[\dimexpr-1.5ex]{Fine-tuning}
        & \multirow{1}{*}[\dimexpr-1.5ex]{CelebA}
        & \multirow{1}{*}[\dimexpr-1.5ex]{60K} \\
      \midrule

      LoRA-WS~\citep{duszenko2025towards}
        & LoRA (stable diffusion models)
        & \multirow{1}{*}[\dimexpr-1.5ex]{Fine-tuning}
        & Subset of ImageNet (10 hierarchies)
        & \multirow{1}{*}[\dimexpr-1.5ex]{161K} \\
      \midrule

      \multirow{2}{*}[\dimexpr-2ex]{\citet{falk2025model}}
        & ViT-S
        & Task (supervised/contrastive), Seed
        & ImageNet
        & 10 \\
      \cmidrule(lr){2-5}
            & \multirow{1}{*}[\dimexpr-1.5ex]{MLP/LP}
        & Learning rate, Optimizer, Seed, Classification head
        & \multirow{1}{*}[\dimexpr-1.5ex]{CIFAR-100}
        & \multirow{1}{*}[\dimexpr-1.5ex]{240} \\
      \bottomrule
    \end{tabular}
  \end{adjustbox}
\end{table*}

Weight space learning treats neural network weights as the primary learning object rather than data or functions.  
This paradigm requires access to large collections of pretrained models, as weights now serve as samples drawn from a distribution over architectures, tasks, or domains.  
However, generating such collections from scratch is computationally expensive and often impractical, especially when diverse hyperparameters or training conditions must be represented.  
To address this, recent research has developed model zoos: benchmark datasets that store pretrained network parameters across architectures, datasets, and training pipelines.  
These model zoos provide a foundation for analyzing geometric structure in weight space, enabling representation learning, generative modeling, retrieval, and model-level generalization.

In this section, we review representative model zoos used in WSL research, summarized in Table~\ref{tab:model_zoo}.  
For clarity, we group them into four broad architecture families: MLPs, CNNs, RNNs, and Transformers.  
We describe the design motivations, scale, and typical use cases for each group below.

\subsection{Multilayer Perceptrons}

MLPs represent the simplest and most structurally uniform family of neural architectures, making them a natural starting point for WSL.  
Because MLPs contain minimal architectural variation, their weight spaces are easier to analyze and serve as a clean setting for studying the basic geometric and statistical properties.  
As such, MLP-based model zoos have played a central role in establishing early empirical evidence that weight space possesses meaningful structure.
One of the earliest systematic MLP zoos is introduced by~\citet{schurholt2021self}, who train 2-layer feedforward networks on a synthetic Tetris-shape discrimination task under diverse hyperparameter configurations.  
This produces a collection of 3.9K pretrained models representing controlled variations in initialization, optimization, and training dynamics, and has been widely used to evaluate weight similarity metrics, alignment strategies, and weight interpolation behaviors.

MLPs further become central when they are used as INRs, where the model weights encode the signal itself rather than features for downstream prediction.  
In this setting, each data instance directly corresponds to one trained MLP, turning datasets into large weight collections.
Implicit-Zoo~\citep{ma2024implicit} exemplifies this paradigm by constructing large-scale INR zoos for CIFAR-10 (60K models), ImageNet-1K (1.4M+ models), Cityscapes (23.4K+ models), and OmniObject3D (5.9K+ models), using SIREN architectures for 2D signals and NeRF-style MLPs for 3D scenes.  
Similarly,~\citet{amaduzzi2025scaling} build a large NeRF-based zoo for 3D objects in G-Buffer Objaverse and ShapeNet, totaling over 320K pretrained models.
Across these efforts, MLP zoos have served as the foundation for progressing from WSR to WSG:  
they provide a structured and scalable environment in which model embeddings, latent weight representations, and generative synthesis of weights can be effectively studied before extending to more complex architectures.

\subsection{Convolutional Neural Networks}

CNNs form the backbone of modern computer vision, and have correspondingly served as one of the earliest and most widely explored settings for constructing model zoos.  
Compared to MLPs, CNNs introduce spatial inductive biases such as local connectivity and weight sharing, making their weight spaces richer and more structured.  
This allows CNN zoos to function as testbeds for analyzing how architectural priors shape weight geometry.

Early CNN zoos focused on small networks (3–5 layers) trained under diverse hyperparameter configurations across standard datasets such as MNIST, CIFAR-10, SVHN, STL-10, and Fashion-MNIST.  
For example, NWS~\citep{eilertsen2020classifying} collects 16K CNNs trained under varied optimizers, learning rates, and augmentations, and further expands the zoo to 320K samples by storing intermediate weight checkpoints. 
Likewise,~\citet{unterthiner2020predicting} train 120K CNNs by varying hyperparameters while fixing architecture, enabling controlled study of weight variation across datasets.  
Following a similar design,~\citet{schurholt2021self} generate over 10K weight samples for MNIST and Fashion-MNIST by varying initialization seeds. 
Another construction strategy appears in~\citet{kurtenbach2025open}, which builds binary classification tasks by subsampling and pairing images to create a diverse LeNet-5 zoo without altering optimization settings.
More recent work expands both architectural diversity and dataset scale.  
\citet{schurholt2022model} construct 27 CNN-based model zoos spanning over 50K architecture variants, including ResNet families, resulting in more than 3.8M pretrained weights.  
Similarly,~\citet{honegger2023sparsified} generate 1.7M sparsified CNN models by systematically pruning networks trained across MNIST, CIFAR, STL-10, SVHN, and USPS.  
These large-scale zoos highlight how architectural and pruning-induced variations contribute to structure in weight space.

In diffusion-based generative models, weight diversity is primarily introduced through adapter-based fine-tuning.  
weights2weights (w2w)~\citep{dravid2024interpreting} collects 60K LoRA-tuned diffusion models, each personalized to a single image from CelebA, forming a zoo of stylistically varied UNet weights.  
Similarly, LoRA-WS~\citep{duszenko2025towards} constructs hierarchical ImageNet-based zoos of size 1K, 10K, 50K, and 100K, enabling weight-level analysis of semantic granularity.
These CNN-based model zoos form essential resources for studying how inductive biases, training procedures, and fine-tuning strategies shape the geometry of weight space, and consequently support downstream weight representation and generation.

\subsection{Recurrent Neural Networks}

Compared to vision, weight space datasets for sequence modeling are less common, reflecting the higher training cost and greater sensitivity of recurrent architectures to optimization dynamics.  
Nevertheless, recurrent model zoos play an important role in extending WSL beyond static inputs, enabling the study of temporal representation and state dynamics directly through weights.
A representative example is the LSTM model zoo constructed by~\citet{herrmann2024learning}, which contains models trained on tasks such as Formal Languages and Sequential MNIST.  
Each model is trained under different learning rates, random seeds, and data orderings, producing diverse trajectories in weight space.  
To capture learning dynamics in addition to end solutions, the authors record multiple checkpoints (nine per training run across 20,000 training steps), resulting in paired observations of intermediate and converged weights.  
This makes the zoo useful not only for analyzing final models, but also for studying weight evolution as a temporal process. 

\subsection{Transformers}

Transformers have become the dominant architecture in both language and vision, and recent work has begun extending weight space datasets into this domain.  
Compared to MLPs and CNNs, transformer model zoos are particularly valuable because the behavior of self-attention layers, normalization, and residual routing introduces richer symmetry and scaling structure in weight space.
A representative effort is the ViT-based zoo introduced by~\citet{falk2025model}, which studies how training paradigms shape weight distributions.  
Specifically, the authors pretrain ViT-S on ImageNet under two learning regimes: seven supervised training seeds and three contrastive self-supervised seeds.  
Each pretrained model is then fine-tuned on CIFAR-100 under 24 distinct hyperparameter configurations, yielding a collection of weights that differ not only in initialization and optimization, but also in training objectives and data regimes.
This enables controlled analysis of how representational geometry varies across supervised vs.\ self-supervised training, and how fine-tuning reshapes weight space structure.

\section{Open Questions}\label{section:conclusion}


Despite rapid progress, Weight Space Learning remains in its early stages.
While current techniques demonstrate that weights contain structured and learnable information, major conceptual, computational, and safety challenges remain.
We highlight three broad research directions that we believe are essential for advancing the field.

\subsection{Weight Space as a First-Class Learning Domain}

Many current successes of WSL appear in INRs, where signals are encoded directly into weights~\citep{dupont2022data,essakine2025where}.
However, these approaches typically treat weights as \emph{containers for data}, rather than as a domain with its own geometry and semantics.
To elevate weight space to a first-class learning domain, two key developments are needed.

\noindent\textbf{Geometric and Functional Foundations.}
Weight space is not an unstructured high-dimensional cloud, it contains symmetries (e.g., neuron permutations, positive scaling) and redundancies that create large equivalence classes of parameters~\citep{zhao2025symmetry}. Formalizing these invariances through quotient structures, a concept widely used in recent equivariant architectures and permutation-invariant functionals~\citep{navon2023equivariant,zhou2023permutation}, is essential for defining meaningful distances and avoiding redundant search directions. Beyond symmetry, overparameterization implies that functional variation often lies in low-dimensional subspaces, as evidenced by mode connectivity~\citep{entezari2022the} and low-rank adaptation methods~\citep{han2024parameterefficient}. Understanding these structures could enable reliable model interpolation, merging, and cross-task transfer directly in weight space. Future work should aim to couple these geometric insights with practical algorithms that operate across architectures without bespoke design.

\noindent\textbf{Universal Weight Space Learner.}
Existing representation models are typically tailored to specific architectures, requiring separate encoders for MLPs, CNNs, RNNs, Transformers, or diffusion models. A key open question is whether a structure-aware yet architecture-agnostic learner can operate directly on raw model weights. Recent efforts have explored MLP-based and message-passing models for embedding multi-modal and INR-style weight parameterizations~\citep{dupont2022data,bauer2023spatial,pala2025gnn}, but none provide a unified family that generalizes across heterogeneous architectures. Developing such a universal weight processor would move WSL toward learning over models themselves, enabling cross-architecture retrieval, warm-start initialization, and model-level transfer directly in weight space. Another promising direction is leveraging LoRA as a universal low-rank adaptation module~\citep{zhang2025reimagining,wu2024difflora}. Because LoRA represents weights through a compact linear transformation, it naturally exhibits strong symmetry properties and aligns well with weight space generation paradigms. If LoRA can serve as an architecture-agnostic representation, WSL could unify optimization across diverse model families, enabling scalable fine-tuning and transferability without bespoke design.

\subsection{Scaling Weight Space Learning to Large Models}

Current WSL methods are validated mostly on small or medium-sized networks such as MLPs and CNNs~\citep{schurholt2021self,ma2024implicit,schurholt2022model}. Although some works attempt to extend WSL to large-scale architectures such as transformers or diffusion models~\citep{dravid2024interpreting,duszenko2025towards,falk2025model}, these efforts typically explore partial weight spaces or undertrained checkpoints. We identify that scaling WSL faces a challenge: the input size of weight space grows dramatically with model depth and width, coupled with complex symmetry constraints and inter-layer dependencies. Directly operating on full weights becomes computationally prohibitive and hard to generalize. We highlight two key directions:

\noindent\textbf{Modular and Hierarchical Processing.}
Large-scale models exhibit repeated modules (e.g., attention blocks, feed-forward layers) with minor functional variations. Existing WSL approaches either decompose small networks~\citep{dupont2022data,bauer2023spatial} or combine a frozen base model with LoRA adapters~\citep{ruiz2024hyperdreambooth,khan2025oral,shao2025context,liu2022polyhistor}, but they rarely exploit the inherent redundancy and hierarchical organization of large architectures. A promising direction is to learn shared weight templates and distinguish instances via lightweight positional or contextual embeddings, analogous to positional encoding in transformers. Hypernetworks~\citep{ha2017hypernetworks} and recent modulation-based methods~\citep{dupont2022data,bauer2023spatial} demonstrate feasibility at smaller scales, yet the question remains: can we modulate an entire large-scale model through shared bases plus layer-specific tokens? Such approaches could drastically reduce dimensionality while preserving architectural fidelity, enabling scalable retrieval, editing, and generation in weight space.

\noindent\textbf{Fine-Tuning Modules.}
Recent studies indicate that weight space generation in WSL often resembles compositional generation~\citep{wiedemer2023compositional}, where learned components or modules are recombined rather than synthesized from scratch~\citep{zeng2025generative}. While WSG offers modularity and reuse, it inherently limits the ability to introduce new knowledge beyond what is already encoded in the components. This raises a critical question: can weight space generation move beyond pure composition toward a paradigm that actively leverages pretrained knowledge?
A promising direction is to learn to generate fine-tuning modules, such as adapters, prefix-tuning layers, or low-rank transformations, instead of full weight tensors~\citep{shao2025context,jin2024conditional}. These modules operate on top of a foundation model, which means the generated component can exploit the rich semantic and structural priors embedded in the base network. This capability allows WSL to fill gaps that compositional strategies cannot address, effectively transforming generation from simple recombination into knowledge-augmented synthesis. Such an approach could enable scalable adaptation across architectures, reduce computational overhead, and preserve functional fidelity by grounding new weights in pretrained representations.

\noindent\textbf{Compression and Approximation.}
Overparameterization suggests that large-scale weight spaces contain substantial redundancy, with functional variation concentrated in low-dimensional directions~\citep{entezari2022the,zhao2025symmetry}. Existing approaches~\citep{soro2025diffusionbased,schurholt2021self,huang2025few} that directly encode weights into latent vectors often ignore these structural properties. Thus a promising direction is to approximate weights in a symmetry-reduced subspace (similar to LoRA, with few weights influencing the function~\citep{hu2022lora}), effectively treating weights as data representations rather than raw large scale parameters. Such compression could drastically reduce computational and storage costs while preserving essential functional properties.

\subsection{Robustness and Safety in Weight Space Operations}

Weight space operations such as editing and generation introduce unique safety and robustness challenges. Most existing works prioritize performance and efficiency, leaving security largely unexplored. With the rapid development of treating weight space as data, approaches that safeguard weights against adversarial perturbations are becoming essential~\citep{sun2017hypernetworks,shor2025adversarial}. Here we outline two key research directions:

\noindent\textbf{Controllable Weight Generation.}
Hypernetworks and generative models demonstrate the feasibility of synthesizing weights~\citep{ha2017hypernetworks,soro2025diffusionbased}, but controllability remains an open challenge. Current generation pipelines often produce weights without explicit guarantees on their intended functionality, raising risks of harmful or unpredictable behaviors. Future work should explore weight functionals that integrate task descriptors, domain constraints, and safety rules into the generation process, ensuring that synthesized weights are auditable and aligned with desired objectives. Such controllable generation would shift WSL from opaque inner-loop adaptation to transparent, reliable weight synthesis.

\noindent\textbf{Adversarial Risk Detection and Defense in Weight Space.}
In WSL, weights themselves become the primary data modality, making adversarial attacks fundamentally different from input-level perturbations. These attacks operate in an abstract weight space, where malicious edits or injected subspaces can compromise model integrity without altering the original data. Future research should focus on two aspects: (i) \emph{detection}, by developing anomaly metrics and embedding-based probes to identify suspicious weight patterns or trajectories~\citep{shor2025adversarial}; and (ii) \emph{defense}, through symmetry-aware projections and reversible editing strategies that neutralize harmful components without full retraining. Establishing such protocols would make WSL robust not only to adversarial attacks but also to systemic risks inherited from contaminated training data.





\section{Conclusion}

Weight space learning reframes neural networks not merely as function approximators trained on data, but as structured objects inhabiting a rich and informative parameter space. 
This survey consolidates WSL into a coherent paradigm based on three complementary dimensions: weight space understanding, which studies the intrinsic structure, symmetries, and theoretical properties of weight space; weight space representation, which learns embeddings, metrics, and probing methods to support model-level reasoning; and weight space generation, which synthesizes or transforms weights through hypernetworks, and generative models. We further reviewed practical applications where WSL offers new capabilities, and summarized model zoo benchmarks that enable systematic evaluation and scaling.
Looking forward, we hope this survey helps unify ongoing research and encourages the community to view neural networks in a new light: not only as functions we train, but as entities we can analyze, represent, and create directly in weight space. As pretrained models continue to proliferate and model-centric learning becomes increasingly important, WSL stands poised to become a foundational perspective in the study and design of intelligent systems.

\appendix
\newpage

\newpage
\bibliography{reference}

@incollection{eilertsen2020classifying,
  title={Classifying the Classifier: Dissecting the Weight Space of Neural Networks},
  author={Eilertsen, Gabriel and J{\"o}nsson, Daniel and Ropinski, Timo and Unger, Jonas and Ynnerman, Anders},
  booktitle={ECAI 2020},
  pages={1119--1126},
  year={2020},
  publisher={IOS Press}
}

@article{unterthiner2020predicting,
  title={Predicting neural network accuracy from weights},
  author={Unterthiner, Thomas and Keysers, Daniel and Gelly, Sylvain and Bousquet, Olivier and Tolstikhin, Ilya},
  journal={arXiv preprint arXiv:2002.11448},
  year={2020}
}

@article{martin2021predicting,
  title={Predicting trends in the quality of state-of-the-art neural networks without access to training or testing data},
  author={Martin, Charles H and Peng, Tongsu and Mahoney, Michael W},
  journal={Nature Communications},
  volume={12},
  number={1},
  pages={4122},
  year={2021},
  publisher={Nature Publishing Group UK London}
}

@article{zhou2023neural,
  title={Neural functional transformers},
  author={Zhou, Allan and Yang, Kaien and Jiang, Yiding and Burns, Kaylee and Xu, Winnie and Sokota, Samuel and Kolter, J Zico and Finn, Chelsea},
  journal={Advances in neural information processing systems},
  volume={36},
  pages={77485--77502},
  year={2023}
}

@inproceedings{dupont2022data,
  title={From data to functa: Your data point is a function and you can treat it like one},
  author={Dupont, Emilien and Kim, Hyunjik and Eslami, SM Ali and Rezende, Danilo Jimenez and Rosenbaum, Dan},
  booktitle={International Conference on Machine Learning},
  pages={5694--5725},
  year={2022},
  organization={PMLR}
}

@article{bauer2023spatial,
  title={Spatial functa: Scaling functa to imagenet classification and generation},
  author={Bauer, Matthias and Dupont, Emilien and Brock, Andy and Rosenbaum, Dan and Schwarz, Jonathan Richard and Kim, Hyunjik},
  journal={arXiv preprint arXiv:2302.03130},
  year={2023}
}

@inproceedings{navon2023equivariant,
  title={Equivariant architectures for learning in deep weight spaces},
  author={Navon, Aviv and Shamsian, Aviv and Achituve, Idan and Fetaya, Ethan and Chechik, Gal and Maron, Haggai},
  booktitle={International Conference on Machine Learning},
  pages={25790--25816},
  year={2023},
  organization={PMLR}
}

@inproceedings{zhou2023permutation,
 author = {Zhou, Allan and Yang, Kaien and Burns, Kaylee and Cardace, Adriano and Jiang, Yiding and Sokota, Samuel and Kolter, J. Zico and Finn, Chelsea},
 booktitle = {Advances in Neural Information Processing Systems},
 editor = {A. Oh and T. Naumann and A. Globerson and K. Saenko and M. Hardt and S. Levine},
 pages = {24966--24992},
 publisher = {Curran Associates, Inc.},
 title = {Permutation Equivariant Neural Functionals},
 volume = {36},
 year = {2023}
}

@article{zhao2025symmetry,
  title={Symmetry in Neural Network Parameter Spaces},
  author={Zhao, Bo and Walters, Robin and Yu, Rose},
  journal={arXiv preprint arXiv:2506.13018},
  year={2025}
}

@article{kofinas2024mlp,
  title={From MLP to NeoMLP: Leveraging Self-Attention for Neural Fields},
  author={Kofinas, Miltiadis and Papa, Samuele and Gavves, Efstratios},
  journal={arXiv preprint arXiv:2412.08731},
  year={2024}
}

@inproceedings{shamsian2024improved,
  title={Improved generalization of weight space networks via augmentations},
  author={Shamsian, Aviv and Navon, Aviv and Zhang, David W and Zhang, Yan and Fetaya, Ethan and Chechik, Gal and Maron, Haggai},
  booktitle={Proceedings of the 41st International Conference on Machine Learning},
  pages={44378--44393},
  year={2024}
}

@inproceedings{dupont2022generative,
  title={Generative Models as Distributions of Functions},
  author={Dupont, Emilien and Teh, Yee Whye and Doucet, Arnaud},
  booktitle={International Conference on Artificial Intelligence and Statistics},
  pages={2989--3015},
  year={2022},
  organization={PMLR}
}

@article{pala2025gnn,
  title={GNN-based Unified Deep Learning},
  author={Pala, Furkan and Rekik, Islem},
  journal={arXiv preprint arXiv:2508.10583},
  year={2025}
}

@inproceedings{
dravid2024interpreting,
title={Interpreting the Weight Space of Customized Diffusion Models},
author={Amil Dravid and Yossi Gandelsman and Kuan-Chieh Wang and Rameen Abdal and Gordon Wetzstein and Alexei A Efros and Kfir Aberman},
booktitle={The Thirty-eighth Annual Conference on Neural Information Processing Systems},
year={2024},
}

@inproceedings{
lim2024graph,
title={Graph Metanetworks for Processing Diverse Neural Architectures},
author={Derek Lim and Haggai Maron and Marc T. Law and Jonathan Lorraine and James Lucas},
booktitle={The Twelfth International Conference on Learning Representations},
year={2024},
}

@article{zaheer2017deep,
  title={Deep sets},
  author={Zaheer, Manzil and Kottur, Satwik and Ravanbakhsh, Siamak and Poczos, Barnabas and Salakhutdinov, Russ R and Smola, Alexander J},
  journal={Advances in neural information processing systems},
  volume={30},
  year={2017}
}

@inproceedings{yang2024neural,
  title={Neural metamorphosis},
  author={Yang, Xingyi and Wang, Xinchao},
  booktitle={European Conference on Computer Vision},
  pages={1--19},
  year={2024},
  organization={Springer}
}

@article{andreis2024set,
  title={Set-based neural network encoding without weight tying},
  author={Andreis, Bruno and Soro, Bedionita and Torr, Philip and Hwang, Sung Ju},
  journal={Advances in Neural Information Processing Systems},
  volume={37},
  pages={76500--76528},
  year={2024}
}

@inproceedings{moskalev2023genuine,
  title={On genuine invariance learning without weight-tying},
  author={Moskalev, Artem and Sepliarskaia, Anna and Bekkers, Erik J and Smeulders, Arnold WM},
  booktitle={Topological, Algebraic and Geometric Learning Workshops 2023},
  pages={218--227},
  year={2023},
  organization={PMLR}
}

@inproceedings{hartford2018deep,
  title={Deep models of interactions across sets},
  author={Hartford, Jason and Graham, Devon and Leyton-Brown, Kevin and Ravanbakhsh, Siamak},
  booktitle={International Conference on Machine Learning},
  pages={1909--1918},
  year={2018},
  organization={PMLR}
}

@inproceedings{
kahana2025deep,
title={Deep Linear Probe Generators for Weight Space Learning},
author={Jonathan Kahana and Eliahu Horwitz and Imri Shuval and Yedid Hoshen},
booktitle={The Thirteenth International Conference on Learning Representations},
year={2025},
}

@article{kahana2025can,
  title={Can this model also recognize dogs? zero-shot model search from weights},
  author={Kahana, Jonathan and Nathan, Or and Horwitz, Eliahu and Hoshen, Yedid},
  journal={arXiv preprint arXiv:2502.09619},
  year={2025}
}

@inproceedings{horwitz2025learning,
  title={Learning on model weights using tree experts},
  author={Horwitz, Eliahu and Cavia, Bar and Kahana, Jonathan and Hoshen, Yedid},
  booktitle={Proceedings of the Computer Vision and Pattern Recognition Conference},
  pages={20468--20478},
  year={2025}
}

@inproceedings{herrmann2024learning,
  title={Learning useful representations of recurrent neural network weight matrices},
  author={Herrmann, Vincent and Faccio, Francesco and Schmidhuber, J{\"u}rgen},
  booktitle={Proceedings of the 41st International Conference on Machine Learning},
  pages={18205--18227},
  year={2024}
}

@inproceedings{
ashkenazi2023nern,
title={Ne{RN}: Learning Neural Representations for Neural Networks},
author={Maor Ashkenazi and Zohar Rimon and Ron Vainshtein and Shir Levi and Elad Richardson and Pinchas Mintz and Eran Treister},
booktitle={The Eleventh International Conference on Learning Representations },
year={2023},
}

@inproceedings{
ha2017hypernetworks,
title={HyperNetworks},
author={David Ha and Andrew M. Dai and Quoc V. Le},
booktitle={International Conference on Learning Representations},
year={2017},
}

@article{krueger2017bayesian,
  title={Bayesian hypernetworks},
  author={Krueger, David and Huang, Chin-Wei and Islam, Riashat and Turner, Ryan and Lacoste, Alexandre and Courville, Aaron},
  journal={arXiv preprint arXiv:1710.04759},
  year={2017}
}

@article{pawlowski2017implicit,
  title={Implicit weight uncertainty in neural networks},
  author={Pawlowski, Nick and Brock, Andrew and Lee, Matthew CH and Rajchl, Martin and Glocker, Ben},
  journal={arXiv preprint arXiv:1711.01297},
  year={2017}
}

@article{henning2018approximating,
  title={Approximating the predictive distribution via adversarially-trained hypernetworks},
  author={Henning, Christian and von Oswald, Johannes and Sacramento, Jo{\~a}o and Surace, Simone Carlo and Pfister, Jean-Pascal and Grewe, Benjamin F},
  year={2018},
  publisher={Yarin}
}

@inproceedings{ratzlaff2019hypergan,
  title={Hypergan: A generative model for diverse, performant neural networks},
  author={Ratzlaff, Neale and Fuxin, Li},
  booktitle={International Conference on Machine Learning},
  pages={5361--5369},
  year={2019},
  organization={PMLR}
}

@inproceedings{
brock2018smash,
title={{SMASH}: One-Shot Model Architecture Search through HyperNetworks},
author={Andrew Brock and Theo Lim and J.M. Ritchie and Nick Weston},
booktitle={International Conference on Learning Representations},
year={2018},
}

@inproceedings{
zhang2018graph,
title={Graph HyperNetworks for Neural Architecture Search},
author={Chris Zhang and Mengye Ren and Raquel Urtasun},
booktitle={International Conference on Learning Representations},
year={2019},
}

@article{knyazev2021parameter,
  title={Parameter prediction for unseen deep architectures},
  author={Knyazev, Boris and Drozdzal, Michal and Taylor, Graham W and Romero Soriano, Adriana},
  journal={Advances in Neural Information Processing Systems},
  volume={34},
  pages={29433--29448},
  year={2021}
}

@inproceedings{
Oswald2020Continual,
title={Continual learning with hypernetworks},
author={Johannes von Oswald and Christian Henning and Benjamin F. Grewe and João Sacramento},
booktitle={International Conference on Learning Representations},
year={2020},
}

@inproceedings{zhang2024hypereditor,
  title={Hypereditor: Achieving both authenticity and cross-domain capability in image editing via hypernetworks},
  author={Zhang, Hai and Wu, Chunwei and Cao, Guitao and Wang, Hailing and Cao, Wenming},
  booktitle={Proceedings of the AAAI Conference on Artificial Intelligence},
  volume={38},
  number={7},
  pages={7051--7059},
  year={2024}
}

@inproceedings{ruiz2024hyperdreambooth,
  title={Hyperdreambooth: Hypernetworks for fast personalization of text-to-image models},
  author={Ruiz, Nataniel and Li, Yuanzhen and Jampani, Varun and Wei, Wei and Hou, Tingbo and Pritch, Yael and Wadhwa, Neal and Rubinstein, Michael and Aberman, Kfir},
  booktitle={Proceedings of the IEEE/CVF conference on computer vision and pattern recognition},
  pages={6527--6536},
  year={2024}
}

@inproceedings{nirkin2021hyperseg,
  title={Hyperseg: Patch-wise hypernetwork for real-time semantic segmentation},
  author={Nirkin, Yuval and Wolf, Lior and Hassner, Tal},
  booktitle={Proceedings of the IEEE/CVF conference on computer vision and pattern recognition},
  pages={4061--4070},
  year={2021}
}

@inproceedings{beck2023hypernetworks,
  title={Hypernetworks in meta-reinforcement learning},
  author={Beck, Jacob and Jackson, Matthew Thomas and Vuorio, Risto and Whiteson, Shimon},
  booktitle={Conference on Robot Learning},
  pages={1478--1487},
  year={2023},
  organization={PMLR}
}

@article{chauhan2024brief,
  title={A brief review of hypernetworks in deep learning},
  author={Chauhan, Vinod Kumar and Zhou, Jiandong and Lu, Ping and Molaei, Soheila and Clifton, David A},
  journal={Artificial Intelligence Review},
  volume={57},
  number={9},
  pages={250},
  year={2024},
  publisher={Springer}
}

@inproceedings{li2020dhp,
  title={Dhp: Differentiable meta pruning via hypernetworks},
  author={Li, Yawei and Gu, Shuhang and Zhang, Kai and Van Gool, Luc and Timofte, Radu},
  booktitle={European conference on computer vision},
  pages={608--624},
  year={2020},
  organization={Springer}
}

@article{wang2024learning,
  title={Learning to generate parameters of convnets for unseen image data},
  author={Wang, Shiye and Feng, Kaituo and Li, Changsheng and Yuan, Ye and Wang, Guoren},
  journal={IEEE Transactions on Image Processing},
  year={2024},
  publisher={IEEE}
}

@inproceedings{littwin2019deep,
  title={Deep meta functionals for shape representation},
  author={Littwin, Gidi and Wolf, Lior},
  booktitle={Proceedings of the IEEE/CVF international conference on computer vision},
  pages={1824--1833},
  year={2019}
}

@inproceedings{dinh2022hyperinverter,
  title={Hyperinverter: Improving stylegan inversion via hypernetwork},
  author={Dinh, Tan M and Tran, Anh Tuan and Nguyen, Rang and Hua, Binh-Son},
  booktitle={Proceedings of the IEEE/CVF conference on computer vision and pattern recognition},
  pages={11389--11398},
  year={2022}
}

@inproceedings{alaluf2022hyperstyle,
  title={Hyperstyle: Stylegan inversion with hypernetworks for real image editing},
  author={Alaluf, Yuval and Tov, Omer and Mokady, Ron and Gal, Rinon and Bermano, Amit},
  booktitle={Proceedings of the IEEE/CVF conference on computer Vision and pattern recognition},
  pages={18511--18521},
  year={2022}
}

@inproceedings{zhmoginov2022hypertransformer,
  title={Hypertransformer: Model generation for supervised and semi-supervised few-shot learning},
  author={Zhmoginov, Andrey and Sandler, Mark and Vladymyrov, Maksym},
  booktitle={International Conference on Machine Learning},
  pages={27075--27098},
  year={2022},
  organization={PMLR}
}

@article{liu2022polyhistor,
  title={Polyhistor: Parameter-efficient multi-task adaptation for dense vision tasks},
  author={Liu, Yen-Cheng and Ma, Chih-Yao and Tian, Junjiao and He, Zijian and Kira, Zsolt},
  journal={Advances in Neural Information Processing Systems},
  volume={35},
  pages={36889--36901},
  year={2022}
}

@inproceedings{mahabadi2021parameter,
  title={Parameter-efficient Multi-task Fine-tuning for Transformers via Shared Hypernetworks},
  author={Mahabadi, Rabeeh Karimi and Ruder, Sebastian and Dehghani, Mostafa and Henderson, James},
  booktitle={Proceedings of the 59th Annual Meeting of the Association for Computational Linguistics and the 11th International Joint Conference on Natural Language Processing (Volume 1: Long Papers)},
  pages={565--576},
  year={2021}
}

@inproceedings{shamsian2021personalized,
  title={Personalized federated learning using hypernetworks},
  author={Shamsian, Aviv and Navon, Aviv and Fetaya, Ethan and Chechik, Gal},
  booktitle={International conference on machine learning},
  pages={9489--9502},
  year={2021},
  organization={PMLR}
}

@inproceedings{lai2025pfedgpa,
  title={pFedGPA: Diffusion-based Generative Parameter Aggregation for Personalized Federated Learning},
  author={Lai, Jiahao and Li, Jiaqi and Xu, Jian and Wu, Yanru and Tang, Boshi and Chen, Siqi and Huang, Yongfeng and Ding, Wenbo and Li, Yang},
  booktitle={Proceedings of the AAAI Conference on Artificial Intelligence},
  volume={39},
  number={17},
  pages={17999--18007},
  year={2025}
}

@inproceedings{zhang2025reimagining,
  title={Reimagining Parameter Space Exploration with Diffusion Models},
  author={Zhang, Lijun and Liu, Xiao and Guan, Hui},
  booktitle={The Exploration in AI Today Workshop at ICML 2025},
  year={2025}
}

@article{xie2024weight,
  title={Weight Diffusion for Future: Learn to Generalize in Non-Stationary Environments},
  author={Xie, Mixue and Li, Shuang and Xie, Binhui and Liu, Chi and Liang, Jian and Sun, Zixun and Feng, Ke and Zhu, Chengwei},
  journal={Advances in Neural Information Processing Systems},
  volume={37},
  pages={6367--6392},
  year={2024}
}

@inproceedings{yin2022sylph,
  title={Sylph: A hypernetwork framework for incremental few-shot object detection},
  author={Yin, Li and Perez-Rua, Juan M and Liang, Kevin J},
  booktitle={Proceedings of the IEEE/CVF conference on computer vision and pattern recognition},
  pages={9035--9045},
  year={2022}
}

@inproceedings{zhao2020meta,
  title={Meta-learning via hypernetworks},
  author={Zhao, Dominic and Kobayashi, Seijin and Sacramento, Jo{\~a}o and von Oswald, Johannes},
  booktitle={4th Workshop on Meta-Learning at NeurIPS 2020 (MetaLearn 2020)},
  year={2020},
  organization={NeurIPS}
}

@article{wang2025neurogen,
  title={NeuroGen: Neural Network Parameter Generation via Large Language Models},
  author={Wang, Jiaqi and Zhang, Yusen and Li, Xi},
  journal={arXiv preprint arXiv:2505.12470},
  year={2025}
}

@inproceedings{schurholt2024towards,
  title={Towards Scalable and Versatile Weight Space Learning},
  author={Sch{\"u}rholt, Konstantin and Mahoney, Michael W and Borth, Damian},
  booktitle={International Conference on Machine Learning},
  pages={43947--43966},
  year={2024},
  organization={PMLR}
}

@article{shao2025context,
  title={In-Context Meta LoRA Generation},
  author={Shao, Yihua and Yan, Minxi and Liu, Yang and Chen, Siyu and Chen, Wenjie and Long, Xinwei and Yan, Ziyang and Li, Lei and Zhang, Chenyu and Sebe, Nicu and others},
  journal={arXiv preprint arXiv:2501.17635},
  year={2025}
}

@inproceedings{
yuan2024spatiotemporal,
title={Spatio-Temporal Few-Shot Learning via Diffusive Neural Network Generation},
author={Yuan Yuan and Chenyang Shao and Jingtao Ding and Depeng Jin and Yong Li},
booktitle={The Twelfth International Conference on Learning Representations},
year={2024},
}

@article{jin2024conditional,
  title={Conditional lora parameter generation},
  author={Jin, Xiaolong and Wang, Kai and Tang, Dongwen and Zhao, Wangbo and Zhou, Yukun and Tang, Junshu and You, Yang},
  journal={arXiv preprint arXiv:2408.01415},
  year={2024}
}

@article{li2025continual,
  title={Continual Adaptation: Environment-Conditional Parameter Generation for Object Detection in Dynamic Scenarios},
  author={Li, Deng and Wu, Aming and Li, Yang and Wang, Yaowei and Han, Yahong},
  journal={arXiv preprint arXiv:2506.24063},
  year={2025}
}

@article{wu2024difflora,
  title={Difflora: Generating personalized low-rank adaptation weights with diffusion},
  author={Wu, Yujia and Shi, Yiming and Wei, Jiwei and Sun, Chengwei and Yang, Yang and Shen, Heng Tao},
  journal={arXiv preprint arXiv:2408.06740},
  year={2024}
}

@article{wang2025recurrent,
  title={Recurrent diffusion for large-scale parameter generation},
  author={Wang, Kai and Tang, Dongwen and Zhao, Wangbo and Sch{\"u}rholt, Konstantin and Wang, Zhangyang and You, Yang},
  journal={arXiv preprint arXiv:2501.11587},
  year={2025}
}

@article{khan2025oral,
  title={ORAL: Prompting Your Large-Scale LoRAs via Conditional Recurrent Diffusion},
  author={Khan, Rana Muhammad Shahroz and Tang, Dongwen and Li, Pingzhi and Wang, Kai and Chen, Tianlong},
  journal={arXiv preprint arXiv:2503.24354},
  year={2025}
}

@inproceedings{huang2025few,
  title={Few-shot Implicit Function Generation via Equivariance},
  author={Huang, Suizhi and Yang, Xingyi and Lu, Hongtao and Wang, Xinchao},
  booktitle={Proceedings of the Computer Vision and Pattern Recognition Conference},
  pages={16262--16272},
  year={2025}
}

@article{sun2017hypernetworks,
  title={Hypernetworks with statistical filtering for defending adversarial examples},
  author={Sun, Zhun and Ozay, Mete and Okatani, Takayuki},
  journal={arXiv preprint arXiv:1711.01791},
  year={2017}
}

@inproceedings{
shor2025adversarial,
title={Adversarial Robustness in Parameter-Space Classifiers},
author={Tamir Shor and Ethan Fetaya and Chaim Baskin and Alex M. Bronstein},
booktitle={Workshop on Neural Network Weights as a New Data Modality},
year={2025},
}

@article{schurholt2021self,
  title={Self-supervised representation learning on neural network weights for model characteristic prediction},
  author={Sch{\"u}rholt, Konstantin and Kostadinov, Dimche and Borth, Damian},
  journal={Advances in Neural Information Processing Systems},
  volume={34},
  pages={16481--16493},
  year={2021}
}

@article{schurholt2022hyper,
  title={Hyper-representations as generative models: Sampling unseen neural network weights},
  author={Sch{\"u}rholt, Konstantin and Knyazev, Boris and Gir{\'o}-i-Nieto, Xavier and Borth, Damian},
  journal={Advances in Neural Information Processing Systems},
  volume={35},
  pages={27906--27920},
  year={2022}
}

@inproceedings{
luigi2023deep,
title={Deep Learning on Implicit Neural Representations of Shapes},
author={Luca De Luigi and Adriano Cardace and Riccardo Spezialetti and Pierluigi Zama Ramirez and Samuele Salti and Luigi di Stefano},
booktitle={The Eleventh International Conference on Learning Representations },
year={2023},
}

@inproceedings{
kofinas2024graph,
title={Graph Neural Networks for Learning Equivariant Representations of Neural Networks},
author={Miltiadis Kofinas and Boris Knyazev and Yan Zhang and Yunlu Chen and Gertjan J. Burghouts and Efstratios Gavves and Cees G. M. Snoek and David W. Zhang},
booktitle={The Twelfth International Conference on Learning Representations},
year={2024},
}

@article{falk2025learning,
  title={Learning Model Representations Using Publicly Available Model Hubs},
  author={Falk, Damian and Sch{\"u}rholt, Konstantin and Tzevelekakis, Konstantinos and Meynent, L{\'e}o and Borth, Damian},
  journal={arXiv preprint arXiv:2510.02096},
  year={2025}
}

@article{horwitz2025we,
  title={We Should Chart an Atlas of All the World's Models},
  author={Horwitz, Eliahu and Kurer, Nitzan and Kahana, Jonathan and Amar, Liel and Hoshen, Yedid},
  journal={arXiv preprint arXiv:2503.10633},
  year={2025}
}

@article{hanna2026GeoSANE,
  title={GeoSANE: Learning Geospatial Representations from Models, Not Data},
  author={Hanna, Joelle and Falk, Damian and Yu, Stella and Borth, Damian},
  journal={Proceedings of the IEEE/CVF Computer Vision and Pattern Recognition (CVPR)},
  year={2026}
}

@article{peebles2022learning,
  title={Learning to learn with generative models of neural network checkpoints},
  author={Peebles, William and Radosavovic, Ilija and Brooks, Tim and Efros, Alexei A and Malik, Jitendra},
  journal={arXiv preprint arXiv:2209.12892},
  year={2022}
}

@article{liang2025drag,
  title={Drag-and-Drop LLMs: Zero-Shot Prompt-to-Weights},
  author={Liang, Zhiyuan and Tang, Dongwen and Zhou, Yuhao and Zhao, Xuanlei and Shi, Mingjia and Zhao, Wangbo and Li, Zekai and Wang, Peihao and Sch{\"u}rholt, Konstantin and Borth, Damian and others},
  journal={arXiv preprint arXiv:2506.16406},
  year={2025}
}

@inproceedings{
soro2025instructionguided,
title={Instruction-Guided Autoregressive Neural Network Parameter Generation},
author={Bedionita Soro and Bruno Andreis and Song Chong and Sung Ju Hwang},
booktitle={Workshop on Neural Network Weights as a New Data Modality},
year={2025},
}

@inproceedings{erkocc2023hyperdiffusion,
  title={Hyperdiffusion: Generating implicit neural fields with weight-space diffusion},
  author={Erko{\c{c}}, Ziya and Ma, Fangchang and Shan, Qi and Nie{\ss}ner, Matthias and Dai, Angela},
  booktitle={Proceedings of the IEEE/CVF international conference on computer vision},
  pages={14300--14310},
  year={2023}
}

@article{nava2023meta,
  title={Meta-Learning via Classifier (-free) Diffusion Guidance},
  author={Nava, Elvis and Kobayashi, Seijin and Yin, Yifei and Katzschmann, Robert K and Grewe, Benjamin},
  journal={Transactions on Machine Learning Research},
  volume={2023},
  number={8},
  year={2023},
  publisher={OpenReview}
}

@article{wang2024neural,
  title={Neural network diffusion},
  author={Wang, Kai and Tang, Dongwen and Zeng, Boya and Yin, Yida and Xu, Zhaopan and Zhou, Yukun and Zang, Zelin and Darrell, Trevor and Liu, Zhuang and You, Yang},
  journal={arXiv preprint arXiv:2402.13144},
  year={2024}
}

@article{wei2024bend,
  title={Bend: Bagging deep learning training based on efficient neural network diffusion},
  author={Wei, Jia and Zhang, Xingjun and Pedrycz, Witold},
  journal={arXiv preprint arXiv:2403.15766},
  year={2024}
}

@inproceedings{li2024beyond,
  title={Beyond Aggregation: Efficient Federated Model Consolidation with Heterogeneity-Adaptive Weights Diffusion},
  author={Li, Jiaqi and Qu, Xiaoyang and Ding, Wenbo and Zhao, Zihao and Wang, Jianzong},
  booktitle={Proceedings of the 33rd ACM International Conference on Information and Knowledge Management},
  pages={3867--3871},
  year={2024}
}

@article{ballerini2025weight,
  title={Weight Space Representation Learning on Diverse NeRF Architectures},
  author={Ballerini, Francesco and Ramirez, Pierluigi Zama and Salti, Samuele and Di Stefano, Luigi},
  journal={arXiv preprint arXiv:2502.09623},
  year={2025}
}

@inproceedings{
meynent2025structure,
title={Structure Is Not Enough: Leveraging Behavior for Neural Network Weight Reconstruction},
author={L{\'e}o Meynent and Ivan Melev and Konstantin Sch{\"u}rholt and Goeran Kauermann and Damian Borth},
booktitle={Workshop on Neural Network Weights as a New Data Modality},
year={2025},
}

@inproceedings{
soro2025diffusionbased,
title={Diffusion-based Neural Network Weights Generation},
author={Bedionita Soro and Bruno Andreis and Hayeon Lee and Wonyong Jeong and Song Chong and Frank Hutter and Sung Ju Hwang},
booktitle={The Thirteenth International Conference on Learning Representations},
year={2025},
}

@inproceedings{zhang2024metadiff,
  title={Metadiff: Meta-learning with conditional diffusion for few-shot learning},
  author={Zhang, Baoquan and Luo, Chuyao and Yu, Demin and Li, Xutao and Lin, Huiwei and Ye, Yunming and Zhang, Bowen},
  booktitle={Proceedings of the AAAI conference on artificial intelligence},
  volume={38},
  number={15},
  pages={16687--16695},
  year={2024}
}

@article{guan2025learning,
  title={Learning to Learn Weight Generation via Local Consistency Diffusion},
  author={Guan, Yunchuan and Liu, Yu and Zhou, Ke and Shen, Zhiqi and Hwang, Jenq-Neng and Li, Lei},
  journal={arXiv preprint arXiv:2502.01117},
  year={2025}
}

@article{
saul2023weightbalancing,
title={Weight-balancing fixes and flows for deep learning},
author={Lawrence K. Saul},
journal={Transactions on Machine Learning Research},
issn={2835-8856},
year={2023},
note={}
}

@article{armenta2023neural,
  title={Neural teleportation},
  author={Armenta, Marco and Judge, Thierry and Painchaud, Nathan and Skandarani, Youssef and Lemaire, Carl and Gibeau Sanchez, Gabriel and Spino, Philippe and Jodoin, Pierre-Marc},
  journal={Mathematics},
  volume={11},
  number={2},
  pages={480},
  year={2023},
  publisher={MDPI}
}

@article{zhao2022symmetry,
  title={Symmetry teleportation for accelerated optimization},
  author={Zhao, Bo and Dehmamy, Nima and Walters, Robin and Yu, Rose},
  journal={Advances in neural information processing systems},
  volume={35},
  pages={16679--16690},
  year={2022}
}

@article{neyshabur2015path,
  title={Path-sgd: Path-normalized optimization in deep neural networks},
  author={Neyshabur, Behnam and Salakhutdinov, Russ R and Srebro, Nati},
  journal={Advances in neural information processing systems},
  volume={28},
  year={2015}
}

@inproceedings{
meng2018gsgd,
title={G-{SGD}: Optimizing Re{LU} Neural Networks in its Positively Scale-Invariant Space},
author={Qi Meng and Shuxin Zheng and Huishuai Zhang and Wei Chen and Zhi-Ming Ma and Tie-Yan Liu},
booktitle={International Conference on Learning Representations},
year={2019},
}

@article{huang2020projection,
  title={Projection based weight normalization: Efficient method for optimization on oblique manifold in DNNs},
  author={Huang, Lei and Liu, Xianglong and Qin, Jie and Zhu, Fan and Liu, Li and Shao, Ling},
  journal={Pattern Recognition},
  volume={105},
  pages={107317},
  year={2020},
  publisher={Elsevier}
}

@inproceedings{yi2022accelerating,
  title={Accelerating training of batch normalization: A manifold perspective},
  author={Yi, Mingyang},
  booktitle={Uncertainty in Artificial Intelligence},
  pages={1128--1137},
  year={2022},
  organization={PMLR}
}

@inproceedings{shamsian2023data,
  title={Data Augmentations in Deep Weight Spaces},
  author={Shamsian, Aviv and Zhang, David and Navon, Aviv and Zhang, Yan and Kofinas, Miltiadis and Achituve, Idan and Valperga, Riccardo and Burghouts, Gertjan and Gavves, Efstratios and Snoek, Cees and others},
  booktitle={NeurIPS 2023 Workshop on Symmetry and Geometry in Neural Representations},
  year={2023}
}

@article{
essakine2025where,
title={Where Do We Stand with Implicit Neural Representations? A Technical and Performance Survey},
author={Amer Essakine and Yanqi Cheng and Chun-Wun Cheng and Lipei Zhang and Zhongying Deng and Lei Zhu and Carola-Bibiane Sch{\"o}nlieb and Angelica I Aviles-Rivero},
journal={Transactions on Machine Learning Research},
issn={2835-8856},
year={2025},
note={Survey Certification}
}

@inproceedings{rahaman2019spectral,
  title={On the spectral bias of neural networks},
  author={Rahaman, Nasim and Baratin, Aristide and Arpit, Devansh and Draxler, Felix and Lin, Min and Hamprecht, Fred and Bengio, Yoshua and Courville, Aaron},
  booktitle={International conference on machine learning},
  pages={5301--5310},
  year={2019},
  organization={PMLR}
}

@inproceedings{fathony2020multiplicative,
  title={Multiplicative filter networks},
  author={Fathony, Rizal and Sahu, Anit Kumar and Willmott, Devin and Kolter, J Zico},
  booktitle={International conference on learning representations},
  year={2020}
}

@article{sitzmann2020implicit,
  title={Implicit neural representations with periodic activation functions},
  author={Sitzmann, Vincent and Martel, Julien and Bergman, Alexander and Lindell, David and Wetzstein, Gordon},
  journal={Advances in neural information processing systems},
  volume={33},
  pages={7462--7473},
  year={2020}
}

@inproceedings{saragadam2023wire,
  title={Wire: Wavelet implicit neural representations},
  author={Saragadam, Vishwanath and LeJeune, Daniel and Tan, Jasper and Balakrishnan, Guha and Veeraraghavan, Ashok and Baraniuk, Richard G},
  booktitle={Proceedings of the IEEE/CVF Conference on Computer Vision and Pattern Recognition},
  pages={18507--18516},
  year={2023}
}

@inproceedings{ramasinghe2022beyond,
  title={Beyond periodicity: Towards a unifying framework for activations in coordinate-mlps},
  author={Ramasinghe, Sameera and Lucey, Simon},
  booktitle={European Conference on Computer Vision},
  pages={142--158},
  year={2022},
  organization={Springer}
}

@inproceedings{liu2024finer,
  title={Finer: Flexible spectral-bias tuning in implicit neural representation by variable-periodic activation functions},
  author={Liu, Zhen and Zhu, Hao and Zhang, Qi and Fu, Jingde and Deng, Weibing and Ma, Zhan and Guo, Yanwen and Cao, Xun},
  booktitle={Proceedings of the IEEE/CVF Conference on Computer Vision and Pattern Recognition},
  pages={2713--2722},
  year={2024}
}

@inproceedings{chan2021pi,
  title={pi-gan: Periodic implicit generative adversarial networks for 3d-aware image synthesis},
  author={Chan, Eric R and Monteiro, Marco and Kellnhofer, Petr and Wu, Jiajun and Wetzstein, Gordon},
  booktitle={Proceedings of the IEEE/CVF conference on computer vision and pattern recognition},
  pages={5799--5809},
  year={2021}
}

@article{wang2024comprehensive,
  title={A comprehensive survey of continual learning: Theory, method and application},
  author={Wang, Liyuan and Zhang, Xingxing and Su, Hang and Zhu, Jun},
  journal={IEEE transactions on pattern analysis and machine intelligence},
  volume={46},
  number={8},
  pages={5362--5383},
  year={2024},
  publisher={IEEE}
}

@article{vettoruzzo2024advances,
  title={Advances and challenges in meta-learning: A technical review},
  author={Vettoruzzo, Anna and Bouguelia, Mohamed-Rafik and Vanschoren, Joaquin and R{\"o}gnvaldsson, Thorsteinn and Santosh, KC},
  journal={IEEE transactions on pattern analysis and machine intelligence},
  volume={46},
  number={7},
  pages={4763--4779},
  year={2024},
  publisher={IEEE}
}

@article{wen2023survey,
  title={A survey on federated learning: challenges and applications},
  author={Wen, Jie and Zhang, Zhixia and Lan, Yang and Cui, Zhihua and Cai, Jianghui and Zhang, Wensheng},
  journal={International journal of machine learning and cybernetics},
  volume={14},
  number={2},
  pages={513--535},
  year={2023},
  publisher={Springer}
}

@article{han2024sadenas,
  title={SaDENAS: A self-adaptive differential evolution algorithm for neural architecture search},
  author={Han, Xiaolong and Xue, Yu and Wang, Zehong and Zhang, Yong and Muravev, Anton and Gabbouj, Moncef},
  journal={Swarm and Evolutionary Computation},
  volume={91},
  pages={101736},
  year={2024},
  publisher={Elsevier}
}

@article{schurholt2022model,
  title={Model zoos: A dataset of diverse populations of neural network models},
  author={Sch{\"u}rholt, Konstantin and Taskiran, Diyar and Knyazev, Boris and Gir{\'o}-i-Nieto, Xavier and Borth, Damian},
  journal={Advances in Neural Information Processing Systems},
  volume={35},
  pages={38134--38148},
  year={2022}
}

@inproceedings{honegger2023sparsified,
  title={Sparsified Model Zoo Twins: Investigating Populations of Sparsified Neural Network Models},
  author={Honegger, Dominik and Sch{\"u}rholt, Konstantin and Borth, Damian},
  booktitle={ICLR 2023 Workshop on Sparsity in Neural Networks},
  year={2023}
}

@inproceedings{duszenko2025towards,
  title={Towards Weight-Space Interpretation of Low-Rank Adapters for Diffusion Models},
  author={Duszenko, Jacek and Bielak, Piotr},
  booktitle={International Conference on Computational Science},
  pages={108--121},
  year={2025},
  organization={Springer}
}

@inproceedings{
ma2024implicit,
title={Implicit Zoo: A Large-Scale Dataset of Neural Implicit Functions for 2D Images and 3D Scenes},
author={Qi Ma and Danda Pani Paudel and Ender Konukoglu and Luc Van Gool},
booktitle={The Thirty-eight Conference on Neural Information Processing Systems Datasets and Benchmarks Track},
year={2024},
}

@article{kurtenbach2025open,
  title={An open dataset of neural networks for hypernetwork research},
  author={Kurtenbach, David and Shamir, Lior},
  journal={Electronics},
  volume={14},
  number={14},
  pages={2831},
  year={2025},
  publisher={MDPI}
}

@inproceedings{falk2025model,
  title={A Model Zoo of Vision Transformers},
  author={Falk, Damian and Meynent, L{\'e}o and Pfammatter, Florence and Sch{\"u}rholt, Konstantin and Borth, Damian},
  booktitle={ICLR Workshop on Neural Network Weights as a New Data Modality},
  year={2025}
}

@article{amaduzzi2025scaling,
  title={Scaling LLaNA: Advancing NeRF-Language Understanding Through Large-Scale Training},
  author={Amaduzzi, Andrea and Ramirez, Pierluigi Zama and Lisanti, Giuseppe and Salti, Samuele and Di Stefano, Luigi},
  journal={arXiv preprint arXiv:2504.13995},
  year={2025}
}

@incollection{hecht1990algebraic,
  title={On the algebraic structure of feedforward network weight spaces},
  author={Hecht-Nielsen, Robert},
  booktitle={Advanced Neural Computers},
  pages={129--135},
  year={1990},
  publisher={Elsevier}
}

@inproceedings{navon2024equivariant,
  title={Equivariant deep weight space alignment},
  author={Navon, Aviv and Shamsian, Aviv and Fetaya, Ethan and Chechik, Gal and Dym, Nadav and Maron, Haggai},
  booktitle={Proceedings of the 41st International Conference on Machine Learning},
  pages={37376--37395},
  year={2024}
}

@article{kalogeropoulos2024scale,
  title={Scale equivariant graph metanetworks},
  author={Kalogeropoulos, Ioannis and Bouritsas, Giorgos and Panagakis, Yannis},
  journal={Advances in neural information processing systems},
  volume={37},
  pages={106800--106840},
  year={2024}
}

@article{zhou2024universal,
  title={Universal neural functionals},
  author={Zhou, Allan and Finn, Chelsea and Harrison, James},
  journal={Advances in neural information processing systems},
  volume={37},
  pages={104754--104775},
  year={2024}
}

@article{pal2024model,
  title={Model lakes},
  author={Pal, Koyena and Bau, David and Miller, Ren{\'e}e J},
  journal={arXiv preprint arXiv:2403.02327},
  year={2024}
}

@inproceedings{
sourek2021lossless,
title={Lossless Compression of Structured Convolutional Models via Lifting},
author={Gustav Sourek and Filip Zelezny and Ondrej Kuzelka},
booktitle={International Conference on Learning Representations},
year={2021},
}

@article{ganev2021universal,
  title={Universal approximation and model compression for radial neural networks},
  author={Ganev, Iordan and van Laarhoven, Twan and Walters, Robin},
  journal={arXiv preprint arXiv:2107.02550},
  year={2021}
}

@article{schwarz2020graf,
  title={Graf: Generative radiance fields for 3d-aware image synthesis},
  author={Schwarz, Katja and Liao, Yiyi and Niemeyer, Michael and Geiger, Andreas},
  journal={Advances in neural information processing systems},
  volume={33},
  pages={20154--20166},
  year={2020}
}

@inproceedings{wortsman2022model,
  title={Model soups: averaging weights of multiple fine-tuned models improves accuracy without increasing inference time},
  author={Wortsman, Mitchell and Ilharco, Gabriel and Gadre, Samir Ya and Roelofs, Rebecca and Gontijo-Lopes, Raphael and Morcos, Ari S and Namkoong, Hongseok and Farhadi, Ali and Carmon, Yair and Kornblith, Simon and others},
  booktitle={International conference on machine learning},
  pages={23965--23998},
  year={2022},
  organization={PMLR}
}

@inproceedings{tang2024merging,
  title={Merging Multi-Task Models via Weight-Ensembling Mixture of Experts},
  author={Tang, Anke and Shen, Li and Luo, Yong and Yin, Nan and Zhang, Lefei and Tao, Dacheng},
  booktitle={International Conference on Machine Learning},
  pages={47778--47799},
  year={2024},
  organization={PMLR}
}

@article{zeng2025generative,
  title={Generative Modeling of Weights: Generalization or Memorization?},
  author={Zeng, Boya and Yin, Yida and Xu, Zhiqiu and Liu, Zhuang},
  journal={arXiv preprint arXiv:2506.07998},
  year={2025}
}

@inproceedings{jang2023learning,
  title={Learning to boost training by periodic nowcasting near future weights},
  author={Jang, Jinhyeok and Yun, Woo-han and Kim, Won Hwa and Yoon, Youngwoo and Kim, Jaehong and Lee, Jaeyeon and Han, ByungOk},
  booktitle={International Conference on Machine Learning},
  pages={14730--14757},
  year={2023},
  organization={PMLR}
}

@inproceedings{
knyazev2025accelerating,
title={Accelerating Training with Neuron Interaction and Nowcasting Networks},
author={Boris Knyazev and Abhinav Moudgil and Guillaume Lajoie and Eugene Belilovsky and Simon Lacoste-Julien},
booktitle={The Thirteenth International Conference on Learning Representations},
year={2025},
}

@article{tatro2020optimizing,
  title={Optimizing mode connectivity via neuron alignment},
  author={Tatro, Norman and Chen, Pin-Yu and Das, Payel and Melnyk, Igor and Sattigeri, Prasanna and Lai, Rongjie},
  journal={Advances in Neural Information Processing Systems},
  volume={33},
  pages={15300--15311},
  year={2020}
}

@inproceedings{
ainsworth2023git,
title={Git Re-Basin: Merging Models modulo Permutation Symmetries},
author={Samuel Ainsworth and Jonathan Hayase and Siddhartha Srinivasa},
booktitle={The Eleventh International Conference on Learning Representations },
year={2023},
}

@inproceedings{
entezari2022the,
title={The Role of Permutation Invariance in Linear Mode Connectivity of Neural Networks},
author={Rahim Entezari and Hanie Sedghi and Olga Saukh and Behnam Neyshabur},
booktitle={International Conference on Learning Representations},
year={2022},
}

@article{brea2019weight,
  title={Weight-space symmetry in deep networks gives rise to permutation saddles, connected by equal-loss valleys across the loss landscape},
  author={Brea, Johanni and Simsek, Berfin and Illing, Bernd and Gerstner, Wulfram},
  journal={arXiv preprint arXiv:1907.02911},
  year={2019}
}

@inproceedings{simsek2021geometry,
  title={Geometry of the loss landscape in overparameterized neural networks: Symmetries and invariances},
  author={Simsek, Berfin and Ged, Fran{\c{c}}ois and Jacot, Arthur and Spadaro, Francesco and Hongler, Cl{\'e}ment and Gerstner, Wulfram and Brea, Johanni},
  booktitle={International Conference on Machine Learning},
  pages={9722--9732},
  year={2021},
  organization={PMLR}
}

@article{theus2025generalized,
  title={Generalized Linear Mode Connectivity for Transformers},
  author={Theus, Alexander and Cabodi, Alessandro and Anagnostidis, Sotiris and Orvieto, Antonio and Singh, Sidak Pal and Boeva, Valentina},
  journal={arXiv preprint arXiv:2506.22712},
  year={2025}
}

@inproceedings{
zhao2025understanding,
title={Understanding Mode Connectivity via Parameter Space Symmetry},
author={Bo Zhao and Nima Dehmamy and Robin Walters and Rose Yu},
booktitle={Forty-second International Conference on Machine Learning},
year={2025},
}

@inproceedings{martinez2021permute,
  title={Permute, quantize, and fine-tune: Efficient compression of neural networks},
  author={Martinez, Julieta and Shewakramani, Jashan and Liu, Ting Wei and B{\^a}rsan, Ioan Andrei and Zeng, Wenyuan and Urtasun, Raquel},
  booktitle={Proceedings of the IEEE/CVF conference on computer vision and pattern recognition},
  pages={15699--15708},
  year={2021}
}

@inproceedings{abronin2024tqcompressor,
  title={TQCompressor: improving tensor decomposition methods in neural networks via permutations},
  author={Abronin, Vadim and Naumov, Aleksei and Mazur, Denis and Bystrov, Dmitriy and Tsarova, Katerina and Melnikov, Artem and Dolgov, Sergey and Brasher, Reuben and Perelshein, Michael},
  booktitle={2024 IEEE 7th International Conference on Multimedia Information Processing and Retrieval (MIPR)},
  pages={503--506},
  year={2024},
  organization={IEEE}
}

@inproceedings{
li2024merge,
title={Merge, Then Compress: Demystify Efficient {SM}oE with Hints from Its Routing Policy},
author={Pingzhi Li and Zhenyu Zhang and Prateek Yadav and Yi-Lin Sung and Yu Cheng and Mohit Bansal and Tianlong Chen},
booktitle={The Twelfth International Conference on Learning Representations},
year={2024},
}

@inproceedings{shen2024exploring,
  title={Exploring the Complexity of Deep Neural Networks through Functional Equivalence},
  author={Shen, Guohao},
  booktitle={International Conference on Machine Learning},
  pages={44643--44664},
  year={2024},
  organization={PMLR}
}

@inproceedings{
zhang2025beyond,
title={Beyond the Permutation Symmetry of Transformers: The Role of Rotation for Model Fusion},
author={Binchi Zhang and Zaiyi Zheng and Zhengzhang Chen and Jundong Li},
booktitle={Forty-second International Conference on Machine Learning},
year={2025},
}

@inproceedings{xu2024permutation,
  title={Permutation equivariance of transformers and its applications},
  author={Xu, Hengyuan and Xiang, Liyao and Ye, Hangyu and Yao, Dixi and Chu, Pengzhi and Li, Baochun},
  booktitle={Proceedings of the IEEE/CVF Conference on Computer Vision and Pattern Recognition},
  pages={5987--5996},
  year={2024}
}

@inproceedings{
lavie2024transformers,
title={Transformers' Spectral Bias and The Symmetric Group},
author={Itay Lavie and Guy Gur-Ari and Zohar Ringel},
booktitle={ICLR 2024 Workshop on Mathematical and Empirical Understanding of Foundation Models},
year={2024},
}

@inproceedings{zhang2022set,
  title={Set norm and equivariant skip connections: Putting the deep in deep sets},
  author={Zhang, Lily and Tozzo, Veronica and Higgins, John and Ranganath, Rajesh},
  booktitle={International Conference on Machine Learning},
  pages={26559--26574},
  year={2022},
  organization={PMLR}
}

@article{zhang2025symmetry,
  title={Symmetry breaking in neural network optimization: insights from input dimension expansion},
  author={Zhang, Jun-Jie and Cheng, Nan and Li, Fu-Peng and Wang, Xiu-Cheng and Chen, Jian-Nan and Pang, Long-Gang and Meng, Deyu},
  journal={npj Artificial Intelligence},
  volume={1},
  number={1},
  pages={12},
  year={2025},
  publisher={Nature Publishing Group UK London}
}

@article{imani2024exploring,
  title={Exploring group and symmetry principles in large language models},
  author={Imani, Shima and Palangi, Hamid},
  journal={arXiv preprint arXiv:2402.06120},
  year={2024}
}

@article{du2021learning,
  title={Learning signal-agnostic manifolds of neural fields},
  author={Du, Yilun and Collins, Katie and Tenenbaum, Josh and Sitzmann, Vincent},
  journal={Advances in Neural Information Processing Systems},
  volume={34},
  pages={8320--8331},
  year={2021}
}

@article{
han2024parameterefficient,
title={Parameter-Efficient Fine-Tuning for Large Models: A Comprehensive Survey},
author={Zeyu Han and Chao Gao and Jinyang Liu and Jeff Zhang and Sai Qian Zhang},
journal={Transactions on Machine Learning Research},
issn={2835-8856},
year={2024},
note={}
}

@inproceedings{
hu2022lora,
title={Lo{RA}: Low-Rank Adaptation of Large Language Models},
author={Edward J Hu and yelong shen and Phillip Wallis and Zeyuan Allen-Zhu and Yuanzhi Li and Shean Wang and Lu Wang and Weizhu Chen},
booktitle={International Conference on Learning Representations},
year={2022},
}

@article{tran2024monomial,
  title={Monomial matrix group equivariant neural functional networks},
  author={Tran, Hoang and Vo, Thieu and Huu, Tho and Nguyen, Tan and others},
  journal={Advances in Neural Information Processing Systems},
  volume={37},
  pages={48628--48665},
  year={2024}
}

@inproceedings{
sverdlov2025revisiting,
title={Revisiting Multi-Permutation Equivariance through the Lens of Irreducible Representations},
author={Yonatan Sverdlov and Ido Springer and Nadav Dym},
booktitle={The Thirteenth International Conference on Learning Representations},
year={2025},
}

@inproceedings{
vo2025equivariant,
title={Equivariant Polynomial Functional Networks},
author={Thieu Vo and Hoang V. Tran and Tho Tran Huu and An Nguyen The and Thanh Tran and Minh-Khoi Nguyen-Nhat and Duy-Tung Pham and Tan Minh Nguyen},
booktitle={Forty-second International Conference on Machine Learning},
year={2025},
}

@inproceedings{gielisse2025end,
  title={End-to-End Implicit Neural Representations for Classification},
  author={Gielisse, Alexander and van Gemert, Jan},
  booktitle={Proceedings of the Computer Vision and Pattern Recognition Conference},
  pages={18728--18737},
  year={2025}
}

@inproceedings{
kiwitt2025symmetries,
title={Symmetries in Weight Space Learning: To Retain or Remove?},
author={Fynn Kiwitt and Behrooz Tahmasebi and Stefanie Jegelka},
booktitle={High-dimensional Learning Dynamics 2025},
year={2025},
}

@inproceedings{knyazev2023can,
  title={Can we scale transformers to predict parameters of diverse imagenet models?},
  author={Knyazev, Boris and Hwang, Doha and Lacoste-Julien, Simon},
  booktitle={International Conference on Machine Learning},
  pages={17243--17259},
  year={2023},
  organization={PMLR}
}

@article{
nava2023metalearning,
title={Meta-Learning via Classifier(-free) Diffusion Guidance},
author={Elvis Nava and Seijin Kobayashi and Yifei Yin and Robert K. Katzschmann and Benjamin F Grewe},
journal={Transactions on Machine Learning Research},
issn={2835-8856},
year={2023},
note={}
}

@article{tang2025data,
  title={Data-adaptive weight-ensembling for multi-task model fusion},
  author={Tang, Anke and Shen, Li and Luo, Yong and Liu, Shiwei and Hu, Han and Du, Bo and Tao, Dacheng},
  journal={International Journal of Computer Vision},
  pages={1--17},
  year={2025},
  publisher={Springer}
}

@inproceedings{tian2025text2weight,
  title={Text2Weight: Bridging Natural Language and Neural Network Weight Spaces},
  author={Tian, Bowen and Chen, Wenshuo and Li, Zexi and Lai, Songning and Wu, Jiemin and Yue, Yutao},
  booktitle={Proceedings of the 33rd ACM International Conference on Multimedia},
  pages={10152--10160},
  year={2025}
}

@inproceedings{hedlin2025hypernet,
  title={HyperNet Fields: Efficiently Training Hypernetworks without Ground Truth by Learning Weight Trajectories},
  author={Hedlin, Eric and Hayat, Munawar and Porikli, Fatih and Yi, Kwang Moo and Mahajan, Shweta},
  booktitle={Proceedings of the Computer Vision and Pattern Recognition Conference},
  pages={22129--22138},
  year={2025}
}

@inproceedings{
saragih2025flow,
title={Flow to Learn: Flow Matching on Neural Network Parameters},
author={Daniel Saragih and Deyu Cao and Tejas Balaji and Ashwin Santhosh},
booktitle={Workshop on Neural Network Weights as a New Data Modality},
year={2025},
}

@inproceedings{finn2017model,
  title={Model-agnostic meta-learning for fast adaptation of deep networks},
  author={Finn, Chelsea and Abbeel, Pieter and Levine, Sergey},
  booktitle={International conference on machine learning},
  pages={1126--1135},
  year={2017},
  organization={PMLR}
}

@article{voulodimos2018deep,
  title={Deep learning for computer vision: A brief review},
  author={Voulodimos, Athanasios and Doulamis, Nikolaos and Doulamis, Anastasios and Protopapadakis, Eftychios},
  journal={Computational intelligence and neuroscience},
  volume={2018},
  number={1},
  pages={7068349},
  year={2018},
  publisher={Wiley Online Library}
}

@article{khurana2023natural,
  title={Natural language processing: state of the art, current trends and challenges},
  author={Khurana, Diksha and Koli, Aditya and Khatter, Kiran and Singh, Sukhdev},
  journal={Multimedia tools and applications},
  volume={82},
  number={3},
  pages={3713--3744},
  year={2023},
  publisher={Springer}
}

@article{zhang2025survey,
  title={A survey of reinforcement learning for large reasoning models},
  author={Zhang, Kaiyan and Zuo, Yuxin and He, Bingxiang and Sun, Youbang and Liu, Runze and Jiang, Che and Fan, Yuchen and Tian, Kai and Jia, Guoli and Li, Pengfei and others},
  journal={arXiv preprint arXiv:2509.08827},
  year={2025}
}

@inproceedings{gower2019sgd,
  title={SGD: General analysis and improved rates},
  author={Gower, Robert Mansel and Loizou, Nicolas and Qian, Xun and Sailanbayev, Alibek and Shulgin, Egor and Richt{\'a}rik, Peter},
  booktitle={International conference on machine learning},
  pages={5200--5209},
  year={2019},
  organization={PMLR}
}

@article{moradi2020survey,
  title={A survey of regularization strategies for deep models},
  author={Moradi, Reza and Berangi, Reza and Minaei, Behrouz},
  journal={Artificial Intelligence Review},
  volume={53},
  number={6},
  pages={3947--3986},
  year={2020},
  publisher={Springer}
}

@inproceedings{wang2024gft,
  title     = {{GFT}: Graph Foundation Model with Transferable Tree Vocabulary},
  author    = {Zehong Wang and Zheyuan Zhang and Nitesh V Chawla and Chuxu Zhang and Yanfang Ye},
  booktitle = {NeurIPS},
  year      = {2024}
}

@inproceedings{wang2025git,
  title     = {Towards Graph Foundation Models: Learning Generalities Across Graphs via Task-Trees},
  author    = {Wang, Zehong and Zhang, Zheyuan and Ma, Tianyi and Chawla, Nitesh V and Zhang, Chuxu and Ye, Yanfang},
  booktitle = {ICML},
  year      = {2025}
}

@article{wang2025scalable,
  title   = {Generative Graph Pattern Machine},
  author  = {Wang, Zehong and Zhang, Zheyuan and Ma, Tianyi and Zhang, Chuxu and Ye, Yanfang},
  journal = {NeurIPS},
  year    = {2025}
}

@inproceedings{wang2025beyond,
  title     = {Beyond Message Passing: Neural Graph Pattern Machine},
  author    = {Zehong Wang and Zheyuan Zhang and Tianyi Ma and Nitesh V Chawla and Chuxu Zhang and Yanfang Ye},
  booktitle = {ICML},
  year      = {2025}
}

@article{amari1998natural,
  title={Natural gradient works efficiently in learning},
  author={Amari, Shun-Ichi},
  journal={Neural computation},
  volume={10},
  number={2},
  pages={251--276},
  year={1998},
  publisher={MIT Press}
}

@article{bronstein2021geometric,
  title={Geometric deep learning: Grids, groups, graphs, geodesics, and gauges},
  author={Bronstein, Michael M and Bruna, Joan and Cohen, Taco and Veli{\v{c}}kovi{\'c}, Petar},
  journal={arXiv preprint arXiv:2104.13478},
  year={2021}
}

@article{wiedemer2023compositional,
  title={Compositional generalization from first principles},
  author={Wiedemer, Thadd{\"a}us and Mayilvahanan, Prasanna and Bethge, Matthias and Brendel, Wieland},
  journal={Advances in Neural Information Processing Systems},
  volume={36},
  pages={6941--6960},
  year={2023}
}

@inproceedings{tran2025equivariant,
  title={Equivariant Neural Functional Networks for Transformers},
  author={Tran, Hoang V and Vo, Thieu and Huu, Tho Tran and Nguyen-Nhat, Minh-Khoi and Tran, Thanh and Pham, Duy-Tung and Nguyen, Tan Minh and others},
  booktitle={International Conference on Learning Representations},
  year={2025},
}

@article{wang2026molecular,
  title={Molecular Representations in Implicit Functional Space via Hyper-Networks},
  author={Wang, Zehong and Han, Xiaolong and Yang, Qi and Tang, Xiangru and Wu, Fang and Guo, Xiaoguang and Sun, Weixiang and Ma, Tianyi and Lio, Pietro and Cong, Le and others},
  journal={arXiv preprint arXiv:2601.22327},
  year={2026}
}

\end{document}